\documentclass{article}

\PassOptionsToPackage{numbers, compress}{natbib}

\usepackage[preprint]{neurips_2024}

\usepackage[utf8]{inputenc} 
\usepackage[T1]{fontenc}    
\usepackage[backref=page]{hyperref}   
\usepackage{url}            
\usepackage{booktabs}       
\usepackage{amsfonts}       
\usepackage{nicefrac}       
\usepackage{microtype}      
\usepackage{xcolor}         

\usepackage{subcaption} 
\usepackage{mathtools}  
\usepackage{changes}    
\usepackage{float}      
\usepackage[acronym]{glossaries} 
\usepackage{array}      

\usepackage{comment}    
\usepackage{csquotes}   
\usepackage{verbatim}   
\usepackage{lipsum}     
\usepackage{booktabs}   
\usepackage{multirow}   
\usepackage{url}        
\usepackage{stfloats}   
\usepackage{todonotes}  
\usepackage{wrapfig}

\hypersetup{
    colorlinks=true,
    citecolor=cyan,
    linkcolor=orange,
    urlcolor=blue
}
\renewcommand{\citet}[1]{\textcolor{blue}{\citeauthor{#1}} \cite{#1}}

\newcommand{\refapp}[1]{\hyperref[#1]{\textcolor{olive}{Appendix~\ref*{#1}}}}

\renewcommand*{\backref}[1]{}
\renewcommand*{\backrefalt}[4]{%
    \ifcase #1 %
        (Not cited.)%
    \or
        (p.~#2.)%
    \else
        (pp.~#2.)%
    \fi
}

\title{Unifying Interpretability and Explainability for Alzheimer’s Disease Progression Prediction}
\author{%
  \textbf{Raja Farrukh Ali}\\
  Kansas State University\\
  \texttt{rfali@ksu.edu} \\
  \And
  \textbf{Stephanie Milani}\\
  Carnegie Mellon University\\
  \texttt{smilani@andrew.cmu.edu} \\
  \AND
  \textbf{John Woods}\\
  Kansas State University\\
  \texttt{jwoods03@ksu.edu} \\
  \And
  \textbf{Emmanuel Adeniji}\\
  Kansas State University\\
  \texttt{adeniji@ksu.edu} \\
  \And
  \textbf{Ayesha Farooq}\\
  Kansas State University\\
  \texttt{ayeshafarooq@ksu.edu} \\
  \AND
  \textbf{Clayton Mansel}\\
  University of Kansas Medical Center\\
  \texttt{cmansel@kumc.edu} \\
  \And
  \textbf{Jeffrey Burns}\\
  Alzheimer's Disease Research Center\\
  University of Kansas Medical Center\\
  \texttt{jburns2@kumc.edu} \\
  \And
  \textbf{William Hsu}\\
  Kansas State University\\
  \texttt{bhsu@ksu.edu} \\
}

\begin{document}
\maketitle

\begin{abstract}

Reinforcement learning (RL) has recently shown promise in predicting Alzheimer's disease (AD) progression due to its unique ability to model domain knowledge. However, it is not clear which RL algorithms are well-suited for this task. Furthermore, these methods are not inherently explainable, limiting their applicability in real-world clinical scenarios. Our work addresses these two important questions. Using a causal, interpretable model of AD, we first compare the performance of four contemporary RL algorithms in predicting brain cognition over 10 years using only baseline (year 0) data. We then apply SHAP (SHapley Additive exPlanations) to explain the decisions made by each algorithm in the model. Our approach combines interpretability with explainability to provide insights into the key factors influencing AD progression, offering both global and individual, patient-level analysis. Our findings show that only one of the RL methods is able to satisfactorily model disease progression, but the post-hoc explanations indicate that all methods fail to properly capture the importance of amyloid accumulation, one of the pathological hallmarks of Alzheimer's disease. Our work aims to merge predictive accuracy with transparency, assisting clinicians and researchers in enhancing disease progression modeling for informed healthcare decisions. Code is available at \url{https://github.com/rfali/xrlad}.

\end{abstract}

\section{Introduction}
\label{sec:introduction}

Alzheimer's disease (AD) is a progressive, irreversible, neurodegenerative disease that contributes to 60–70\% of dementia cases \cite{who2023}, which is the fifth leading cause of death globally \cite{gustavsson2023global}. AD is marked by a gradual reduction in brain volume and the eventual loss of neurons, resulting in declines in memory, language, cognition, and communication skills. Understanding the factors driving AD progression is essential for early diagnosis, intervention, and improved patient outcomes \cite{porsteinsson2021diagnosis} for a major health issue that affects 58 million people worldwide.

\begin{figure}[t!]
    \centering
    \includegraphics[width=0.9\linewidth]{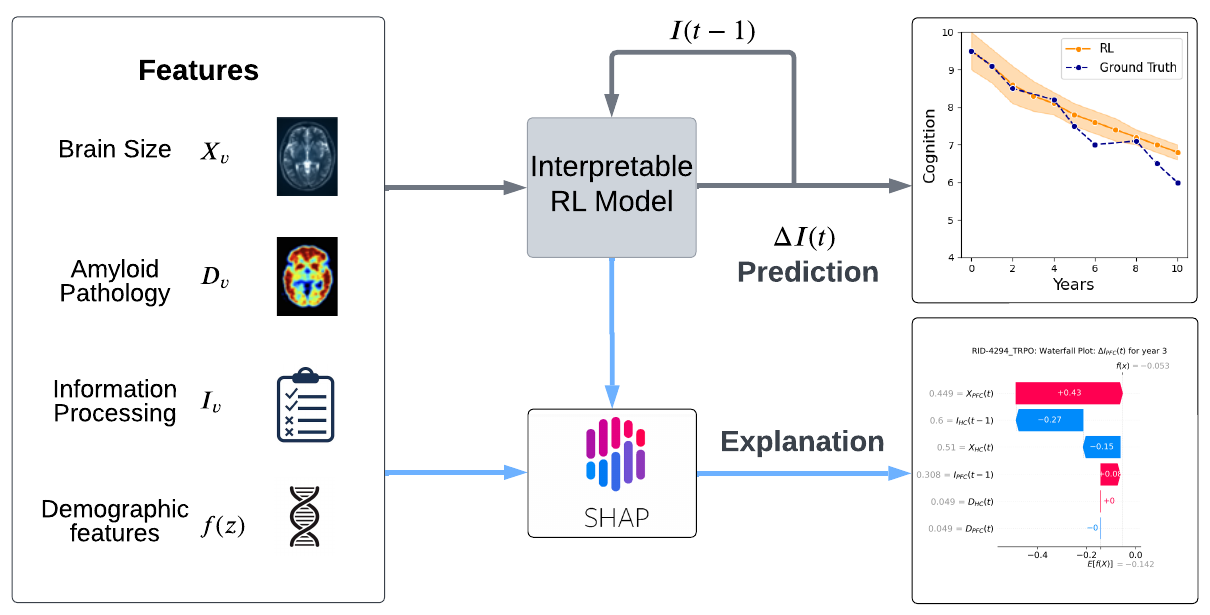}
    \caption{Our proposed framework unifies interpretability with explainability to generate cognition trajectory predictions over 10 years and associated global and local explanations for AD progression.}
    \label{fig:ixrl_framework}
\end{figure}

Clinicians diagnose AD using cognitive tests and brain imaging to assess memory, behavior changes, and brain degeneration. Given the absence of treatments to cure or prevent AD, early diagnosis is critical. While machine learning models have shown promise in predicting AD progression, their lack of interpretability and explainability pose a significant challenge to their adoption in clinical settings \cite{vellido2020importance}.
The terms \emph{interpretability} and \emph{explainability} in AI, while closely related and often used interchangeably, are distinct in their focus and application \cite{marcinkevivcs2020interpretability}. Interpretability involves creating models that are innately understandable, emphasizing a direct and transparent relationship between input and output. On the other hand, explainability focuses on elucidating how black box models generate their outputs, providing insights into their decision-making processes after the fact \cite{rudin2019stop}. We follow this convention in our work.

In response to the call for transparency in AI models, there has been a recent surge in research that uses eXplainable AI (XAI) methods for AD diagnosis \cite{viswan2024explainable}. This trend involves using a secondary, post-hoc model to explain the outputs of an \textit{opaque, black box ML model}. Such explanations derived from secondary models can be unreliable and misleading, especially in the absence of domain knowledge \cite{rudin2019stop}. 
Instead, it is recommended that the focus should be on designing inherently interpretable models. Since interpretability is domain-specific \cite{freitas2014comprehensible}, interpretable models are not only transparent but also conform to domain knowledge, including structural, causal, and physical constraints.

We posit that while interpretability may hold greater significance than explainability, both concepts are not mutually exclusive and can complement each other by offering insights into model decisions from different perspectives, improving model comprehension, and justifying complex outcomes. We focus on developing models that balance transparency with domain relevance, ensuring they are intuitive and adhere to established research, thus facilitating more reliable and interpretable solutions in the disease diagnosis and treatment process. This approach is essential in domains such as healthcare, where accurate decision-making and the ability to explain model outputs effectively can significantly impact patient outcomes.

In this work, we use a domain-knowledge and RL-based, interpretable model tasked with predicting cognition trajectories in AD patients \cite{saboo2021reinforcement} and apply a post-hoc explainability method (SHapley Additive exPlanations, SHAP \cite{lundberg2017unified}) to explain the predictions.
By modeling the brain as a system of differential equations based on causal relationships (hence interpretable), the RL agent learns to predict change in cognition (actions) as it tries to optimize cognition while minimizing associated costs. SHAP provides a method of assigning feature importance scores and quantifying the contribution of each feature to the RL decisions. By leveraging SHAP, we gain valuable insights into the relative importance of different features, shedding light on the factors driving AD progression in a more explainable manner. 

The \textbf{key contributions} of this work are:

\begin{itemize}
    \item We propose a novel Interpretable Explainable RL (IXRL) framework that combines interpretability (ante-hoc) with explainability (post-hoc) to predict 10-year cognition trajectories from baseline year-0 data, and provide explanations for model decisions (Figure \ref{fig:ixrl_framework}). To the best of our knowledge, this is one of the first works to \textit{demonstrate} a unified interpretable and explainable framework in a healthcare application.
    
    \item We compare the predictive performance of four RL methods (TRPO, PPO, DDPG, SAC) to assess their suitability to handle the complexities of this task. Moreover, due to the explainability components of the framework, we are able to understand why some methods perform better than others on the said task.
    
    \item The model's explainability aspect highlights potential failure modes as the prediction model, while being accurate in predicting long-term cognition trajectories, does not consider amyloid accumulation an important feature vector in its prediction, which is contrary to some clinical studies \cite{feld2014decrease}.
\end{itemize}


\section{Related Work}
\label{sec:related_work}

This work focuses on modeling AD \emph{progression through time}, which considers factors that affect its evolution such as brain size, activity, pathology, and cognition \cite{frizzell2022artificial,breijyeh2020comprehensive}. In contrast, a large body of work focuses on the Alzheimer's \emph{diagnosis classification} on patient data \cite{balakrishnan2023alzheimers}, which is not the focus of this work (see \refapp{app:additional_related_work} for a summary). 

\paragraph{Modeling Alzheimer’s Disease Progression.}
\label{sec:related_work_AD_progression}

Modeling AD progression involves two primary approaches: \emph{mechanistic models} and \emph{data-driven models}. The key distinction is that mechanistic models simulate known biochemistry while data-driven models perform statistical analysis on patient datasets. Mechanistic models incorporate existing biological knowledge about the underlying disease processes into mathematical representations and simulations in order to make predictions. These include network diffusion models focusing on how neurodegeneration spreads across brain networks \cite{raj2012network}, graph-based evolution models that capture topological alteration in functional brain networks \cite{vertes2012simple,li2016simulating}, regression dynamic causal modeling to infer and quantify causal relationships among brain regions \cite{frassle2018generative}, RL based models that simulate domain knowledge based differential equations to optimize a reward function \cite{saboo2021reinforcement}, and a combination of control theory and machine learning to understand the controllability of AD progression \cite{dan2023developing}. 
In contrast to mechanistic models, data-driven models utilize statistical and machine learning approaches applied directly to patient datasets to uncover patterns associating risk factors, biomarkers, and indicators of disease stage with cognitive outcomes and decline. These types of black-box models do not directly encode biological knowledge but rather seek to extract predictive signals purely from data. They encompass a spectrum of techniques, including Bayesian models \cite{fruehwirt2018bayesian}, event-based models \cite{fonteijn2012event}, mixed-effects models \cite{oxtoby2018data,liu2023review}, and machine learning models \cite{lin2018convolutional,tabarestani2018profile,diogo2022early,al2023ppad}. Mechanistic models are interpretable and can encode domain-specific knowledge whereas pure data-driven models can capture the statistical patterns within large datasets.

\paragraph{Explainable RL.}
\label{sec:related_work_XRL}

In supervised machine learning, explainability focuses on understanding the relationship between input features and the model's predictions, a task that grows complex in RL due to the inherent sequential decision-making nature of the problem. Nevertheless, recent years have seen a surge in research on XRL, as reviewed in \cite{puiutta2020explainable,heuillet2021explainability,milani2023explainable}, categorizing XRL by aspects like timing and scope of explanations. Since we use SHAP, a feature attribution post-hoc explainability method, we briefly discuss SHAP's position within these categorizations. The most prominent categorization splits XRL methods into \emph{transparent} and \emph{post-hoc} explainability methods \cite{heuillet2021explainability}. SHAP belongs to the latter as it generates explanations after a policy has been trained. Methods can also be organized according to the part of the RL agent they explain \cite{milani2023explainable}. SHAP falls in the \emph{Directly Generate Explanation} subcategory within the \emph{Feature Importance} category since given a state, SHAP generates an explanation for a non-interpretable policy's selected action after training. This enables the understanding of the factors that influence a model towards its final predictions.
Despite its widespread usage, SHAP's application in Reinforcement Learning (RL) remains notably scarce \cite{zhang2021xrlpower,meggetto2023people}, primarily because of the traditionally large state and action spaces associated with RL environments \cite{vouros2022explainable} and the challenge of seamlessly integrating SHAP with existing RL libraries. While RL has been used for AD progression prediction \cite{saboo2021reinforcement} and AD treatment planning \cite{bhattarai2023busing}, the aspect of explainability has not been addressed yet. 

\section{Methodology}
\label{sec:methodology}

\subsection{Interpretable Model for AD Progression}
\label{sec:interpretable_model}

\begin{figure}[t]
    \centering
    \includegraphics[width=0.75\linewidth]{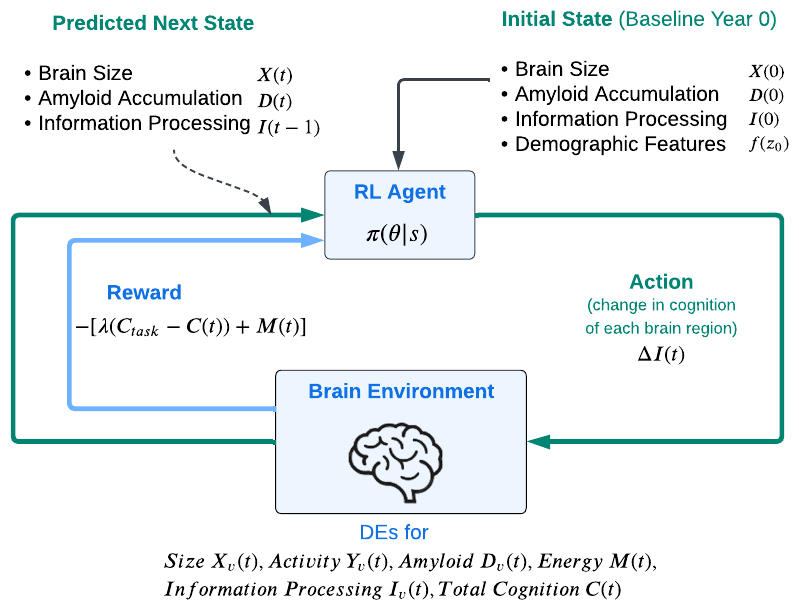}
    \caption{Interpretable RL model for AD progression prediction, expressed through DEs based on domain knowledge. The RL agent aims to predict cognition for next time step while balancing cognitive load and energy costs.} 
    \label{fig:rl_model}
\end{figure}

We provide a brief overview of the interpretable model (Figure \ref{fig:rl_model}) used to predict AD progression, based on prior work \cite{saboo2021reinforcement}. The model leverages domain knowledge to establish causal relationships between various factors involved in AD progression, described briefly as follows. Amyloid beta (A$\beta$), a key factor in AD and measured using florbetapir-PET scans, propagates between brain regions, influencing brain structure (measured via MRI), activity (measured via fMRI), and cognition (measured through tests like Mini-Mental State Examination (MMSE) \cite{folstein1975mini}, Alzheimer's Disease Assessment Scale - Cognitive Subscale 11 and 13 (ADAS11 and ADAS13) \cite{mohs1997development}. The model defines a hypothetical variable, $C_{task}$, which represents cognitive demand (theoretically required cognition) and impacts brain activity. Brain activity, in turn, affects cognition and contributes to neurodegeneration. The model also considers the energetic cost associated with brain activity, which can further contribute to neurodegeneration. 

The model defines these relationships using appropriate sets of differential equations (DEs). These known relationships are specified a priori through the graph $G = (V, E)$, where a node $v \in V$ represents a brain region, and an edge $e \in E$ represents a tract. Multiple brain regions contribute to the overall cognition. This work investigates two brain regions of interest, the hippocampus (HC) and prefrontal cortex (PFC), hence, $|V|=2$. A network diffusion model is used to model the change of amyloid in a region over time as it captures the propagation of A$\beta$ through tracts. $D_v(t)$ is the instantaneous amyloid accumulation in region $v \in V$ at time $t$, so $D(t) = [D_1(t), D_2(t), \ldots, D_{|V|}(t)]$. The total change in amyloid accumulation is represented as ${dD(t)}/{dt} = -\beta H D(t)$, where $H$ is the Laplacian of the adjacency matrix of the graph $G_s$, and $\beta$ is a constant. The total amyloid in a region $\phi_v(t)$ can then be expressed as $\phi_v(t) = \int_{0}^{t} D_v(s) \, ds$.

Cognition in brain regions is measured through the introduction of a hypothetical term \textit{information processing}, $I_v(t) \in \mathbb{R} \geq 0$, which relates a region's size and activity to its ``contribution" to cognition. The resulting model for cognition, $C(t)$, supported by the brain at time $t$ is defined as $C(t) = \sum_{v \in V} I_{v}(t)$. The activity in region $v \in V$ in support of cognition $C(t)$ at time t is denoted as $Y_v(t)$. The activity depends on both its information processing and its size ($X_v$). It is given by $Y_{v}(t) = \gamma I_{v}(t) / X_{v}(t) \: \forall v \in V$. The relationship between activity and information processing is proportional, while the relationship between activity and size is inversely proportional. Although cognition $C(t)$, brain size $X_v$, and activity $Y_v$ are related, the exact relationship among them is unknown and cannot be easily learned from limited data. 
The energetic cost $M(t)$ represents the brain's energy consumption, which is proportional to its overall activity $Y_v(t)$ and serves as a cost associated with supporting cognition. It is given by $M(t) = \sum_{v \in V} Y_{v}(t)$. $X_v(t)$ denotes the size of a brain region $v \in V$ at time $t$, and $X(t) = [X_1(t), X_2(t), ..., X_{|V|}(t)]$. Neurodegeneration is primarily influenced by amyloid deposition and brain activity. The relationship between brain activity, neurodegeneration, and A$\beta$ is given by ${dX_{v}(t)}/{dt} = -\alpha _{1}D_{v}(t) -\alpha _{2}Y_{v}(t)$.

The demographic features of patients, such as age, gender, and education, also affect AD progression. To account for these demographics in the model, parameters $\alpha _{1}$, $\alpha_{2}$, $\beta$, $\gamma$ were introduced in previous equations. For demographic features $Z_{0}$ at baseline, let $f$ be a function that approximates these parameter constants, such that $(\alpha _{1},\alpha _{2},\beta,\gamma) = f(Z_{0})$. For more details on this approximation, please see prior work \cite{saboo2021reinforcement}.

\subsection{RL framework for Optimizing Cognitive Load} 
\label{sec:rl_framework}

DEs provide relationships between some, but not all, factors relevant to AD. To address these unknown relationships, the AD model is used to formulate an optimization problem, which it solves using RL. 
The RL problem is commonly modeled as a Markov Decision Process (MDP) 
\cite{puterman2014markov}, defined by the tuple $({S,A,P,R},\gamma)$, with states $S$, actions $A$, transition function $P(s_{t+1}|s_t,a_t)$, reward function $R(s_t,a_t)$ and discount factor $\gamma$. The objective is to learn a policy $\pi(a_t|s_t)$ that maximizes the discounted cumulative return. 
\cite{bellman1957markovian}, defined as a tuple $(\mathcal{S}, \mathcal{A}, T, R, H, \gamma)$, where $\mathcal{S}$ is the set of states, $\mathcal{A}$ is the set of actions, $T: \mathcal{S} \times \mathcal{A} \times \mathcal{S} \rightarrow [0, 1]$ is the transition function, $R: \mathcal{S} \times \mathcal{A} \rightarrow \mathbb{R}$ is the reward function, $H \in \mathbb{N}$ is the time horizon, and $\gamma \in [0, 1)$ is the discount factor. Solving an MDP requires learning a policy $\pi : \mathcal{S} \times \mathcal{A} \rightarrow [0, 1]$ that maximizes the expected discounted return $\pi^* = \text{arg}\max_{\pi} \mathbb{E}_{s_{t}, a_{t}, r_{t} \sim T, \pi} [\sum_{t=0}^{H-1} \gamma^{t} r_{t+1} | s_{0} ]$.

The environment is represented here as a simulator that encompasses the differential equations that model the brain, including the variables $D_v(t)$, $\phi_v(t)$, $X_v(t)$, $Y_v(t)$, $I_v(t)$, $C(t)$ and $M(t)$. At each timestep $t$, the simulator produces a state $ S(t) = \{X_v(t), D_v(t), I_v(t - 1)\}$. The action $A(t)$ specifies the change in information processed by each brain region from the previous time step, i.e., $\Delta I_v(t)$. Formally, $A(t) = \{\Delta I_v(t) \: | \: \sum_{v \in V} I_v(t) \leq C_{\text{task}}\}$ where $I_v(t) = I_v(t - 1) + \Delta I_v(t)$. Since multiple brain regions are being modeled simultaneously such that the contribution of inter-connected regions influence the overall cognition, the number of actions per time step equals the number of regions $|V|$ being modeled. 

The RL agent aims to calculate the optimal information processing in each brain region, which together add up to the total cognition of the brain. To do so, it must balance the trade-off between two competing criteria: (i) minimizing the discrepancy between the cognitive demand of a task $C_{\text{task}}$ and the actual cognition available in the brain $C(t)$, and (ii) minimizing the cost $M(t)$ associated with supporting cognition. With $\lambda$ as a parameter controlling the trade-off between the mismatch and the cost, the agent's goal is to maximize the reward, given by $R(t) = -\left[\lambda(C_{\text{task}} - C(t)) + M(t)\right]$. This objective function aims to optimize cognitive workload distribution across brain regions. With the patient data as input to the simulator along with a domain-grounded reward function, the task is to learn to predict the change in information processing of each brain region.

\subsection{Explainability of Model Predictions}
\label{sec:explainability}

Once trained, the deep RL agent within this mechanistic model essentially becomes a black box neural network whose decisions are not inherently explainable.  
Applying such a model in critical settings like healthcare requires a comprehensive understanding of the agent's decision-making.
To address these needs in the context of AD diagnosis, we advocate for an IXRL framework that adds post-hoc explainability to the ante-hoc model, enabling a detailed evaluation and explanation of the agent's decisions. We use SHAP \cite{lundberg2017unified} for its consistency with human intuition. SHAP is a model-agnostic framework built on Shapley values \cite{shapley1953value} that has roots in cooperative game theory. Shapley values assign a value to each player based on their marginal contribution to all possible coalitions (subsets of players) to fairly allocate the total payoff of the game to each player. 

In XRL, we want to explain the policy's output in the context of the input states or features. Let $\phi_s$ be the Shapley value of a specific input feature $s \in S$ in the model, $N$ the total number of the input features/states, $\mathcal{C}$ the possible coalitions (or subsets), and $p$ the payoff function which quantifies the value each input feature adds to the prediction. Mathematically, the Shapley value of state $s \in S$  on action $a \in A$ is computed as:

\begin{equation}
\phi_s^a = \sum_{\mathcal{C} \subseteq N \setminus \{s\}} \frac{|\mathcal{C}|! \cdot(|N| - |\mathcal{C}| - 1)!}{|N|!} [p(\mathcal{C} \cup \{s\}) - p(\mathcal{C})].
\label{eq:shap}
\end{equation}

Because SHAP calculates the contribution of each input feature (state) to the action, it provides a means for understanding which state features likely contributed the most to the agent's decisions.
SHAP differs from the basic Shapley values by providing both local and global explanations, efficient computation methods for tree-based models, and by unifying existing additive feature attribution methods like Local Interpretable Model-Agnostic Explanations (LIME) \cite{ribeiro2016should}.
If $\phi_s^a$ denotes the local Shapley value attributing the contribution of state $s$ to the prediction of action $a$, $N$ is the number of states, $z' \in \{0,1\}^N$ is the coalition vector to represent the presence (1) or absence (0) of a state, then the explanation model is given by $g(z') = \phi_0 + \sum_{s=1}^{N} \phi_s z'_s$ for each model predicted action $a$. For an input instance $x$, the coalition vector $x'$ is a vector of all 1’s and the explanation model $g$ simplifies to $g(x') = \phi_0 + \sum_{s=1}^{N} \phi_s$. Similarly, given a trajectory of an RL agent such that $\tau = [S(0), A(0), S(1), A(1), \dots, S(T)]$, the global Shapley value can be computed by $\phi_s^{\text{global}} = \sum_{t \in T} \phi_s^{\text{local}}$ for all local features that contributed to the prediction of that global value. By aggregating SHAP values computed for each instance across the entire dataset, the framework provides a comprehensive perspective on the behavior of the model in predicting AD across a diverse spectrum of cases. This can help to identify significant features that consistently influence predictions. 
Local SHAP explanations ($\phi_s^a$)  provide insight into individual action prediction $\pi(s)$ given a state $s$. This type of explanation allows clinicians to inspect the \textit{specific} features that contributed to an individual prediction of the change in information processed by each brain region, and the resulting overall cognition. 


\section{Experimental Setup}
\label{sec:experimental_setup}

\paragraph{Dataset.}
\label{sec:datasets}
Data used in this work was obtained from the Alzheimer’s Disease Neuroimaging Initiative (ADNI) \cite{adni} database (\href{https://adni.loni.usc.edu}{https://adni.loni.usc.edu}). Owing to the nature of disease progression modeling, we filtered participants with baseline measurements of cognition, demographics, MRI, and florbetapir-PET scans, along with longitudinal cognitive measurements and at least 2 follow-up assessments comprising both PET and MRI scans (details in \refapp{app:implementation}). We test on two different cognitive assessment scores available for each patient as ground truth, MMSE and ADAS13). 

\paragraph{RL Algorithms.}
We evaluate four contemporary RL algorithms for modeling AD progression. Chosen for their adaptability to complex environments, they differ in policy updates, exploration strategies, and computational efficiency. We summarize each algorithm here, while providing technical details in \refapp{app:rl_methods}. \textbf{TRPO} maximizes policy improvement using KL divergence \cite{schulman2015trust}. \textbf{PPO} balances performance and simplicity with a clipping mechanism that allows stable learning \cite{schulman2017proximal}. \textbf{DDPG} learns deterministic policies for precise actions \cite{lillicrap2015continuous}. \textbf{SAC} maximizes entropy alongside reward for exploration in complex settings \cite{haarnoja2018soft}. 
We also augmented TRPO and PPO with memory networks, specifically LSTMs \cite{hochreiter1997long}.

\paragraph{Evaluation.}
Using 5-fold cross-validation, we train each agent for 1 million timesteps. During evaluation, the trained models predict 10-year cognition scores for each individual patient using only baseline year-0 data as input, generating 8800 data points per-fold (5 seeds, 11 years, 160 patients) per algorithm. Kernel-SHAP was used to generate SHAP values, which is a model-agnostic method that uses a kernel-based estimation approach to efficiently calculate Shapley values. 
We assess the performance of the algorithms using Mean Absolute Error (MAE) and Mean Squared Error (MSE).


\section{Results} 
\subsection{AD Progression Prediction using RL}

We first investigate which RL algorithms are well suited for the task of modeling AD progression.
\begin{wrapfigure}{r}{0.5\textwidth}
    \vspace{-10pt} 
    \centering
    \includegraphics[width=0.8\linewidth]{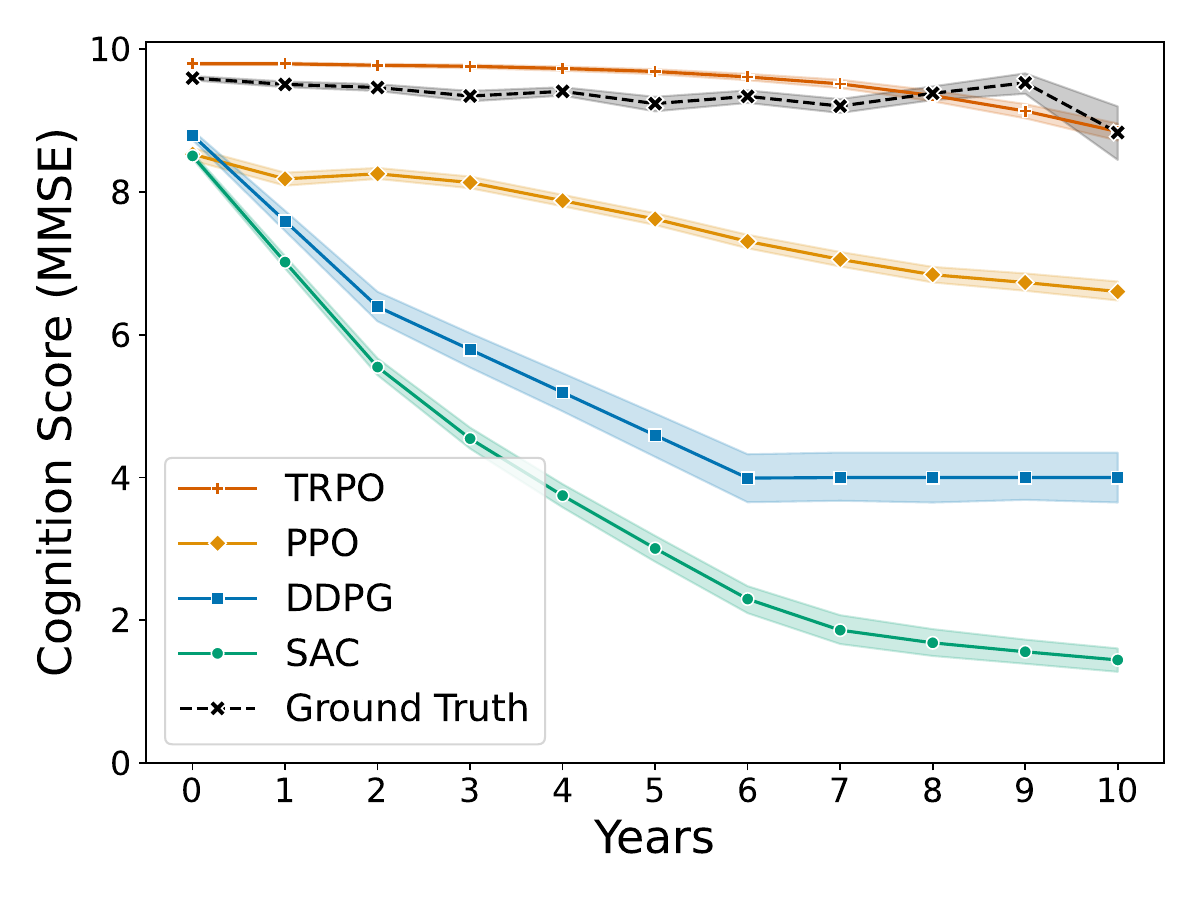}
    \vspace{-10pt} 
    \caption{Predictions on MMSE score}
    \label{fig:rl_trajectories}
\end{wrapfigure}
Figure \ref{fig:rl_trajectories} shows the results on the MMSE score  (\refapp{app:adas13} details the ADAS13 results). 
Results show that TRPO was able to model the cognitive decline quite closely while PPO, DDPG, and SAC failed to do so, on both MMSE and ADAS13 predictions (Table \ref{tab:method_scores}).
Moreover, only TRPO's performance was on par with recurrent neural networks, and augmenting TRPO with LSTMs also did not result in improved performance (\refapp{app:lstm}). 
This suggests that the DE simulator environment is sensitive to updates, highlighting TRPO's stability, especially in predicting longitudinal data, where its ability to maintain predictive update steps through trust region creation proves beneficial for stabilizing the learning process and ensuring accuracy over time.
Moreover, none of the methods except TRPO seem to exhibit recovery/compensatory mechanisms that are indicative of neurodegeneration (\refapp{app:rl_trajectory}).

\begin{table}[ht]
\centering
\setlength{\tabcolsep}{12pt}
\begin{tabular}{c c c c c}
\hline
\textbf{Method} & 
\multicolumn{2}{c}{\textbf{MMSE Score}} & 
\multicolumn{2}{c}{\textbf{ADAS13 Score}} \\
\hline
& MAE & MSE & MAE & MSE \\
\hline
TRPO & \textbf{0.572 (0.062)} & \textbf{0.86 (0.316)} & \textbf{1.153 (0.122)} & \textbf{2.084 (0.454)} \\
PPO  & 1.751 (0.387)  & 4.731 (1.859)  & 1.528 (0.802)  & 5.693 (8.101)  \\
DDPG & 3.94 (3.025)  & 27.901 (24.419)  & 4.592 (1.583)  & 28.918 (13.608)  \\
SAC  & 4.656 (0.732) & 30.764 (6.474)   & 3.979 (0.842) & 22.805 (6.964)  \\
\hline
\end{tabular}
\caption{Results for the cognition trajectory prediction of RL algorithms for two clinical cognition score types (MMSE and ADAS13). MAE: Mean Absolute Error, MSE: Mean Squared Error, parenthesis show standard deviations.}
\label{tab:method_scores}
\end{table}

\subsection{Explaining RL Predictions with SHAP}
\label{sec:explaining_rl_predictions}

\begin{figure*}[t!]
    \centering
    \begin{subfigure}{0.24\linewidth}
        \centering
        \includegraphics[width=\linewidth]{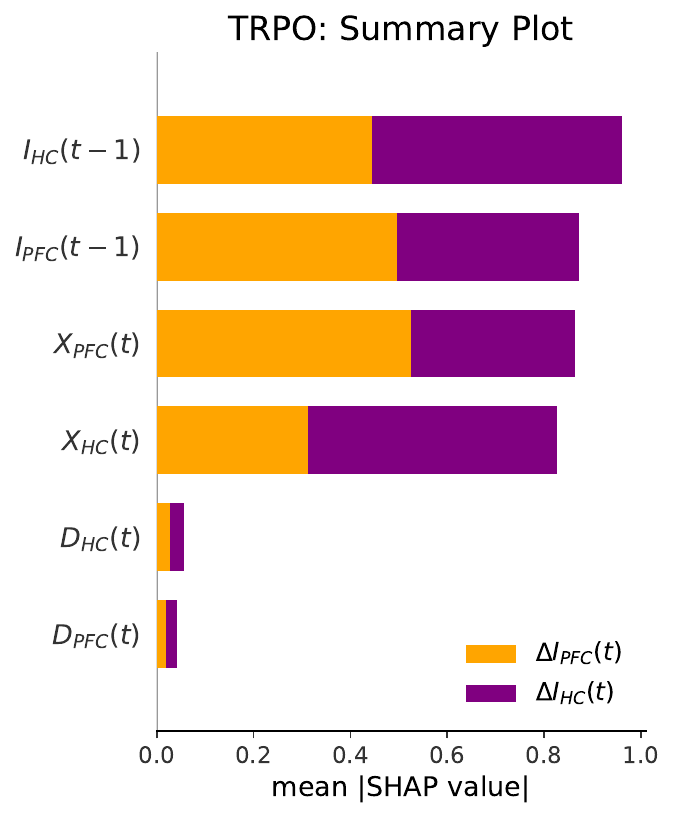}
    \end{subfigure}
    \begin{subfigure}{0.24\linewidth}
        \centering
        \includegraphics[width=\linewidth]{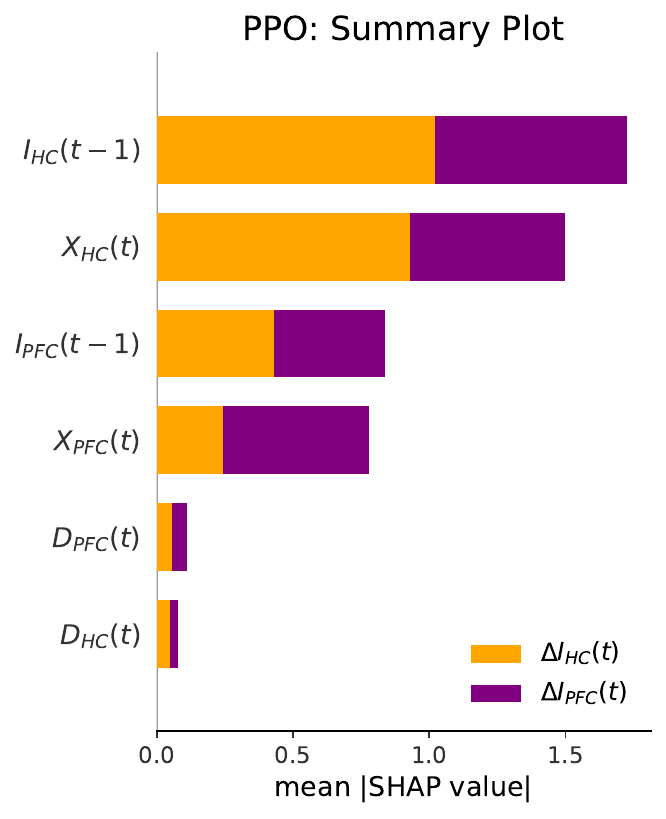}
    \end{subfigure}
    \begin{subfigure}{0.24\linewidth}
        \centering
        \includegraphics[width=\linewidth]{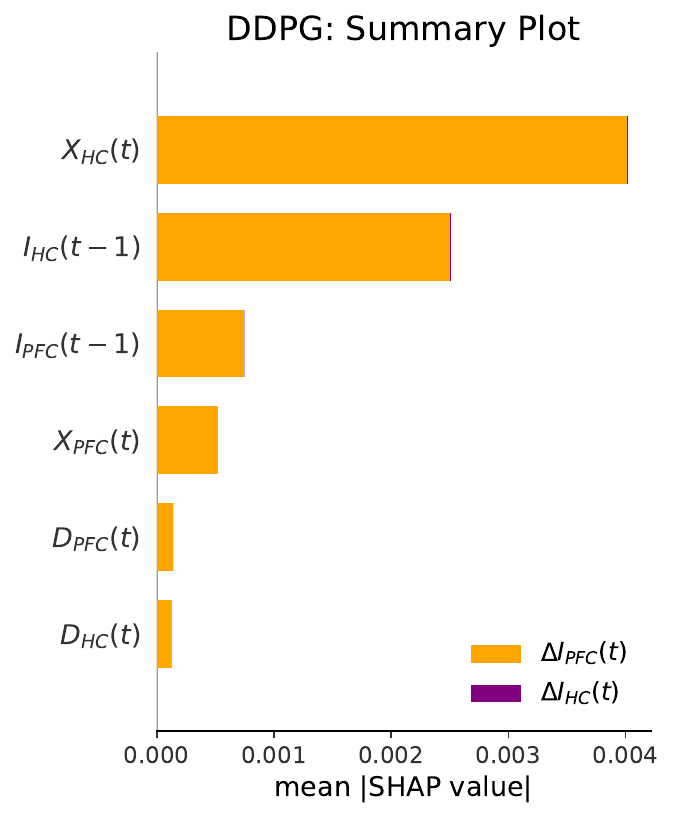}
    \end{subfigure}
    \begin{subfigure}{0.24\linewidth}
        \centering
        \includegraphics[width=\linewidth]{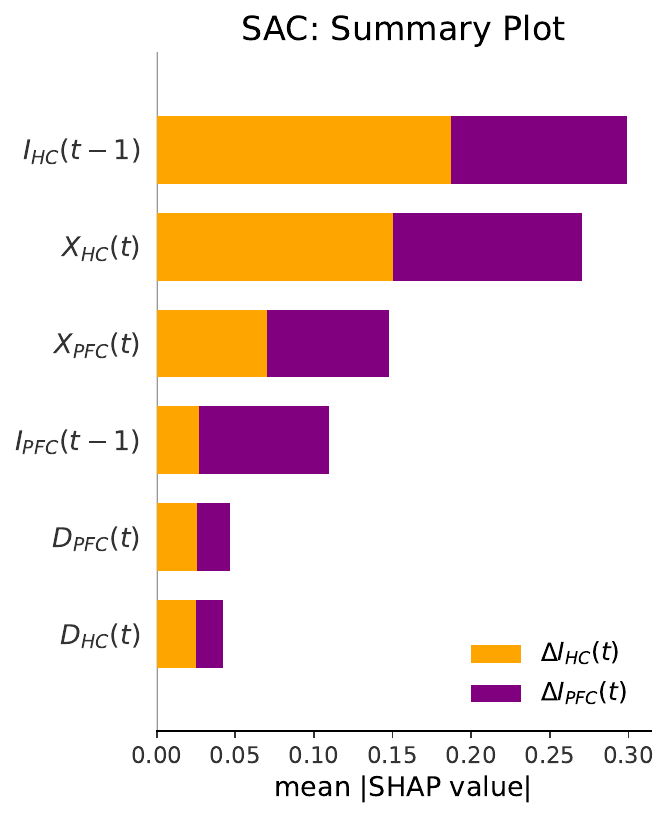}
    \end{subfigure}
    \caption{SHAP summary bar plots for all patient predictions by TRPO, PPO, DDPG, and SAC. The set of 6 input features for the 2 brain regions studied, information processing $I_v(t - 1)$, size $X_v(t)$ and amyloid accumulation $D_v(t)$, are ranked by SHAP according to their feature importance score, whereas the colors orange and purple represent their marginal effect on the two outputs/actions (change in cognition $\Delta I_v(t)$).}
    \label{fig:shap_4_rl}
\end{figure*}

\paragraph{Feature Importance for different RL algorithms.} 
We refer the reader to a primer on how to interpret SHAP plots in \refapp{app:shap_plot_summary}. Figure \ref{fig:shap_4_rl} shows the SHAP plots for all RL methods and gives a global view of which input features influenced the predicted actions (change in cognition score for each of the two regions) by each RL method. 
The bar plot ranks features by mean absolute SHAP value for information processing in the Prefrontal Cortex region $\Delta I_{PFC}$ (orange in TRPO, PPO, DDPG, purple in SAC) and the Hippocampus region $\Delta I_{HC}$. TRPO gives the most weightage to previous timestep values of information processing, hence attributing the past as the most capable predictor of the future. It then ranks the size of each region, and lastly the amyloid accumulation of each region. 
The minimal feature attribution to amyloid by all RL methods seems at first surprising as it goes against some existing clinical research suggesting that the accumulation of $A\beta$ is a significant contributor to AD progression \cite{feld2014decrease}. While a lot of research considers amyloid accumulation being an important biomarker for AD, there are differing opinions, especially from research focused on pathophysiology and treatment paradigms outside of the amyloid cascade hypothesis \cite{ashleigh2023role}. The precise mechanism through which amyloid beta contributes to cognitive decline remains elusive and our results indicate that creating accurate disease preogression models that delineate the pathogenesis of AD is imperative for the advancement of effective therapeutics.
PPO gives the highest importance to information processing in the hippocampus, followed by its size. However, these features influence the cognition prediction of the hippocampus region more than TRPO. It can also be observed that features influence one of the actions disproportionally, which could be why the method was unable to perform well and failed to show recovery/compensatory phenomenon typically linked to brain atrophy (details in Supplementary). For DDPG, the problem seems to be glaring as the magnitude of mean SHAP values is exceedingly small and none of the features seemed to influence the action $\Delta I_{HC}$, evident from the complete absence of SHAP values for that action. SAC, while performing the worst, gave feature importance scores similar to PPO. However, the magnitude of SHAP values seems to be much smaller than PPO. In the remainder of this section, we use the best-performing method (TRPO) to explain the global and local/patient predictions.

\begin{figure*}[t]    
    \begin{subfigure}{0.265\linewidth}
        \centering
        \includegraphics[width=\linewidth]{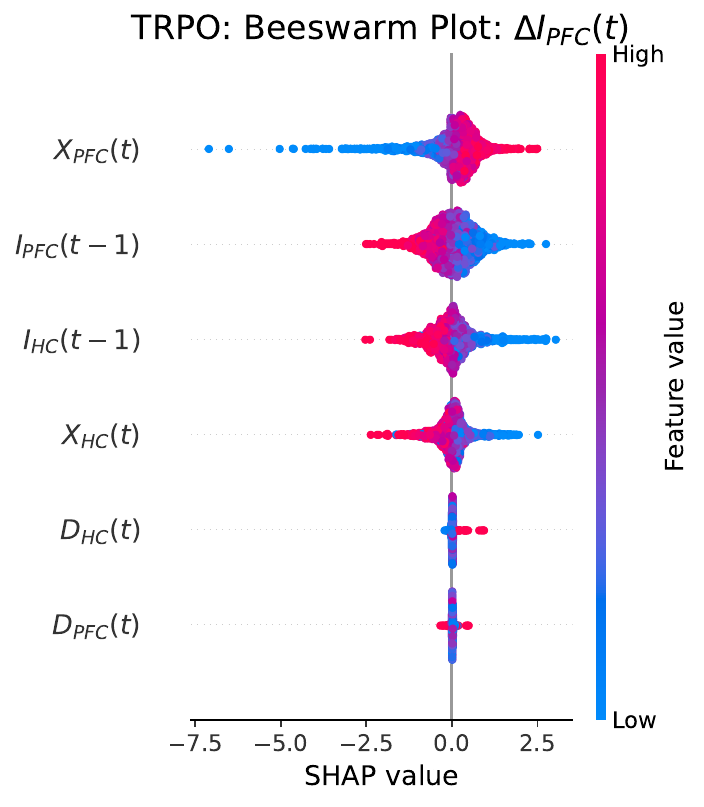}
        \caption{}
        \label{fig:beeswarm_pfc}
    \end{subfigure}
    \begin{subfigure}{0.265\linewidth}
        \centering
        \includegraphics[width=\linewidth]{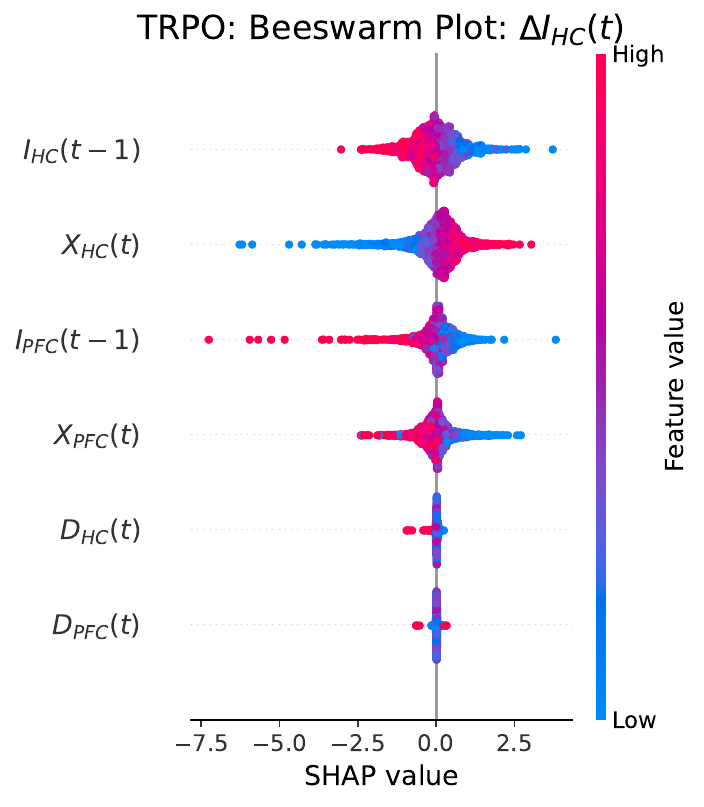}
         \caption{}
        \label{fig:beeswarm_hc}
    \end{subfigure}
    \begin{subfigure}{0.22\linewidth}
        \centering
        \includegraphics[width=\linewidth]{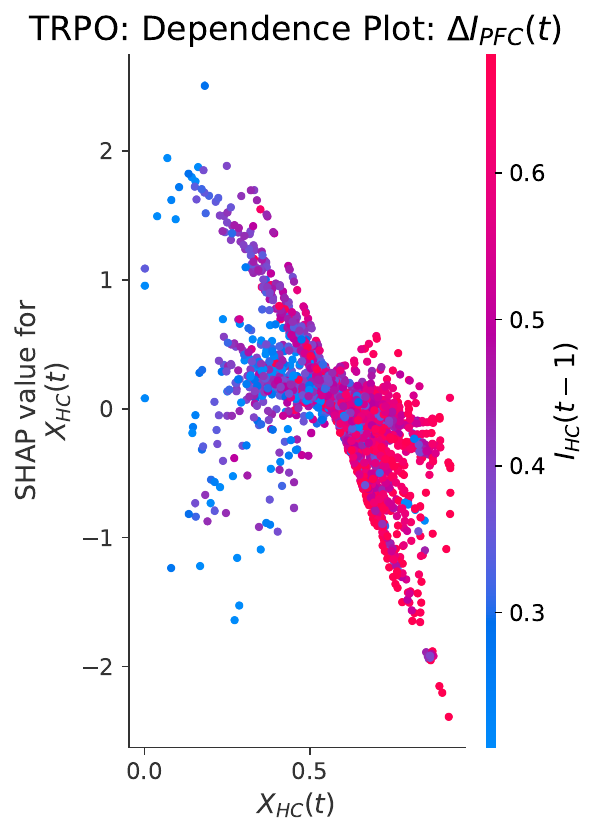}
         \caption{}
        \label{fig:dependence_x_hc}
    \end{subfigure}
    \begin{subfigure}{0.22\linewidth}
        \centering
        \includegraphics[width=\linewidth]{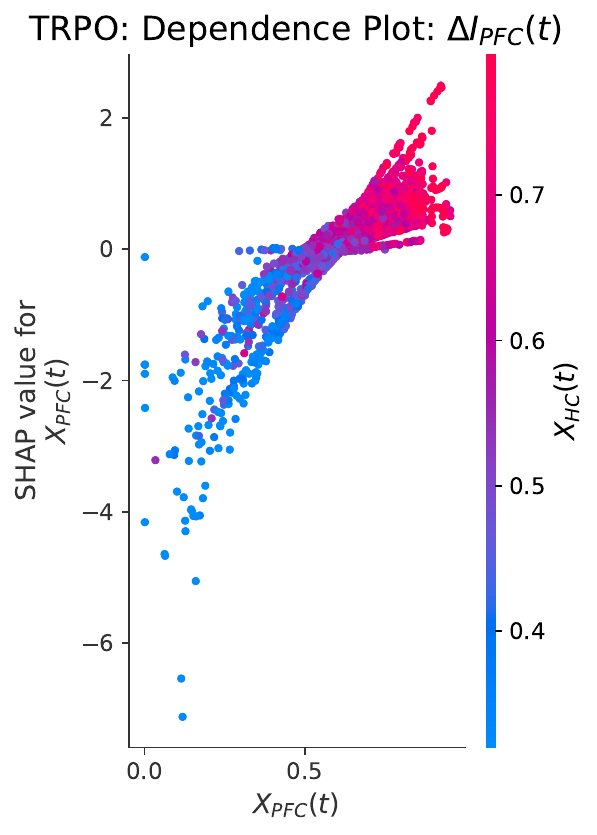}
         \caption{}
        \label{fig:dependence_x_pfc}
    \end{subfigure}
    
    \caption{SHAP plots for global predictions made by TRPO. \textbf{(a)} and \textbf{(b)}: Beeswarm plots for cognition prediction of each region, assigning distinctive colors to sample values (red high, blue low). \textbf{(c)} and \textbf{(d)}: Dependence plots capture the partial dependence between the values of a feature (x-axis, $X_v$) and its associated contribution to model prediction (y-axis, SHAP values of $X_v$).}
     \label{fig:global}
\end{figure*}

\paragraph{Global Explanations.}
Figures \ref{fig:beeswarm_pfc} and \ref{fig:beeswarm_hc} show the Beeswarm plots for each of the two predicted actions $\Delta I_v(t)$, where each dot represents one of the 1660 samples input to the model for prediction, the dot's color represents the magnitude of the feature, and the position of a dot on the x-axis represents how much that feature contributed to the output, as measured by the feature's SHAP value. Features with higher attribution to the output are placed at the top. The model considers region size $X_v(t)$ and information processing at the last timestep $I_v(t-1)$ as the most influential features when predicting change in cognition $\Delta I_v(t)$ for each brain region $v$. The plots also visualize how an increase in the size of a region increases the resulting cognition in that region ($X_{PFC}$ linearly correlates with $\Delta I_{PFC}$ and vice versa).

Figures \ref{fig:dependence_x_hc} and \ref{fig:dependence_x_pfc} are dependence plots that show the connection between brain size and cognition, with the x-axis representing feature value, and y-axis the SHAP values (contribution to the model's prediction). 
We observe that a decrease in prefrontal cortex size $X_{PFC}(t)$ causes a decrease in the model's predicted cognition (Figure \ref{fig:dependence_x_pfc}), which is in line with clinical findings that suggest smaller prefrontal cortex size corresponds with lower scores on cognitive tests \cite{yuan2014prefrontal}. Figure \ref{fig:dependence_x_hc} shows a decrease in hippocampus size $X_{HC}$ causes an increase in predicted cognition, which is evidence for recovery-compensatory effects exhibited by the model that were not programmed by design into the differential equations. Compensatory mechanisms refer to the brain's ability to adapt and partially maintain cognitive functioning in the face of neurodegeneration.
Our results show that a declining contribution of information processing $I_v(t)$, which is proportional to $X_v(t)$, in one region is compensated by an increased contribution to the model's prediction by another to maintain total cognition (Figure \ref{fig:recovery_compensatory}). In this case, the decrease in $X_{PFC}(t)$'s contribution to output $\Delta I_{PFC}(t)$ is compensated by an increase in $X_{HC}(t)$'s contribution.

\begin{figure*}[t!]
    \begin{subfigure}{0.23\textwidth}
        \centering
        \includegraphics[width=\linewidth]{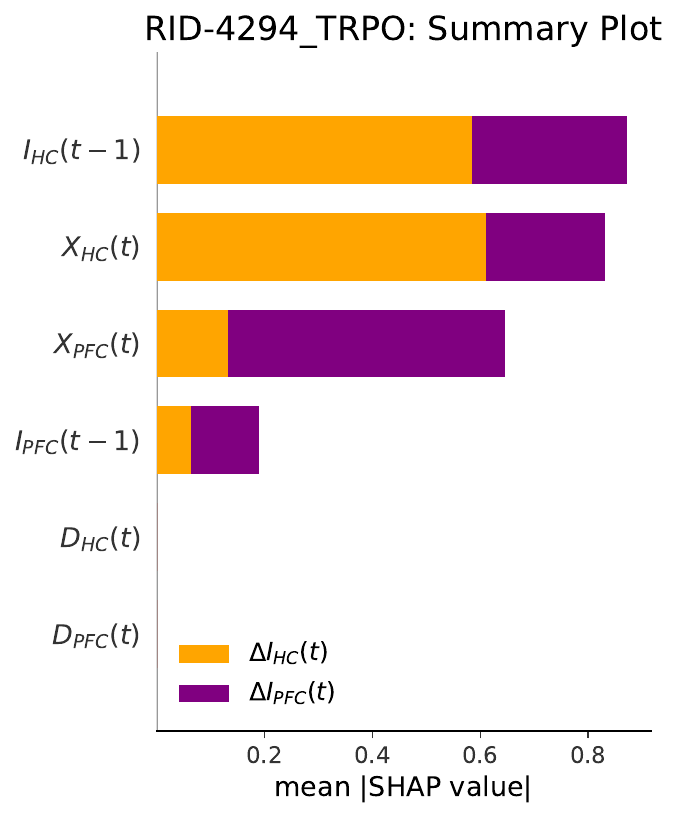}
        \caption{Summary Plot}
        \label{fig:patient_shap}
    \end{subfigure}
    \begin{subfigure}{0.23\textwidth}
        \centering
        \includegraphics[width=0.9\linewidth]{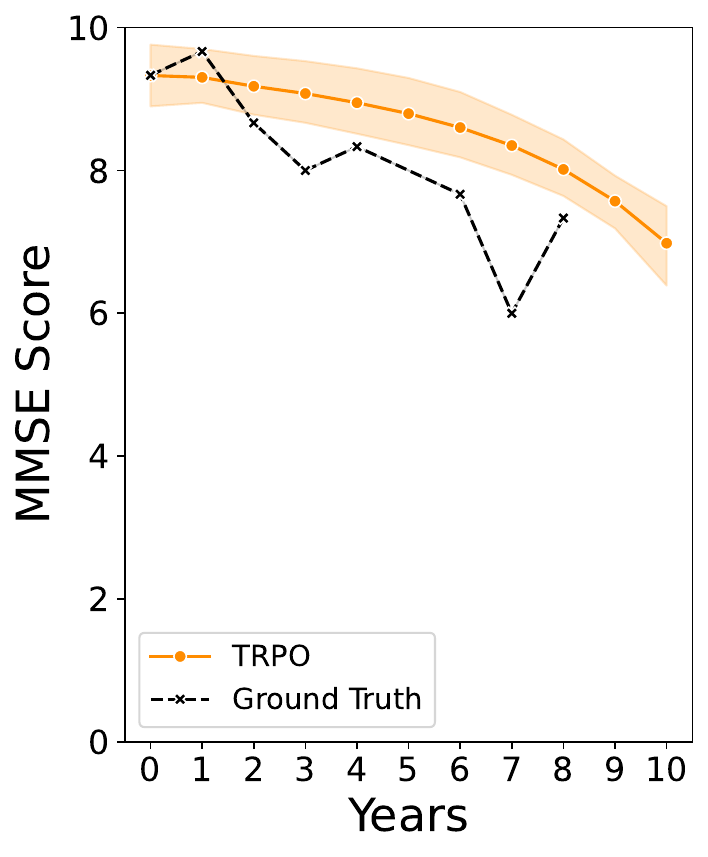}
        \caption{Total Cognition}
        \label{fig:patient_cognition}
    \end{subfigure}
    \begin{subfigure}{0.26\textwidth}
        \centering
        \includegraphics[width=\linewidth]{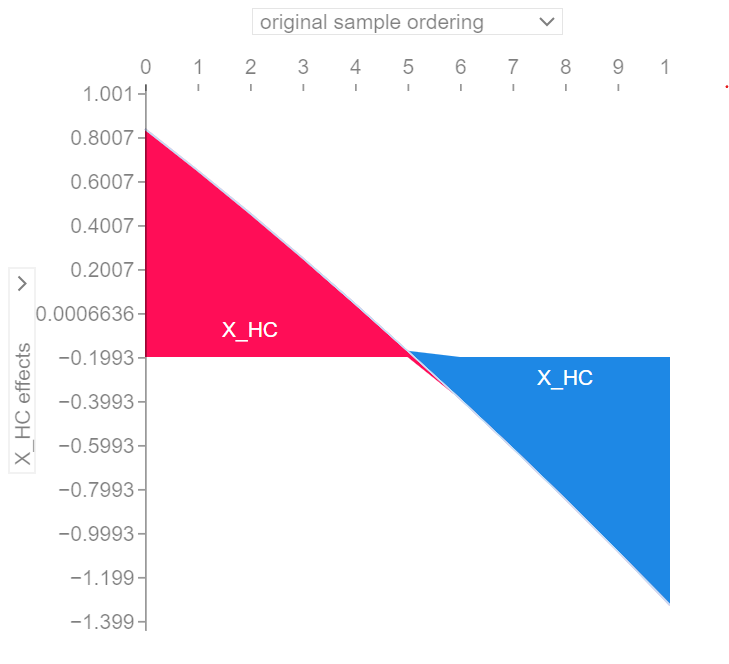}
        \caption{Effect of $X_{HC}$}
        \label{fig:patient_shap_hc_size}
    \end{subfigure}
    \begin{subfigure}{0.26\textwidth}
        \centering
        \includegraphics[width=0.92\linewidth]{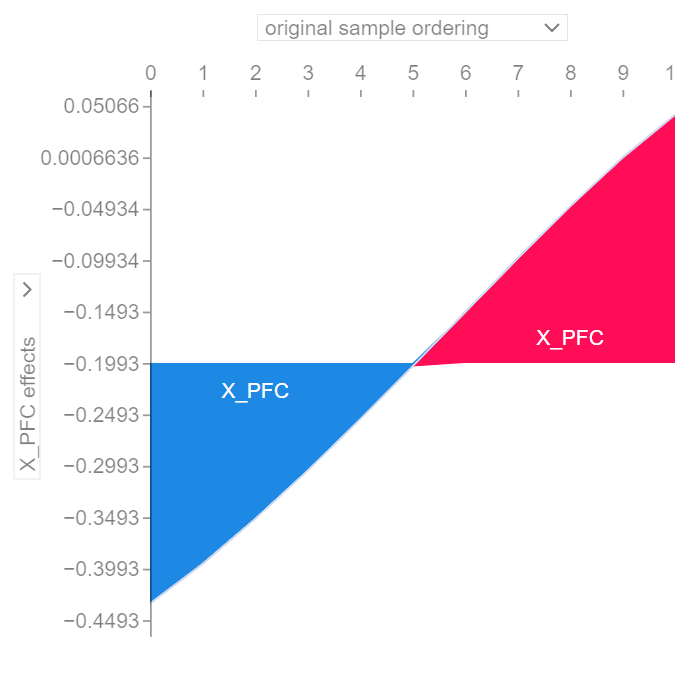}
        \caption{Effect of $X_{PFC}$}
        \label{fig:patient_shap_pfc_size}
    \end{subfigure}
    \caption{RL predicted trajectories and SHAP plots for a representative patient (Record ID 4294)}
    \label{fig:per_patient}
\end{figure*}
\paragraph{Local Explanations.}
Figure \ref{fig:per_patient} shows the local, per-patient explanations for a particular patient selected due to sufficient decline in actual cognition score. Figure \ref{fig:patient_shap} summarizes the effect each of the 6 features has on the model's actions, $\Delta I_{v}(t)$. This patient appears to be representative of the global population since $I_v(t-1)$ and $X_v(t)$ contribute most to the model's prediction of $\Delta I_v(t)$ for each brain region, whereas Figure \ref{fig:patient_cognition} highlights the accuracy of the TRPO model at predicting cognition over time. The model's decisions can be contextualized through time to see how they change as AD progresses, as shown in Figures \ref{fig:patient_shap_hc_size} and \ref{fig:patient_shap_pfc_size} with y-axis depicting years 0 to 10.
They show that during the initial years, a decrease in $X_{PFC}$'s contribution to the output $\Delta I_{HC}(t)$ (shown by blue) is compensated by an increase in $X_{HC}$'s contribution to the output $\Delta I_{HC}(t)$ (shown by red), confirming the existing of compensatory mechanisms. While predictions can be compared against ground truth values, SHAP plots delineate which model inputs influenced its output, thereby also explaining the decision-making of black-box RL models.

\section{Discussion}
\label{sec:discussion}
The integration of ML in healthcare has been approached with caution, and for good reason. These models must be made explainable before they can be confidently utilized in clinical settings. We posit that the methodology presented in this work represents a step in that direction. By explaining the decisions of an RL algorithm trained to predict cognition trajectories for early diagnosis, we aim to facilitate the gradual and manageable adoption of ML into healthcare applications. 

Based on the results of our proposed framework, we identified a potential limitation in current modeling approaches; an RL method that accurately predicted disease progression did not attribute any significance to a hypothesized AD biomarker given as input (amyloid) in its decision-making process.
Amyloid accumulation is not only a pathological hallmark of the disease, but has historically been the predominant hypothesized causal factor in AD pathogenesis \citep{karran2011amyloid}. However, given the relative failure of therapies aimed at removing amyloid from the brain in clinical trials \cite{fedele2023anti, pleen2022alzheimer}, new hypotheses have been proposed to underlie AD progression \cite{levin2022testing,armstrong2011pathogenesis}. Our IXRL framework further supports evidence that amyloid is more of a downstream pathologic marker and not a key cause of neurodegeneration and cognitive decline. Future modeling studies should attempt to incorporate non-amyloid biomarkers as features such as tau accumulation, or neuroinflammation \cite{leng2021neuroinflammation}. 

There are limitations to this work. First, this work focuses on two brain regions, and future work can incorporate more brain regions to model more complex relationships. Second, the training and evaluation sets share the same data distribution (ADNI dataset), which may lead to inconsistent performance evaluation and explainability of RL methods. Third, the efficacy of domain-guided interpretable models hinges on the precision of the embedded domain knowledge; inaccuracies in this knowledge could lead to flawed model outputs. Finally, better explainability methods need to be developed to address some recent critiques of Shapley-value-based explanation methods \cite{kumar2020problems,huang2023inadequacy}.

One can argue that SHAP explains how the model predicts AD progression, not how AD progression occurs. However, using an interpretable model for AD progression prediction alleviates this concern to some extent. Possible future work may include experimenting with a different model or adding new features (such as different brain regions) to the model. If a model that incorporates other brain regions or biomarkers is used to predict the progression of AD, will these features still have the same importance? Additionally, if a new reward model was chosen, will the features still have the same importance rank?  Future work can help answer these questions and shed light into the accuracy of the SHAP explanations while highlighting any weaknesses. 

We demonstrate the value of an IXRL framework for modeling and explaining AD progression prediction. By combining interpretable models that encode causal relationships with explainability methods like SHAP, we show how different RL algorithms predict cognition trajectories up to 10 years post-diagnosis. Our SHAP analysis identified that increased information processing and reduced brain size appear to contribute most to cognitive decline in both of the studied brain regions. We believe that with better explainability tools and more accurate, complex brain models, we can uncover further insights into AD progression modeling.
It would also be interesting to investigate different XAI techniques for this domain, along with feedback and insights from clinicians and domain experts. 
These steps would contribute to our shared goal; to translate these findings into actionable insights for early prediction and intervention for improved AD-related healthcare.

\section*{Acknowledgments}
The authors are grateful to Krishnakant Saboo, Heather Bailey and Michael Young for discussions, and to Vinny Sun for plotting help. This work was supported in part by the KDD Excellence Fund at Kansas State University. We gratefully acknowledge the anonymous reviewers for their critical and insightful feedback. Finally, this research was enriched by the broader academic community, and we appreciate the shared knowledge and resources.

\bibliographystyle{unsrtnat} 
\bibliography{main}

\clearpage
\appendix

\centerline{\textbf{\huge Appendix}}

\section{Summary of SHAP Plots}
\label{app:shap_plot_summary}

We provide a brief description for each of the plot used in this paper (for additional details and examples, please see the SHAP Documentation \href{https://shap.readthedocs.io}{https://shap.readthedocs.io}). Each SHAP plot offers a unique lens to scrutinize the complex decision-making mechanism of a machine learning model.

\begin{itemize}
  \item \textbf{Summary Bar Plot}: This plot ranks features based on the mean absolute SHAP values across all instances. It offers a global view of feature importance, with larger bars indicating greater influence on the model’s output. The plot is presented as a horizontal bar chart, making it clear which features are most important in the model's decision-making process.
  
  \item \textbf{Beeswarm Plot}: For a more detailed global interpretation, the beeswarm plot positions individual SHAP values of all features for all samples on a chart, resembling a swarm of bees. This visualization clusters points to demonstrate the distribution of the impacts each feature has on the prediction, with color intensity often representing the feature value (red means higher feature value and blue being lower, by default).

  \item \textbf{Dependence Plot}: This plot shows the effect of a single feature across the whole dataset, reflecting the relationship between the feature’s value and its SHAP value. It can also highlight potential interactions between features when color coding is applied to represent another feature (typically chosen automatically). If there's an interaction effect between this additional feature and the one being plotted, it will manifest as a unique vertical coloring pattern.

  \item \textbf{Force Plot}: A local explainability tool, the force plot illustrates how each feature’s SHAP value pushes the prediction away from the base value. This is especially useful for dissecting individual predictions and understanding the tug-of-war between features leading to the final model output.

  \item \textbf{Stacked Force Plot}: Similar to the force plot, this variant stacks individual force plots for multiple instances together, useful for comparing the explanations of several predictions at once. In our work, we stacked individual force plots for a patient's predictions over 10 years to create a stacked force plot (see Fig 7c and 7d in main).

  \item \textbf{Waterfall Plot}: This local explainability plot sequentially adds feature contributions on top of a base value to arrive at the final prediction. It provides a step-by-step breakdown of how each feature’s SHAP value cumulatively influences the prediction, depicted as a "waterfall" of contributions.

  \item \textbf{Decision Plot}: The decision plot takes the concept of the waterfall plot further by plotting the cumulative path of SHAP values leading to the final prediction. This plot can be extended to show multiple instances together, providing a decision path comparison among them.
\end{itemize}

\section{Reinforcement Learning Algorithms}
\label{app:rl_methods}

\paragraph{TRPO.}
\label{app:trpo}
Trust Region Policy Optimization (TRPO) \cite{schulman2015trust} is an on-policy algorithm that guarantees monotonic policy improvement. The key idea of TRPO is to constrain the local variation of the parameters to a ``trust region" in the policy space to ensure the update steps of the policy are the biggest possible improvement. The constraint $\delta$ on the variation of parameters is determined by KL Divergence. The theoretical update for the policy $\pi_\theta$ with parameters $\theta$ is defined as:   

\begin{equation}
  \theta_{k+1} = \arg \max_{\theta} \mathcal{L}(\theta_k, \theta) \quad \text{s.t.} \ \Bar{D}_{KL}(\theta || \theta_k) \leq \delta
\end{equation}

where $\mathcal{L}(\theta_k, \theta)$ is the surrogate advantage that measures the relative performance of $\pi_\theta$ to $\pi_{\theta_k}$, defined as:

\begin{equation}
  \mathcal{L}(\theta_k, \theta) = \underset{s,a \sim \pi_{\theta_k}}{\mathbb{E}} \left[ \frac{\pi_\theta(a|s)}{\pi_{\theta_k}(a|s)} A^{\pi_{\theta_k}}(s,a) \right],
\end{equation}

and $\Bar{D}_{KL}(\theta || \theta_k)$ is the KL-Divergence between the two policies, defined as:

\begin{equation}
  \Bar{D}_{KL}(\theta || \theta_k) = \underset{s \sim \pi_{\theta_k}}{\mathbb{E}} \left[ D_{KL}(\pi_\theta(\cdot|s) || \pi_{\theta_k}(\cdot|s)) \right].
\end{equation}

\paragraph{PPO.}
\label{app:ppo}
Proximal Policy Optimization (PPO) \cite{schulman2017proximal} is a model-free policy gradient algorithm designed to address the limitations of previous policy optimization methods like TRPO. Unlike TRPO, which only performs one policy update per step, PPO performs multiple epochs of Stochastic Gradient Ascent for each update, allowing for a more general and sample complex algorithm. To constrain policy updates, PPO clips the objective function instead of using KL Divergence like in TRPO. The policy update for $\pi_\theta$ with parameters $\theta$ is defined as:

\begin{equation}
  \theta_{k+1} = \arg \max_{\theta} \underset{s,a \sim \pi_{\theta_k}}{\mathbb{E}} [L(s,a,\theta_k,\theta)],
\end{equation}

where $L(s,a,\theta_k,\theta)$ is the objective function defined as:

\begin{equation}
    L(s,a,\theta_k,\theta) = \min\left(\frac{\pi_{\theta}(a|s)}{\pi_{\theta_{k}}(a|s)} A^{\pi_{\theta_{k}}}(s,a), \text{clip}\left(\frac{\pi_{\theta}(a|s)}{\pi_{\theta_{k}}(a|s)}, 1 - \epsilon, 1 + \epsilon\right) A^{\pi_{\theta_{k}}}(s,a) \right)
\end{equation}

\paragraph{DDPG.}
\label{app:ddpg}
Deep Deterministic Policy Gradient (DDPG) \cite{lillicrap2015continuous} is a model-free, off-policy actor-critic algorithm for learning policies in high-dimensional, continuous action spaces. It combines ideas from DPG (Deterministic Policy Gradient) \cite{silver2014deterministic} and DQN (Deep Q-Network) \cite{mnih2013playing}, employing a replay buffer to sample experience transitions and target networks to stabilize training. DDPG aims to learn both a Q-function and a policy simultaneously. The Q-function is trained by minimizing the mean-squared Bellman error:

\begin{equation}
    L(\phi, \mathcal{D}) = \underset{(s,a,r,s',d) \sim \mathcal{D}}{\mathbb{E}}\left[ \left( Q_{\phi}(s,a) - \left(r + \gamma Q_{\phi_{\text{targ}}}(s', \mu_{\theta_{\text{targ}}}(s')) \right) \right)^2 \right] 
\end{equation}

where $ \mathcal{D} $ is the replay buffer containing previous experiences, $ \phi $ represents the parameters of the Q-function network, $ \phi_{\text{targ}} $ represents the target Q-network and $\mu_{\theta_{\text{targ}}}$ is the target policy. DDPG approximates $ \max_a Q^*(s,a) $ with $ Q(s,\mu(s)) $, leveraging the differentiability of $ Q^*(s,a) $ with respect to the action. Policy learning in DDPG involves learning a deterministic policy $ \mu_{\theta}(s) $ that maximizes the estimated Q-value. This is achieved by performing gradient ascent with respect to the policy parameters $ \theta $:

\begin{equation}
    \max_{\theta} \underset{s \sim \mathcal{D}}{\mathbb{E}}\left[ Q_{\phi}(s, \mu_{\theta}(s)) \right] 
\end{equation}

\paragraph{SAC.}
\label{app:sac}
Soft Actor-Critic (SAC) \cite{haarnoja2018soft} is an off-policy, model-free algorithm designed for environments with continuous action spaces, utilizing an actor-critic architecture. SAC is based on the maximum entropy reinforcement learning framework, which encourages exploration by maximizing both the expected return and the entropy of the policy, preventing premature convergence to suboptimal policies and improving learning stability. 

SAC concurrently learns a policy $\pi_{\theta}$ and two Q-functions $Q_{\phi_1}, Q_{\phi_2}$. The reward function is modified to include an entropy term $r_t' = r_t + \alpha H(\pi(\cdot|s_t))$ where \( \alpha \) is the trade-off coefficient controlling the importance of the entropy term, and \( H(\pi(\cdot|s_t)) \) is the entropy of the policy. The Q-value function and the loss functions for the Q-networks in SAC are defined as:
\begin{equation}
Q^{\pi}(s,a) \approx r + \gamma\left(Q^{\pi}(s',\tilde{a}') - \alpha \log \pi(\tilde{a}'|s') \right)
\end{equation}
\begin{equation}
L(\phi_i, {\mathcal D}) = \underset{\tau \sim {\mathcal D}}{{\mathrm E}}\left[
    \Bigg( Q_{\phi_i}(s,a) - r + \gamma  \left( \min_{j=1,2} Q_{\phi_{\text{targ},j}}(s', \tilde{a}') - \alpha \log \pi_{\theta}(\tilde{a}'|s') \right), \Bigg)^2
    \right],
\end{equation}
where $\tilde{a}' \sim \pi_{\theta}(\cdot|s')$. To learn the policy $\pi_{\theta}$,it should maximize $V^{\pi}(s)$, which is defined as:
\begin{equation}
V^{\pi}(s) = \mathbb{E}_{a \sim \pi}[Q^{\pi}(s,a)] + \alpha H\left(\pi(\cdot|s)\right)
= \mathbb{E}_{a \sim \pi}[Q^{\pi}(s,a) - \alpha \log \pi(a|s)].
\end{equation}

To get the policy loss, SAC uses $\min_{j=1,2} Q_{\phi_j}$ (the minimum of the two Q approximators). The policy is optimized as follows:
\begin{equation}
\max_{\theta} \mathbb{E}_{s \sim \mathcal{D}, \, \xi \sim \mathcal{N}} \left[ \min_{j=1,2} Q_{\phi_j}(s,\tilde{a}_{\theta}(s,\xi)) - \alpha \log \pi_{\theta}(\tilde{a}_{\theta}(s,\xi)|s) \right].
\end{equation}
where $\tilde{a}_{\theta}(s, \xi) = \tanh\left( \mu_{\theta}(s) + \sigma_{\theta}(s) \odot \xi \right)$ and $ \xi \sim \mathcal{N}(0, I)$.

Unlike DDPG, the actor for SAC employs a stochastic policy so that the actor's output is a probability distribution over actions. SAC also utilizes entropy regularization and automatic temperature tuning. By regularizing entropy, the measure of randomness or uncertainty in the policy, SAC aims to maximize the cumulative reward and entropy. This encourages policies with more diverse and exploratory actions. SAC incorporates automatic temperature tuning, which dynamically adjusts the weight of the entropy term in the objective function. This temperature parameter is crucial for balancing the trade-off between exploration and exploitation, ensuring that the policy does not become overly deterministic or excessively random but instead seeks to maximize the expected cumulative reward while maintaining a beneficial degree of exploratory behavior.

\section{Augmenting RL methods with Memory Based Architectures}
\label{app:lstm}

We integrate Long Short-Term Memory (LSTM) networks \cite{hochreiter1997long} with the two best performing RL methods identified, aiming to enhance their performance in sequential decision problems involving long temporal dependencies. We compare these LSTM-augmented RL algorithms against pure recurrent methods, miniRNN and Suppor Vector Regression (SVR) \cite{nguyen2020predicting}, as baselines. 

LSTMs are a class of artificial neural networks designed to recognize patterns in sequential data, renowned for their ability to retain information over extended periods, a crucial capability for tasks involving temporal dependencies. By integrating LSTMs into RL algorithms, we enable the agents to infer hidden states over time and make informed decisions based on the history of interactions with the environment, particularly in partially observable environments or where the optimal policy depends on sequences of previous observations and actions.

Specifically, we use TRPO-LSTM and PPO-LSTM, which combine the Trust Region Policy Optimization (TRPO) and Proximal Policy Optimization (PPO) algorithms, respectively, with LSTM networks. TRPO-LSTM leverages LSTMs to capture temporal dependencies, making it well-suited for tasks requiring memory of past events, unlike standard TRPO which focuses on single time-step decisions \cite{schulman2015trust}. PPO-LSTM integrates PPO's optimization strategy with LSTMs, enabling effective handling of sequential data while benefiting from simpler and more computationally efficient policy updates compared to the stricter constraints of TRPO-LSTM \cite{schulman2017proximal}.

\begin{figure*}[t!]    
    \begin{subfigure}{0.32\linewidth}
        \centering
        \includegraphics[width=\linewidth]{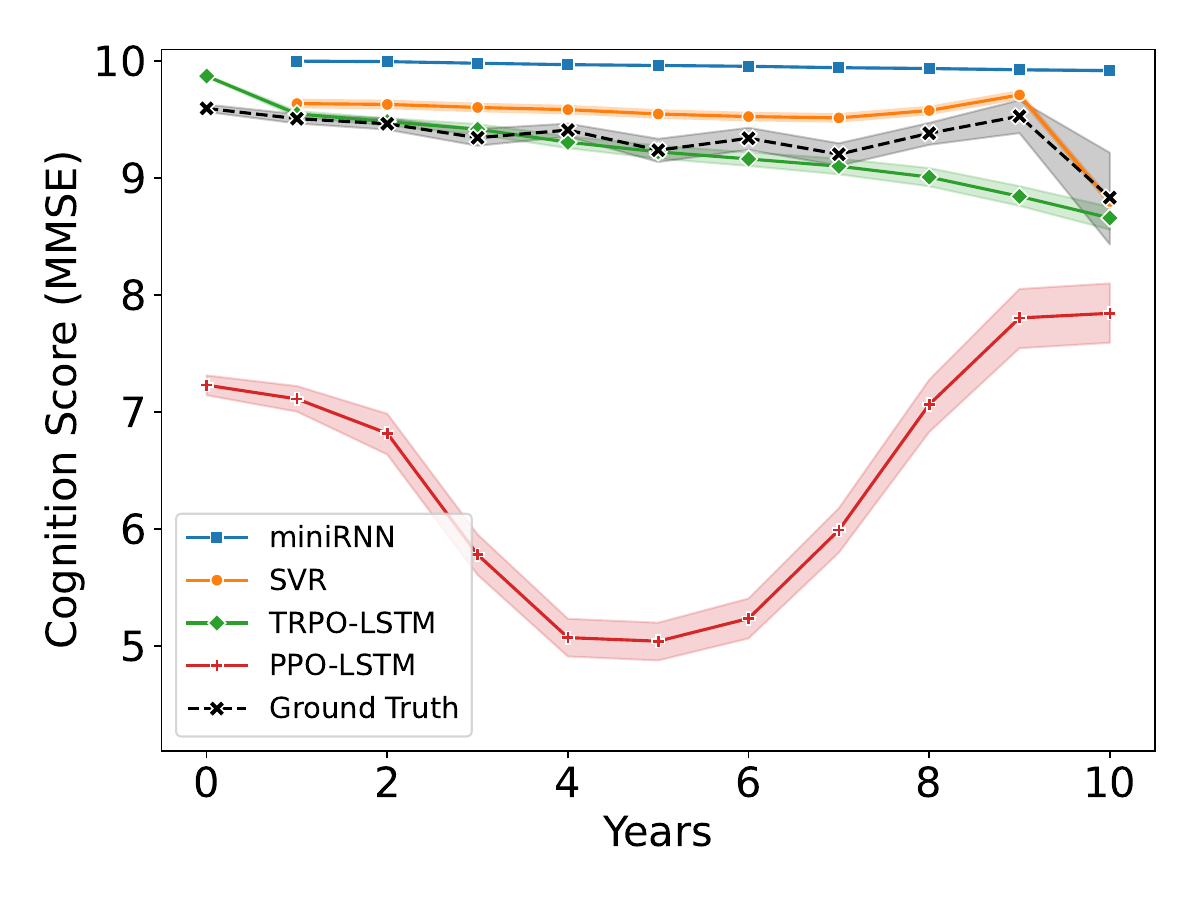}
        \caption{Recurrent baselines}
        \label{fig:lstm_baselines}
    \end{subfigure}
    \begin{subfigure}{0.32\linewidth}
        \centering
        \includegraphics[width=\linewidth]{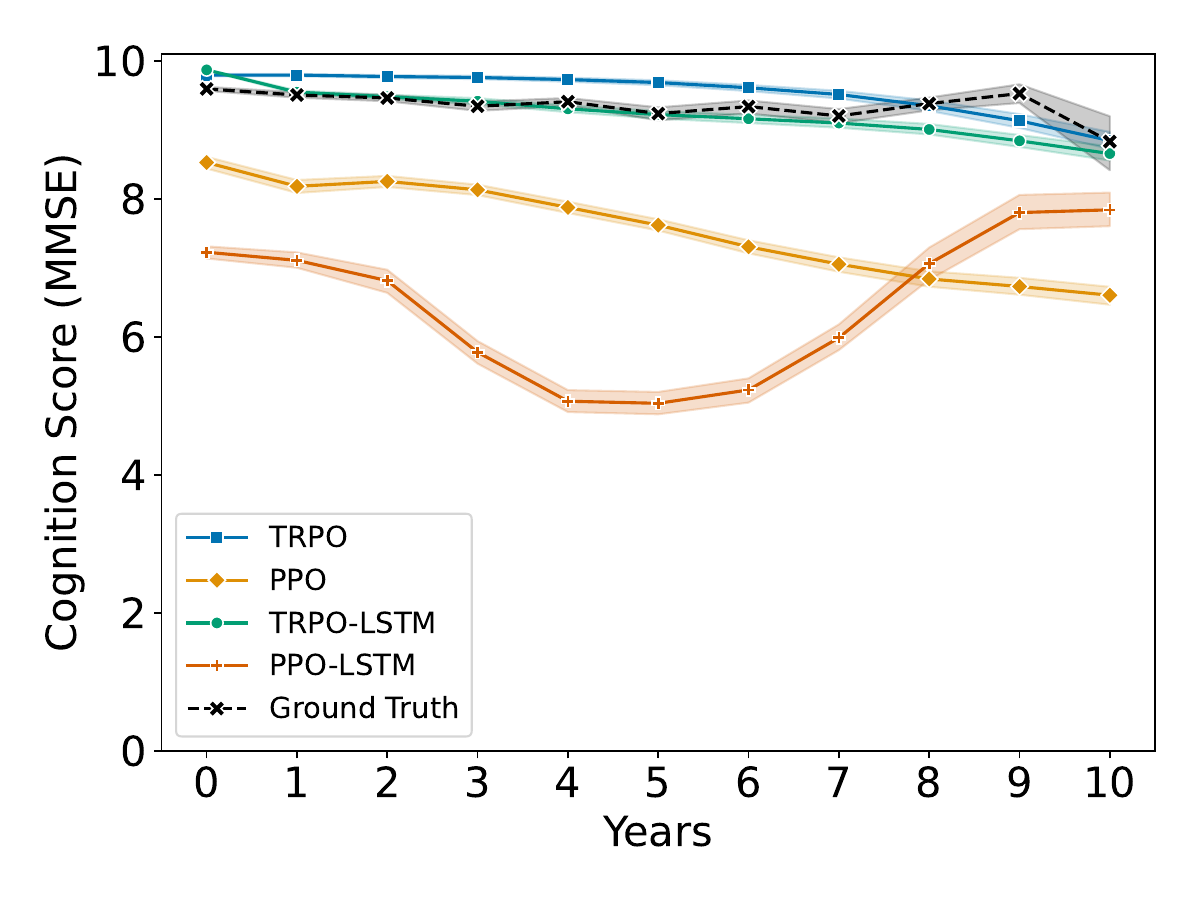}
        \caption{Memory Augmented RL}
        \label{fig:lstm_rl}
    \end{subfigure}
    \begin{subfigure}{0.32\textwidth}
        \includegraphics[width=\linewidth]{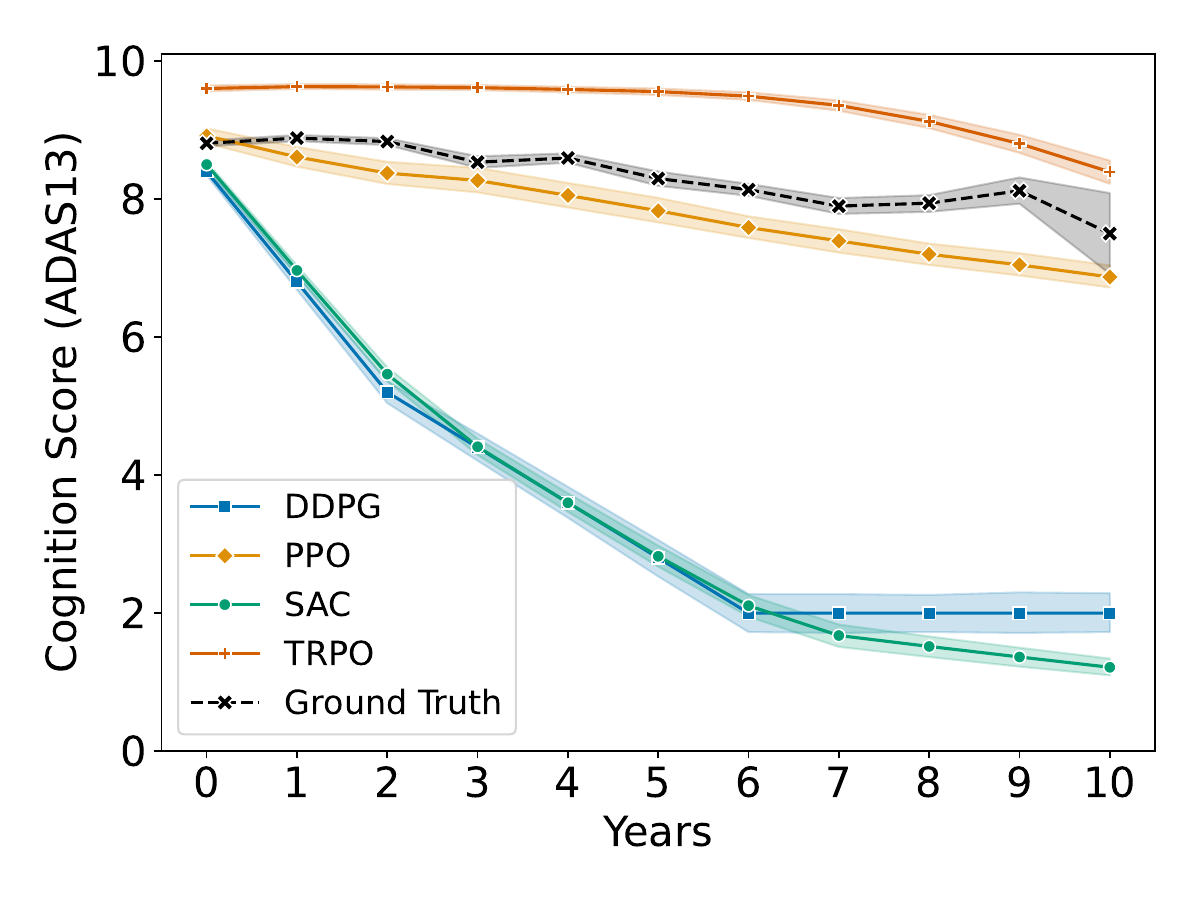}
        \caption{ADAS13 Predictions}
        \label{fig:adas13}
    \end{subfigure}
    
    \caption{\textbf{a}: AD progression prediction of TRPO-LSTM and PPO-LSTM vs Recurrent baselines (miniRNN and SVR). \textbf{b}: AD progression prediction of TRPO and PPO methods vs their recurrent variants. \textbf{c}: Predicted Cognition Trajectories using the ADAS13 score.}
    \label{fig:lstms_adas13}
\end{figure*}

Figure \ref{fig:lstm_baselines} shows the results of the TRPO-LSTM and PPO-LSTM variants, plotted against supervised baselines miniRNN and Suppor Vector Regression (SVR) \cite{nguyen2020predicting}, two state of the art recurrent neural network approaches on the AD prediction tasks. Figure \ref{fig:lstm_rl} shows the results of the TRPO-LSTM and PPO-LSTM variants plotted alongside their non-recurrent versions. Results are presented in Table \ref{tab:lstms}. While TRPO-LSTM achieves performance similar to TRPO in terms of progression prediction, it has a higher MAE. PPO-LSTM on the other hand is unable to predict the cognition trajectory and performs worse than PPO. This relatively poor performance of recurrent model-free RL methods may be attributed to inductive biases encoded within these memory architectures and may require more careful hyperparameter tuning to perform (at least) on par with their non-recurrent counterparts \citep{ni2022recurrent}.

\begin{table}[ht!]
\centering
\setlength{\tabcolsep}{30pt}
{
    \begin{tabular}{c c c} \toprule
    Method  & MAE  & MSE \\
    \midrule
    miniRNN &  0.599 (0.137)  & 0.984 (0.659)  \\
    SVR & \textbf{0.495 (0.067)} & \textbf{0.574 (0.230)}  \\
    TRPO-LSTM & 0.654 (0.127)  & 0.883 (0.283)  \\
    PPO-LSTM  & 3.281 (1.06) & 16.427 (10.107)  \\
    \bottomrule
\end{tabular}}
\caption{AD progression prediction performance of memory augmented methods (RL+LSTM) vs recurrent network baselines. MAE: mean absolute error, MSE: mean squared error}
\label{tab:lstms}
\end{table}

\section{Additional Cognition Tests: ADAS-Cog13}
\label{app:adas13}

There are several cognitive tests commonly used in clinical practice and research settings to assess cognitive function and screen for dementia. The Alzheimer's Disease Assessment Scale-Cognitive subscale (ADAS-Cog 13 or ADAS13) \cite{mohs1997development} and the Mini-Mental State Examination (MMSE) \cite{folstein1975mini} are two widely used cognitive assessment tools in Alzheimer's disease research and clinical practice. While both instruments aim to evaluate cognitive function and screen for dementia, they differ in their focus, structure, and sensitivity. The ADAS13 is an extension of the original ADAS-Cog 11 \cite{rosen1984new}, which includes 11 tasks assessing memory, language, praxis, and orientation. The ADAS-Cog 13 adds two additional components to evaluate delayed word recall and number cancellation tasks, providing a broader assessment of cognitive functions affected by AD. In contrast, MMSE primarily assesses global cognitive function, focusing on areas such as orientation, memory, attention, and language.

\begin{figure}[t!]
    \centering
    \begin{subfigure}{\textwidth}
        \includegraphics[width=\linewidth]{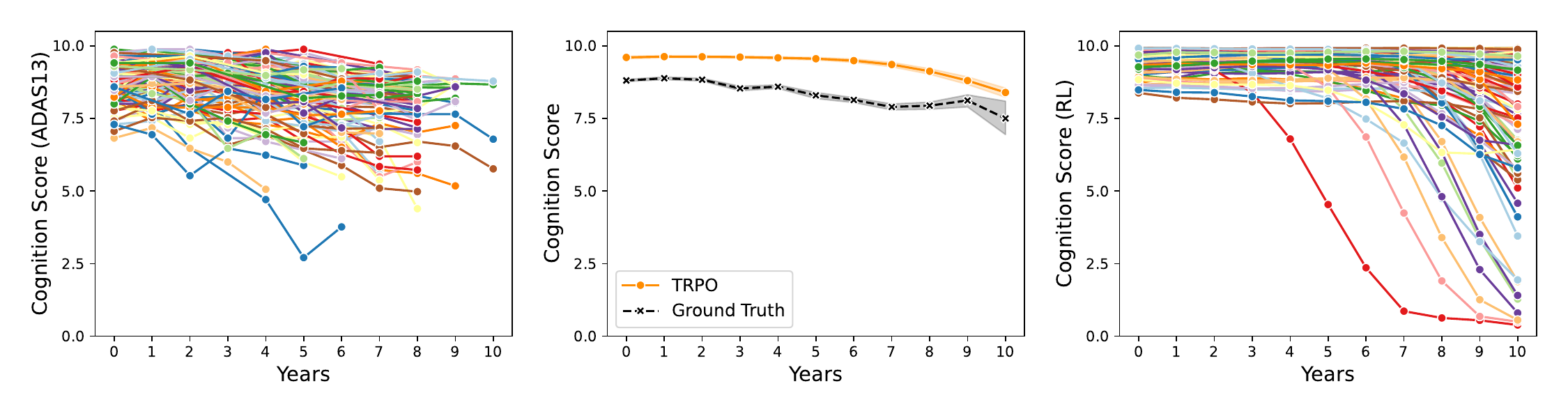}
    \end{subfigure}
    
    \caption{Individual ground truth vs TRPO predicted cognition trajectories when ADAS13 is used as the ground truth cognition score. \textbf{Left}: Ground truth values for all patients. \textbf{Center}: Mean values of cognition for ground truth and TRPO predicted cognition trajectories. \textbf{Right}: TRPO predicted individual cognition trajectories for all samples (mean over 5 seeds).}
    \label{fig:adas13_individual}
\end{figure}

We experimented with using ADAS13 as the underlying cognition ground truth to predict $C(t)$. Figure \ref{fig:adas13} shows a comparison of the four RL methods when tested on the ADAS-Cog 13 score, whereas Figure \ref{fig:adas13_individual} shows the individual cognition trajectories for ground truth and TRPO predicted values. It can be observed that while TRPO performs relatively best among the four methods on ADAS13 as it did on MMSE, its prediction curve remains the same although the ground truth values for ADAS13 are lower than MMSE. This shows that the AD model described in this paper may be limited in its accuracy to predict the MMSE score only, which highlights a crucial flaw in the model. This also raises the question as to how RL-based frameworks can be adapted to suit different types of cognition tests, of which many exist.

\section{Recovery Compensatory Effects in Brain}
\label{app:rl_trajectory}

Recovery compensatory effects in the brain involve complex mechanisms that allow the brain to adapt and compensate for lost functions due to injury or disease \cite{hylin2017understanding}. These mechanisms include neural plasticity, the brain's ability to reorganize itself by forming new neural connections, and the development of compensatory behaviors, where remaining functional areas of the brain take over the functions of the damaged areas. This process can be influenced by factors such as the timing and intensity of rehabilitation, to maximize functional recovery and improve quality of life for individuals affected by brain injuries or neurodegenerative diseases.

\begin{figure}[th!]
    \centering
    \begin{subfigure}{0.24\textwidth}
        \includegraphics[width=\linewidth]{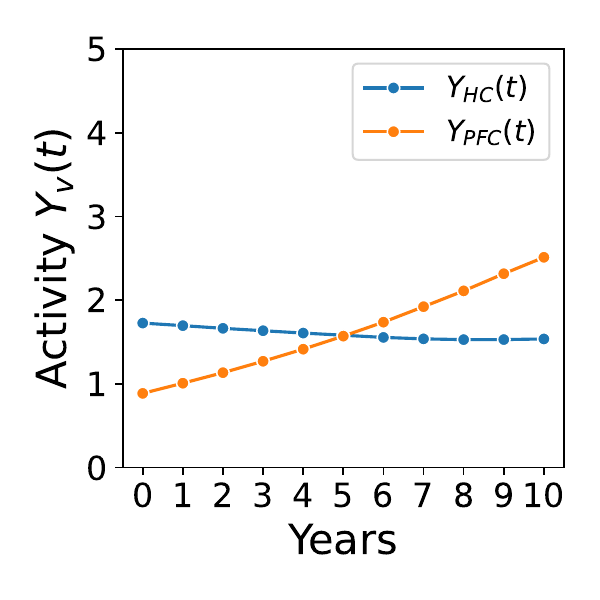}
    \end{subfigure}%
    \hfill
    \begin{subfigure}{0.24\textwidth}
        \includegraphics[width=\linewidth]{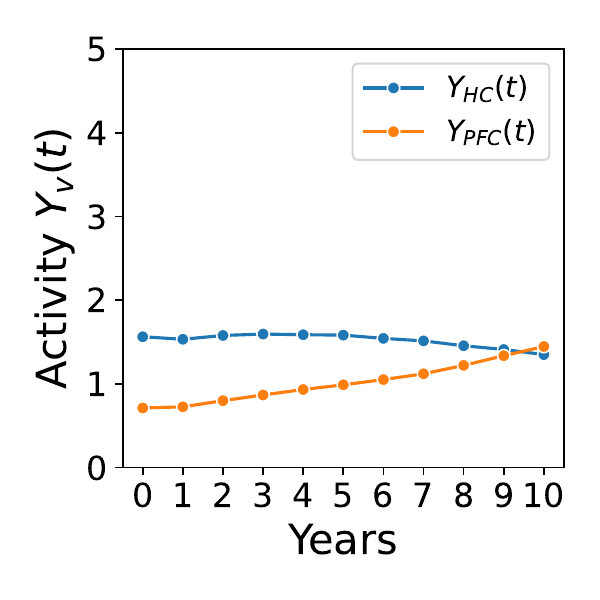}
    \end{subfigure}%
    \hfill
    \begin{subfigure}{0.24\textwidth}
        \includegraphics[width=\linewidth]{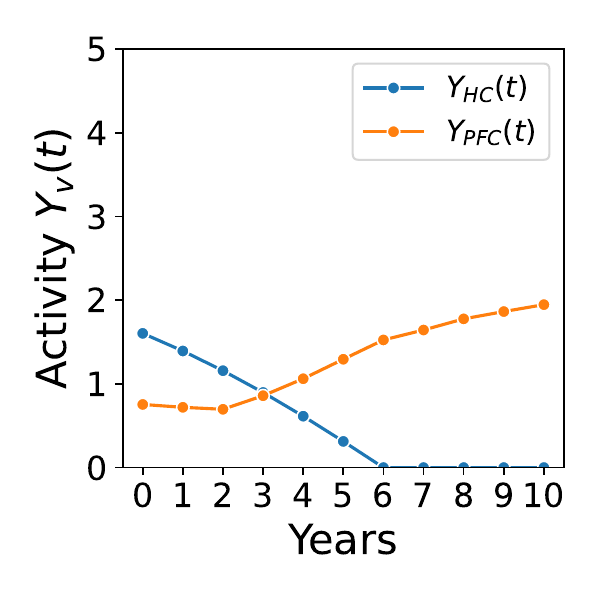}
    \end{subfigure}%
    \hfill
    \begin{subfigure}{0.24\textwidth}
        \includegraphics[width=\linewidth]{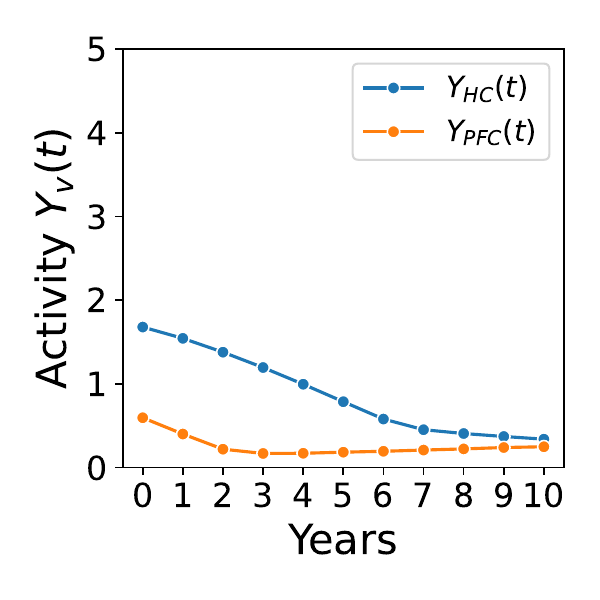}
    \end{subfigure}\\
    
    \begin{subfigure}{0.24\textwidth}
        \includegraphics[width=\linewidth]{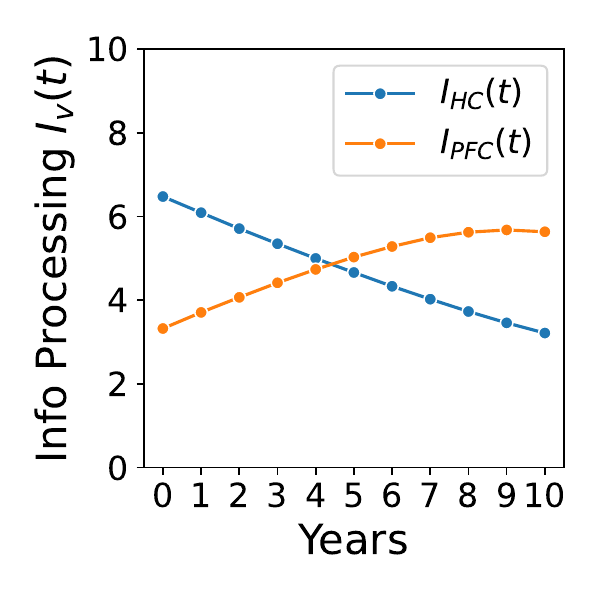}
        \caption{TRPO}
    \end{subfigure}%
    \hfill
    \begin{subfigure}{0.24\textwidth}
        \includegraphics[width=\linewidth]{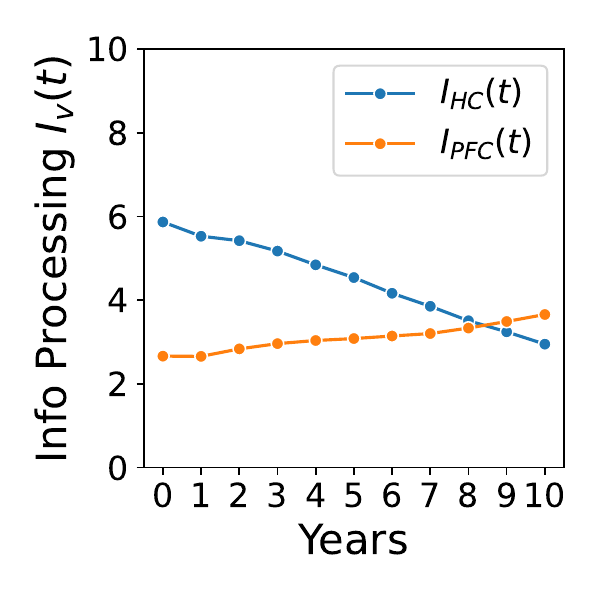}
        \caption{PPO}
    \end{subfigure}%
    \hfill
    \begin{subfigure}{0.24\textwidth}
        \includegraphics[width=\linewidth]{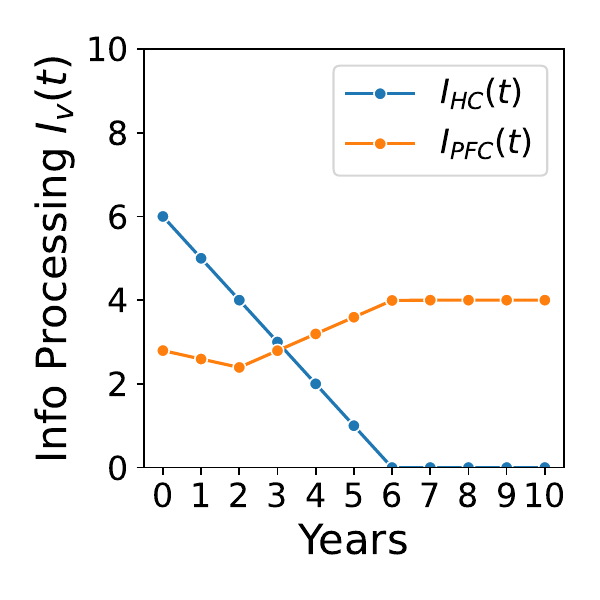}
        \caption{DDPG}
    \end{subfigure}%
    \hfill
    \begin{subfigure}{0.24\textwidth}
        \includegraphics[width=\linewidth]{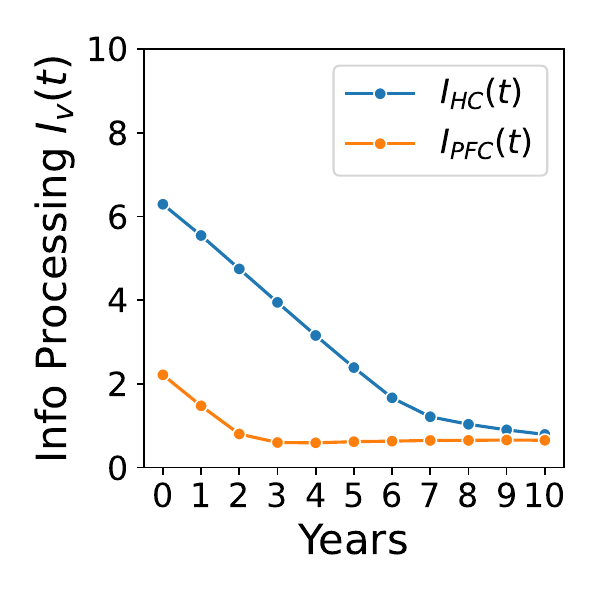}
        \caption{SAC}
    \end{subfigure}\\
    
    \caption{Recovery/compensatory effect for each of the four RL methods. \textbf{Top Row} Brain Activity for each of the two regions. \textbf{Bottom Row} Information Processing (contribution to overall cognition) by each brain region.}
    \label{fig:recovery_compensatory}
\end{figure}

Figure \ref{fig:recovery_compensatory} illustrates the recovery compensatory effect in brain activity and information processing for the four RL methods. The top row depicts brain activity levels for two key regions, showing how each RL method predicts changes over ten years. It can be seen that only TRPO is able to correctly model recovery compensatory effects, whereas none of the other algorithms can do so. This might be the key reason that TRPO can predict cognition trajectories well, although the underlying objective function remains the same. The bottom row details the information processing for each brain region, representing their contribution to overall cognition. These processes highlight the comparative dynamics and recovery patterns captured by each RL method, providing information in support of their predictive performance over time.

\section{Implementation}
\label{app:implementation}

\paragraph{Experiment Design.}
\label{app:experiment_design}
The policy network for each algorithm consists of a two-layer feedforward neural network with 32 hidden units per layer. We train all four methods using the Adam optimizer and implementations made available by the Garage library \cite{contributors2019garage}, using default hyperparameters. We perform k-fold cross-validation with k=5 and repeat each fold 5 times using a different random seed value (25 experiments per method). The data is split in 64:16:20 (train, validation, test). We train each agent for 1 million timesteps on the train split, where each epoch involves sampling 1000 trajectories. Each patient's trajectory comprises 11 time points, including the baseline (8800 data points generated during evaluation per algorithm). To ensure stable learning, we clip the reward within a range of [-2000, 2000] and limit the action space to [-2,2], reflecting MMSE score changes observed in ADNI data over consecutive years. The LSTM variants of TRPO and PPO had a recurrent layer with 32 hidden units. Experiments were run on a 32-core, 128GB RAM machine, and models were trained without using a GPU.

\paragraph{Data Processing.}
\label{app:data_processing}
Data used in preparation of this article were obtained from the \cite{adni} database. Since disease progression modeling required the availability of longitudinal data, patients were filtered based on baseline measurements of cognition, demographics, MRI, and florbetapir-PET scans, along with longitudinal cognitive measurements and at least 2 follow-up assessments comprising both PET and MRI scans. Follow-up visits were not required to be consecutive, and data spanning up to 10 years after baseline were retained. Cognitive assessments were preserved for all available time points up to and including year 10, irrespective of MRI/PET availability. This resulted in a dataset of 160 patients, encompassing 52 cognitively normal (CN), 23 with significant memory concern (SMC), 58 with early mild cognitive impairment (EMCI), and 27 with late mild cognitive impairment (LMCI). Demographic features included age, gender, education, and the presence of the APOE-$\epsilon$4 genotype. Our analysis focused on a 2-node graph representation with nodes denoting the hippocampus (HC) and the prefrontal cortex (PFC) due to their relevance to cognition and Alzheimer's disease pathology. Hippocampal and prefrontal cortex volumes were used to represent brain structure $X(t)$. PET-scan derived Standardized Uptake Value Ratio (SUVR) values for PFC and HC served as measures of A$\beta$ deposition $D(t)$. We used the Mini-Mental State Examination (MMSE) score \cite{folstein1975mini} as a measure of cognition $C(t)$ for most of our experiments (unless stated otherwise), normalizing values to be in the range 0 to 10. 

\paragraph{Hyperparameters.}
\label{app:hyperparams}
We provide the hyperparameters for our experiments in Table \ref{tab:hyperparameters}. These values are used in all our experiments unless specified otherwise.

\begin{table}[ht!]
\centering
\setlength{\tabcolsep}{30pt}
{
\begin{tabular}{c c}
    \toprule
    \bf Hyperparameters & \bf Values \\
    \hline
    Batch size & 1000 \\
    Epochs & 1000 \\
    {GAE $\lambda$} & {0.97} \\
    Learning Rate Clip Range & $0.2$ \\ 
    Policy Entropy Coefficient & $0.02$ \\
    Max Cognition ($C_{task}$)  & 10.0 \\
    Scale Observations & True \\
    Action limit & $\pm$ 2.0 \\
    Score & MMSE (default), ADAS13  \\
    Max timesteps & 11 (years) \\
    Number of seeds & 5 \\
    Hidden Units (TRPO-LSTM, PPO-LSTM) & 32 \\
    Experience Replay Size (DDPG, SAC) & 1M \\ 
    \bottomrule
\end{tabular}}
\caption{Hyperparameters for our experiments}
\label{tab:hyperparameters}
\end{table}

\section{Ground Truth vs RL Prediction for all variables}
\label{app:comparison_gt_rl}

\begin{figure}[ht!]
    \captionsetup{font={footnotesize}} 
    
    \centering
    \begin{subfigure}{0.19\textwidth}
        \includegraphics[width=\linewidth]{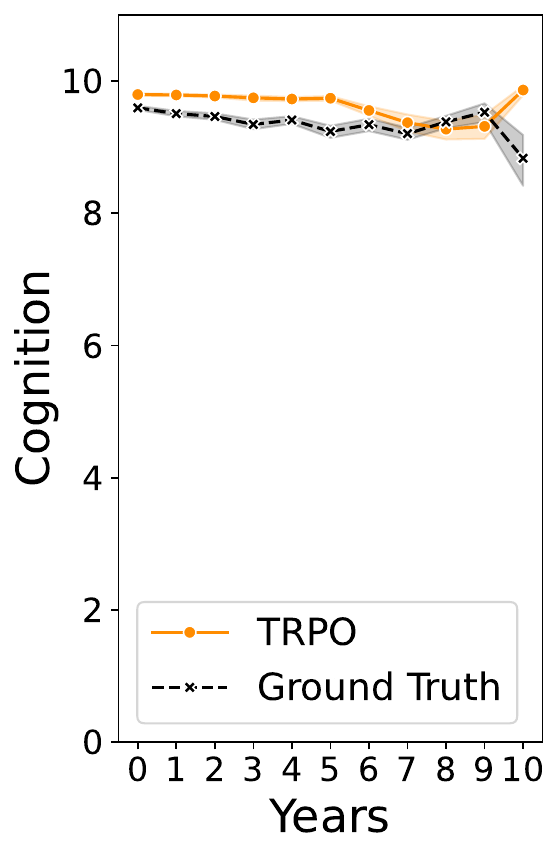}
    \end{subfigure}%
    \begin{subfigure}{0.19\textwidth}
        \includegraphics[width=\linewidth]{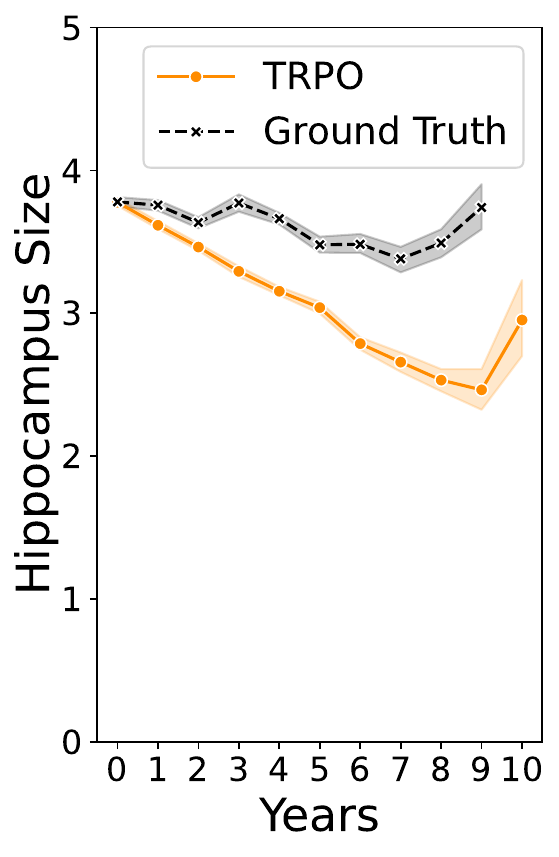}
    \end{subfigure}%
    \begin{subfigure}{0.19\textwidth}
        \includegraphics[width=\linewidth]{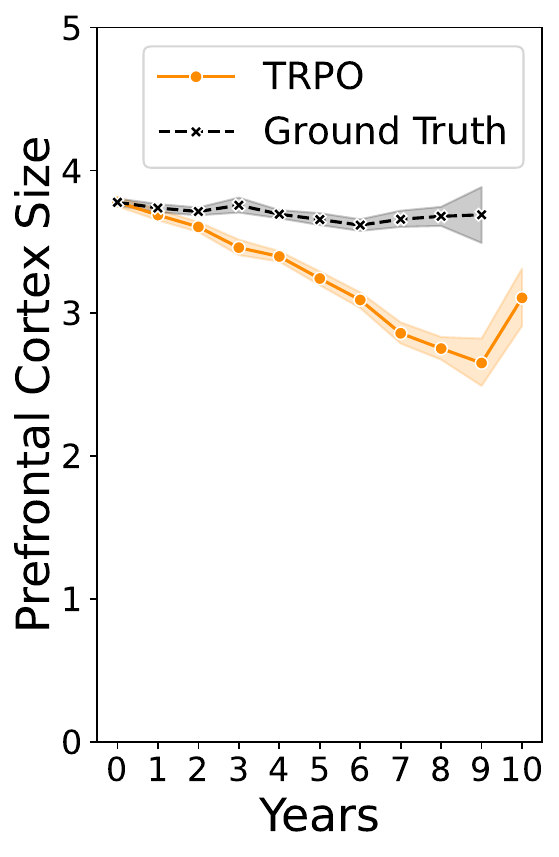}
    \end{subfigure}%
    \begin{subfigure}{0.19\textwidth}
        \includegraphics[width=\linewidth]{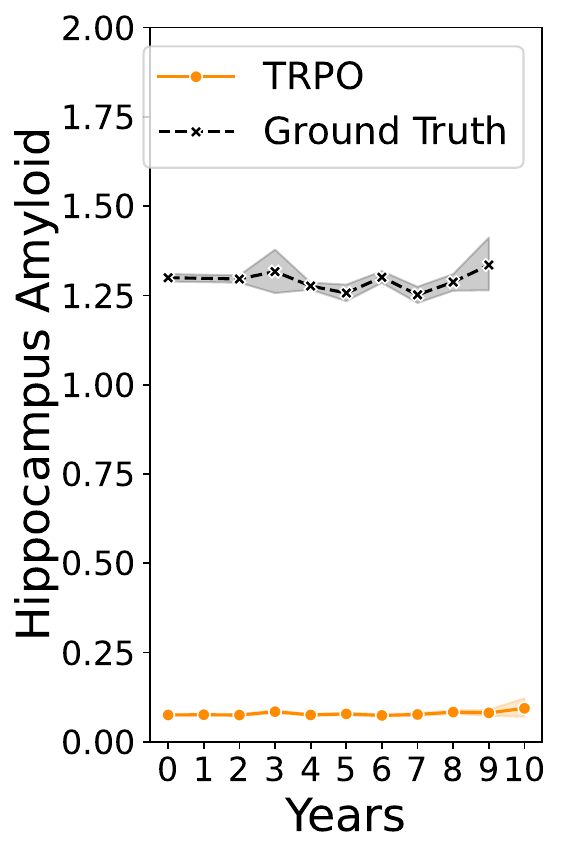}
    \end{subfigure}%
    \begin{subfigure}{0.19\textwidth}
        \includegraphics[width=\linewidth]{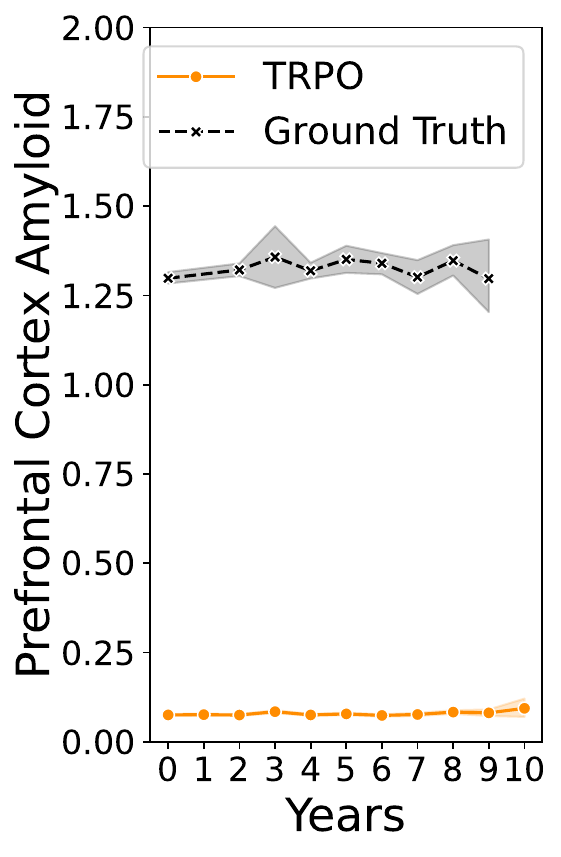}
    \end{subfigure}\\
    
    \begin{subfigure}{0.19\textwidth}
        \includegraphics[width=\linewidth]{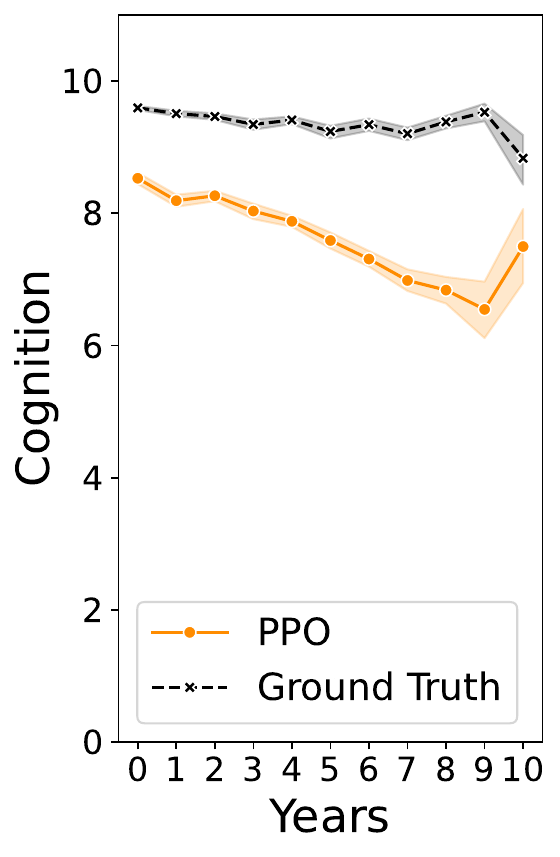}
    \end{subfigure}
    \begin{subfigure}{0.19\textwidth}
        \includegraphics[width=\linewidth]{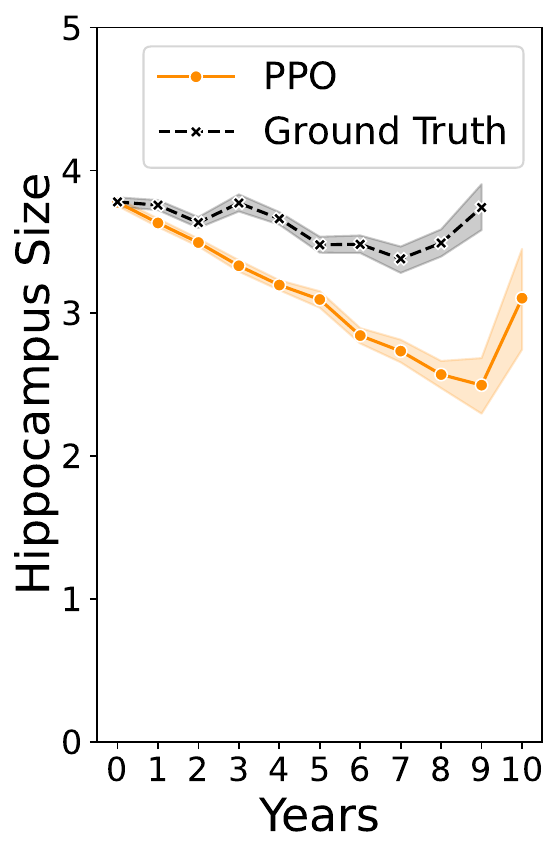}
    \end{subfigure}%
    \begin{subfigure}{0.19\textwidth}
        \includegraphics[width=\linewidth]{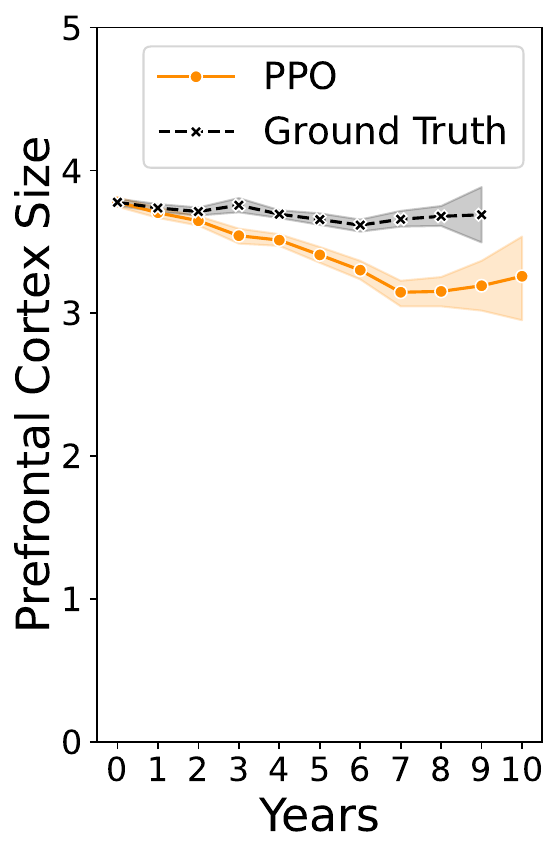}
    \end{subfigure}%
    \begin{subfigure}{0.19\textwidth}
        \includegraphics[width=\linewidth]{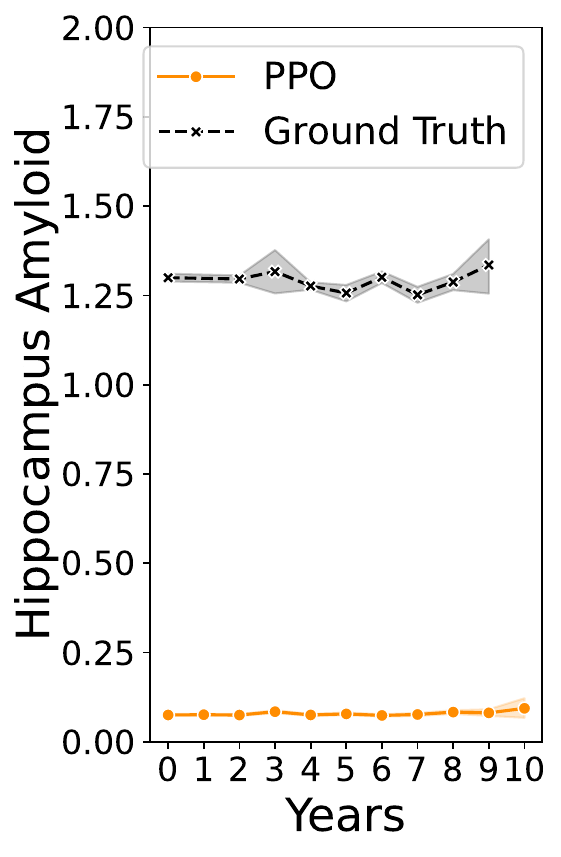}
    \end{subfigure}%
    \begin{subfigure}{0.19\textwidth}
        \includegraphics[width=\linewidth]{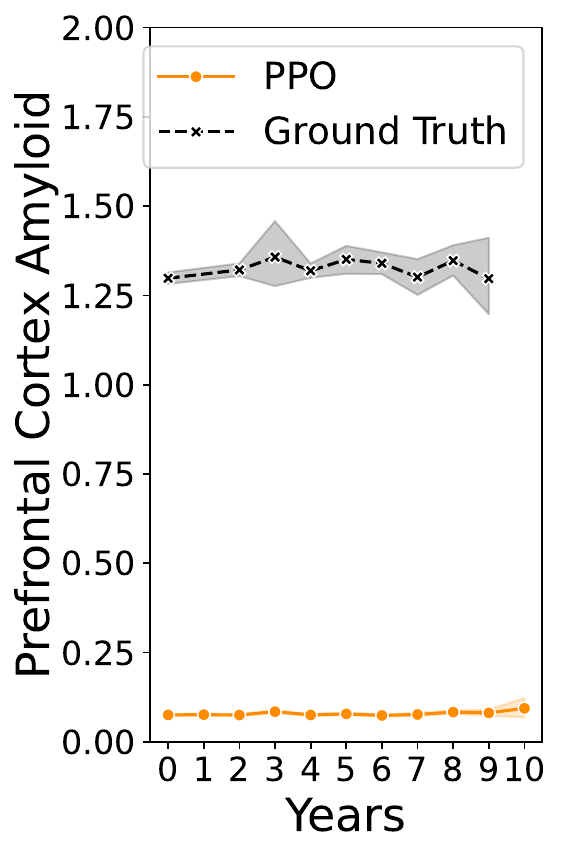}
    \end{subfigure}\\

    \begin{subfigure}{0.19\textwidth}
        \includegraphics[width=\linewidth]{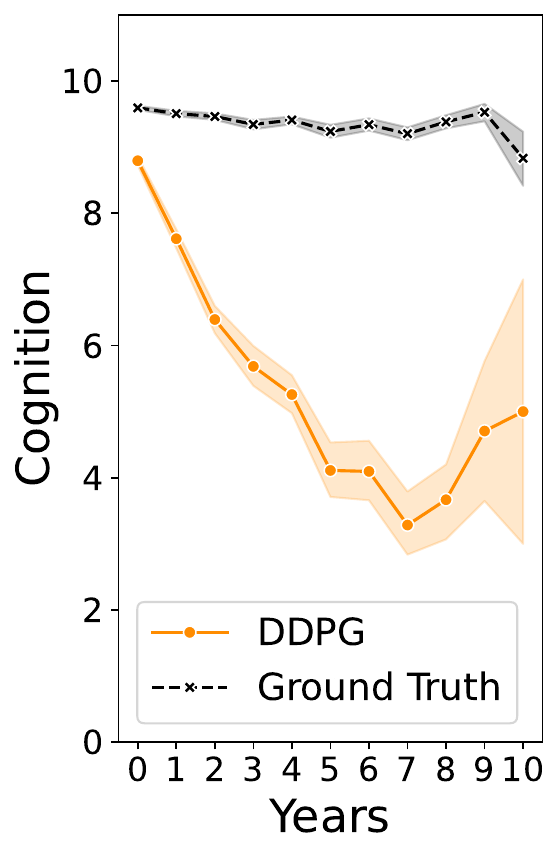}
    \end{subfigure}%
    \begin{subfigure}{0.19\textwidth}
        \includegraphics[width=\linewidth]{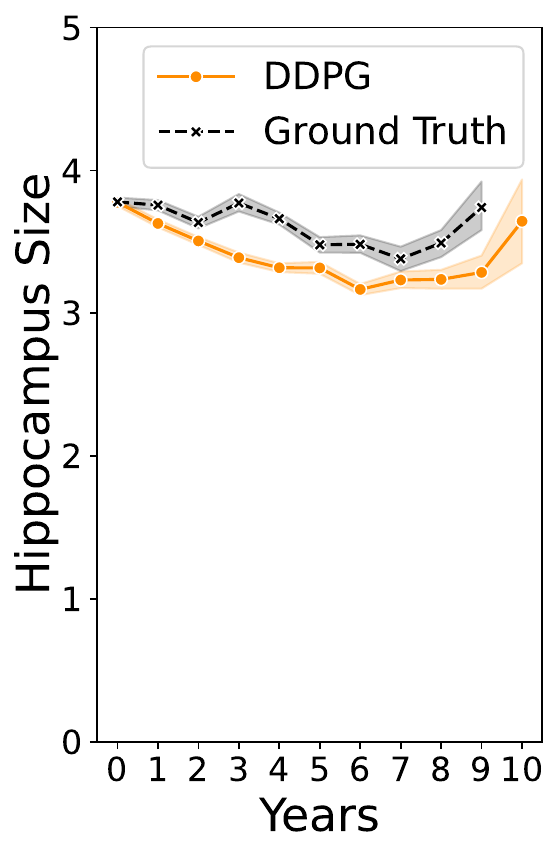}
    \end{subfigure}%
    \begin{subfigure}{0.19\textwidth}
        \includegraphics[width=\linewidth]{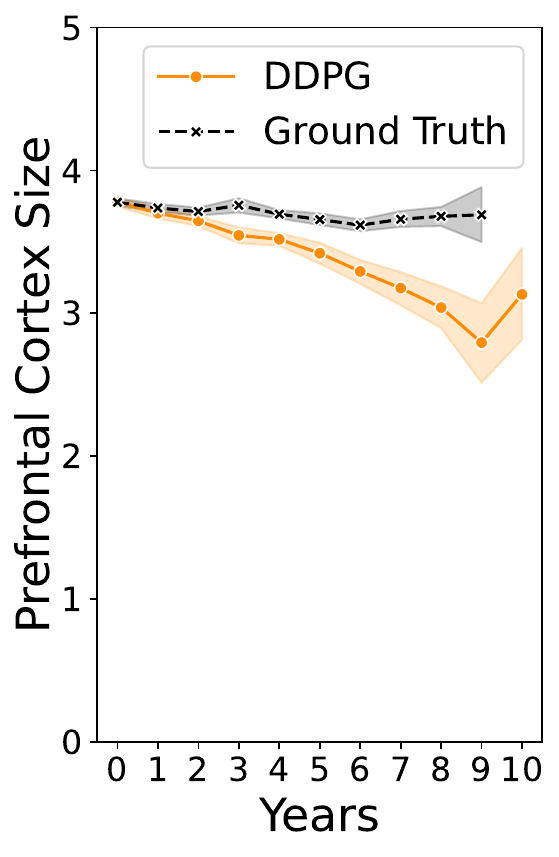}
    \end{subfigure}%
    \begin{subfigure}{0.19\textwidth}
        \includegraphics[width=\linewidth]{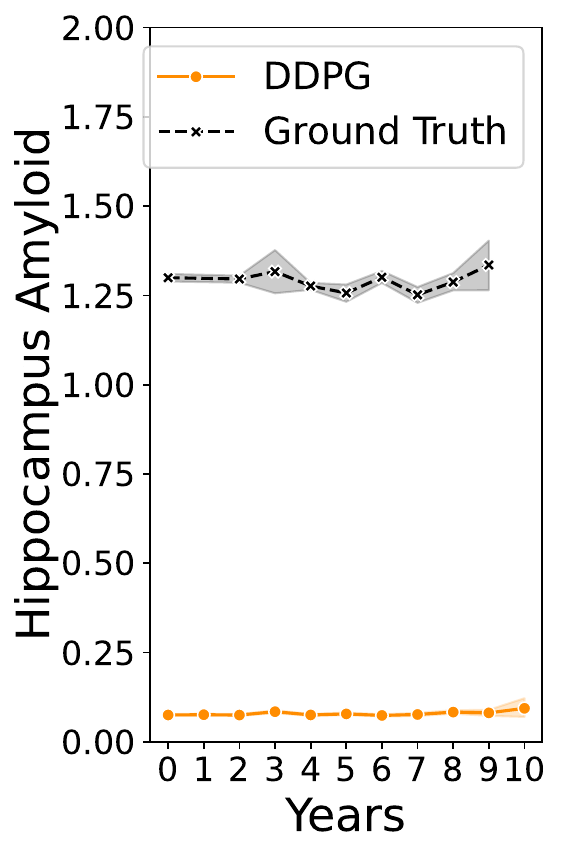}
    \end{subfigure}%
    \begin{subfigure}{0.19\textwidth}
        \includegraphics[width=\linewidth]{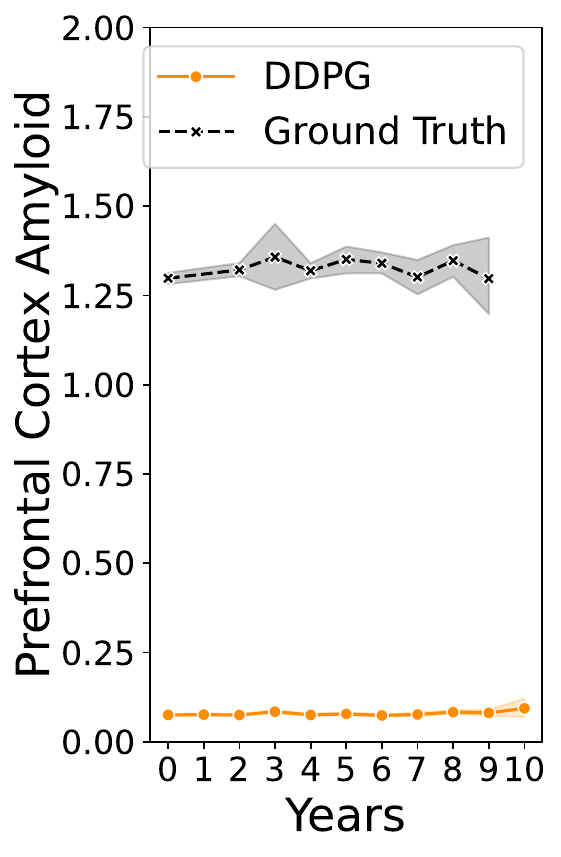}
    \end{subfigure}\\

    \begin{subfigure}{0.19\textwidth}
        \includegraphics[width=\linewidth]{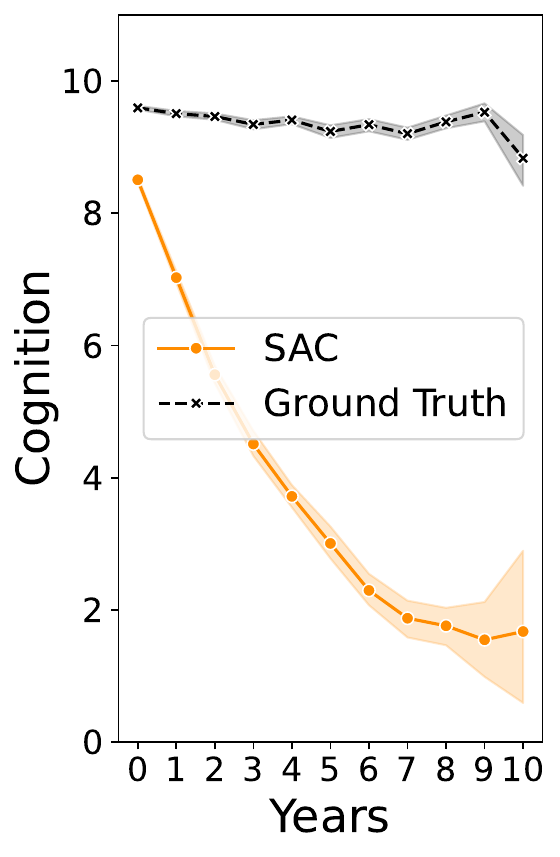}
        \caption{Cognition}
    \end{subfigure}%
    \begin{subfigure}{0.19\textwidth}
        \includegraphics[width=\linewidth]{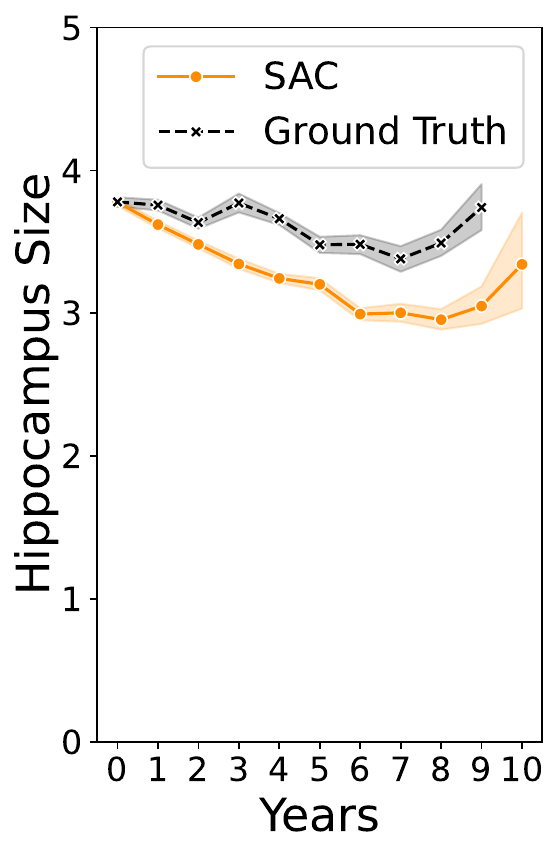}
        \caption{HC Size}
    \end{subfigure}%
    \begin{subfigure}{0.19\textwidth}
        \includegraphics[width=\linewidth]{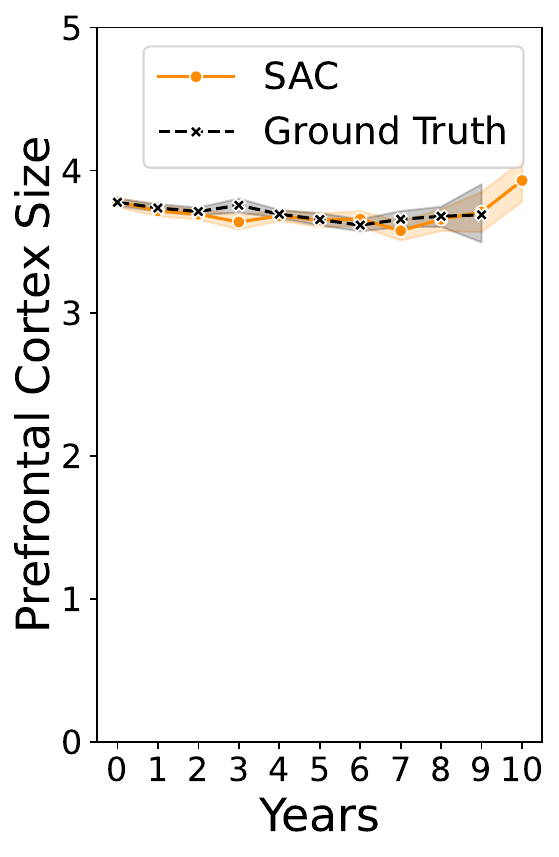}
        \caption{PFC Size}
    \end{subfigure}%
    \begin{subfigure}{0.19\textwidth}
        \includegraphics[width=\linewidth]{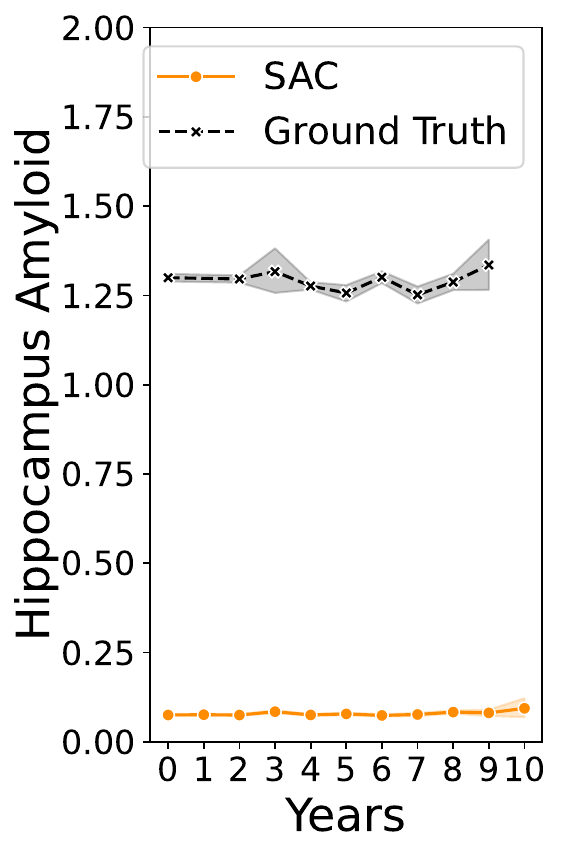}
        \caption{HC Amyloid}
    \end{subfigure}%
    \begin{subfigure}{0.19\textwidth}
        \includegraphics[width=\linewidth]{plots/supp_rl_comparison/DDPG_MMSE_comp_amyloid-pfc.pdf}
        \caption{PFC Amyloid}
    \end{subfigure}\\
    
    \captionsetup{font={}} 
    \caption{Ground truth vs predictions of each RL algorithm for the five variables modeled through the simulator. From Left to Right: Cognition, Hippocampus (HC) Size/Volume, Prefrontal Cortex (PFC) Size/Volume, Hippocampus Amyloid and Prefrontal Cortex Amyloid.}
    \label{fig:comparison_all_5}
\end{figure}

Figure \ref{fig:comparison_all_5} presents a comparative analysis of ground truth versus predictions made by the four RL algorithms—TRPO, PPO, DDPG, and SAC—over a decade-long period, using five key variables from the dataset processed through a simulator. Each plot illustrates the trajectory of predictions against actual data points for cognition, hippocampus size, prefrontal cortex size, hippocampus amyloid accumulation, and prefrontal cortex amyloid accumulation, offering insights into each algorithm's predictive accuracy and temporal consistency with the real-world progression of the measured variables. We can observe that prediction accuracy varies with each RL method. While TRPO can model cognition trajectories accurately, DDPG and SAC are better at predicting brain region size. However, none of the RL methods can correctly predict the amyloid deposition, which necessitates additional measures to better capture amyloid's modeling in the differential equations.  

\clearpage

\section{Individual Cognition Trajectories}
\label{app:cog_trajectory}

\begin{figure}[ht!]
    \centering
    \begin{subfigure}{0.9\textwidth}
        \includegraphics[width=\linewidth]{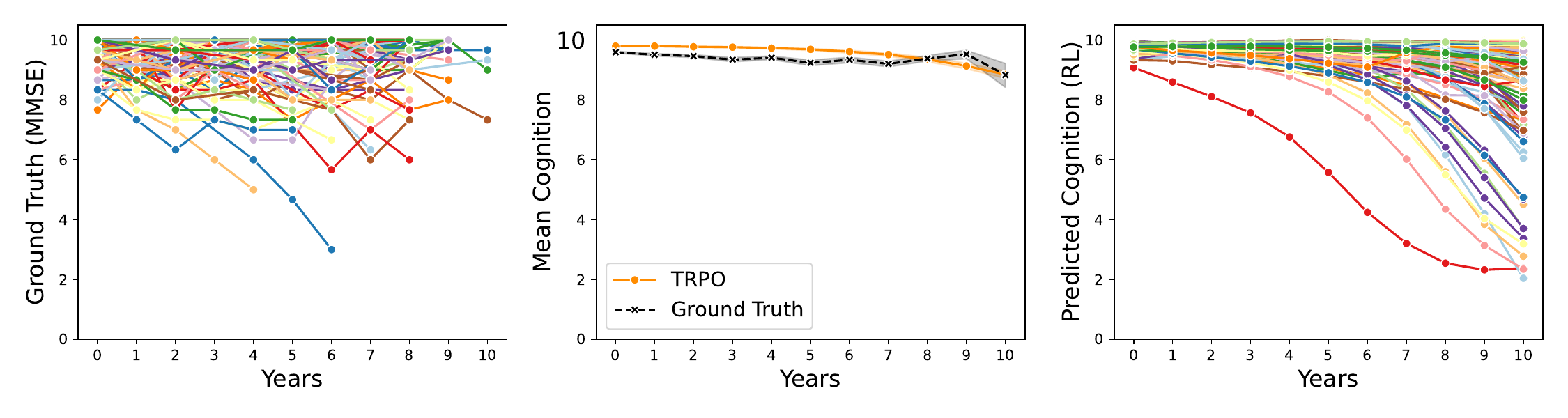}
    \end{subfigure}\\
    
    \begin{subfigure}{0.9\textwidth}
        \includegraphics[width=\linewidth]{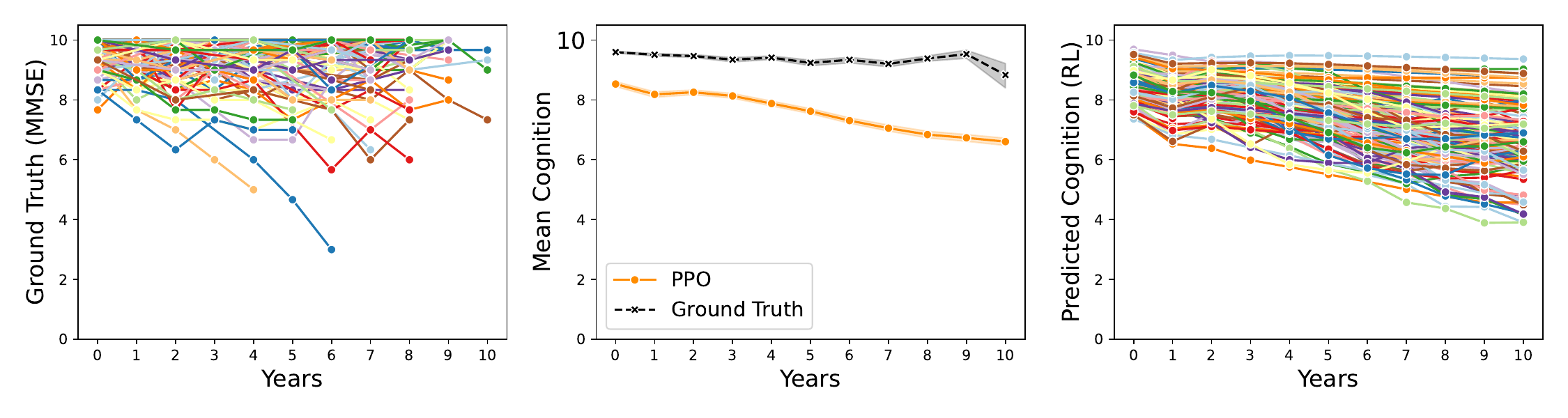}
    \end{subfigure}\\
    
    \begin{subfigure}{0.9\textwidth}
        \includegraphics[width=\linewidth]{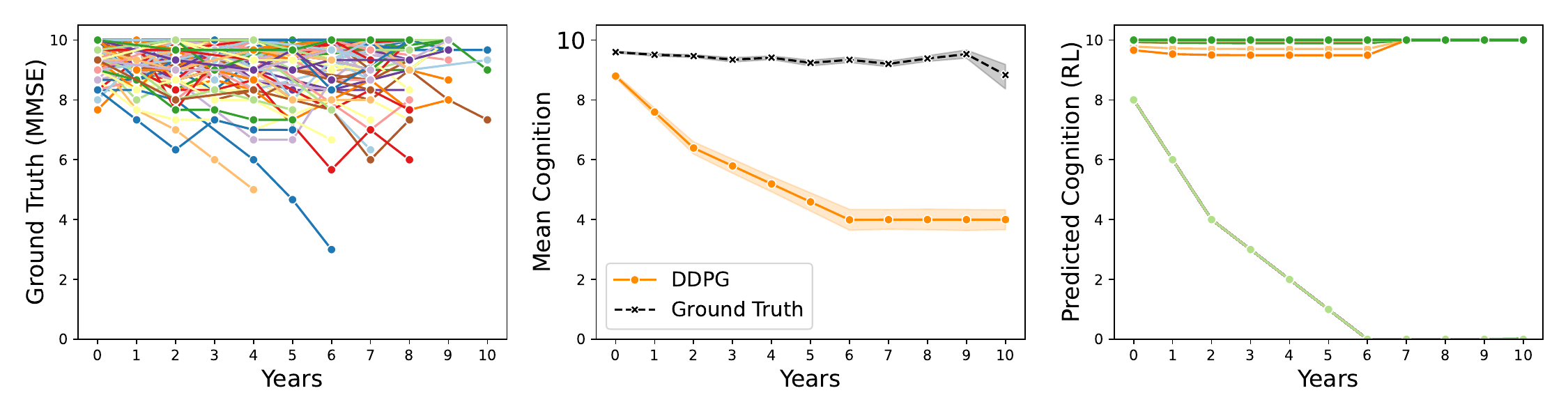}
    \end{subfigure}\\
    
    \begin{subfigure}{0.9\textwidth}
        \includegraphics[width=\linewidth]{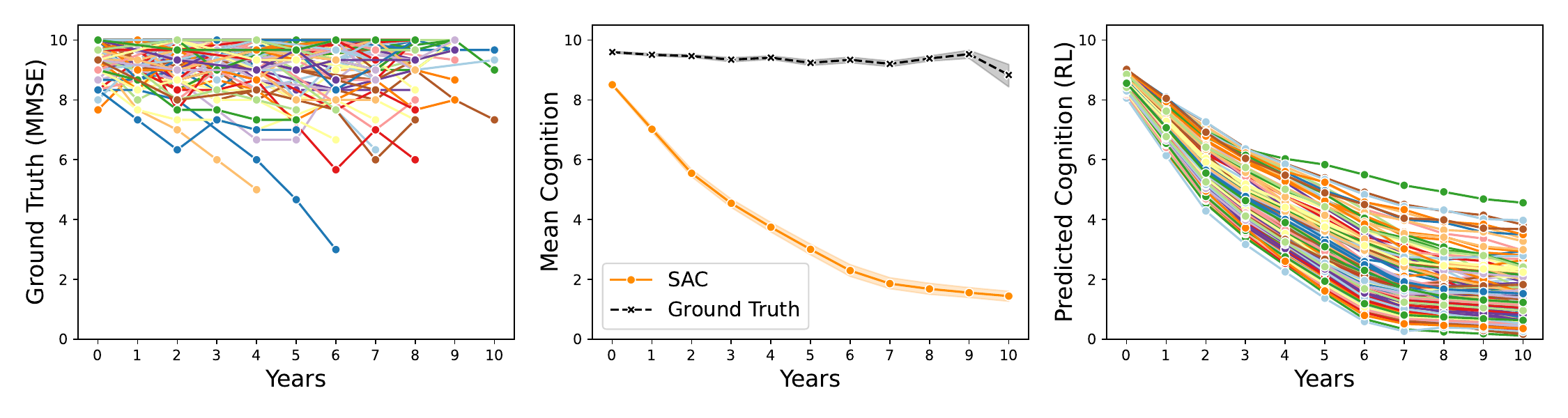}
    \end{subfigure}\\
    
    \caption{Cognition Trajectories for each of the four RL methods. Top to Bottom: TRPO, PPO, DDPG, SAC. \textbf{Left}: Ground truth values for all patients, \textbf{Center}: Mean values of cognition for ground truth and RL method's prediction. \textbf{Right}: RL method's predicted individual cognition trajectories for all samples (mean over 5 seeds).}
    \label{fig:cog_traj_all_3in1}
\end{figure}

Figure \ref{fig:cog_traj_all_3in1} shows the individual cognition trajectories for ground truth and RL predicted values, for each of the four RL algorithms studied in this work.

\section{Comparison of RL Algorithms' Explanations}
\label{app:shap_global_algos}

\begin{figure}[ht!]
    \centering
    \begin{subfigure}{0.3\textwidth}
        \includegraphics[width=\linewidth]{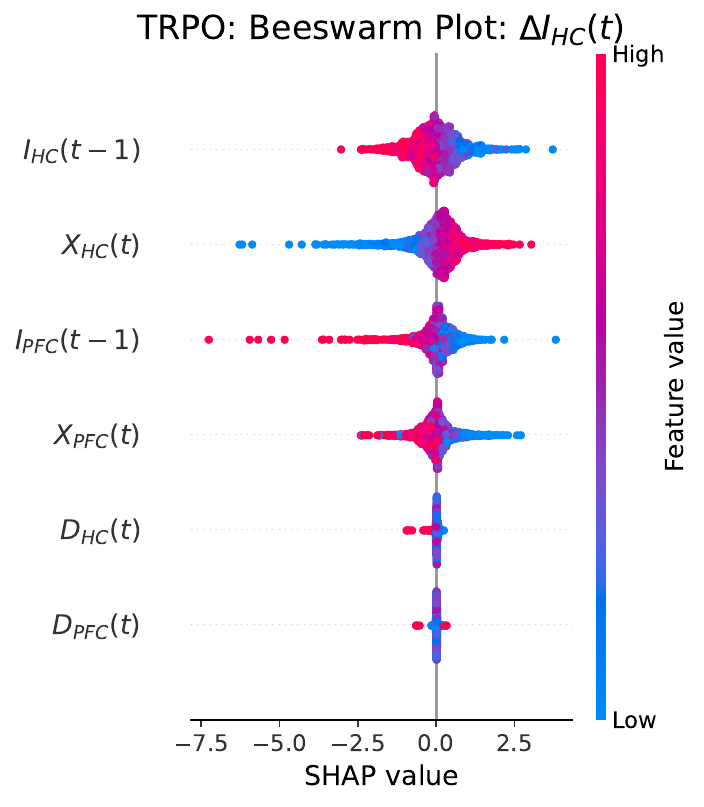}
    \end{subfigure}%
    \hspace{2cm}
    \begin{subfigure}{0.3\textwidth}
        \includegraphics[width=\linewidth]{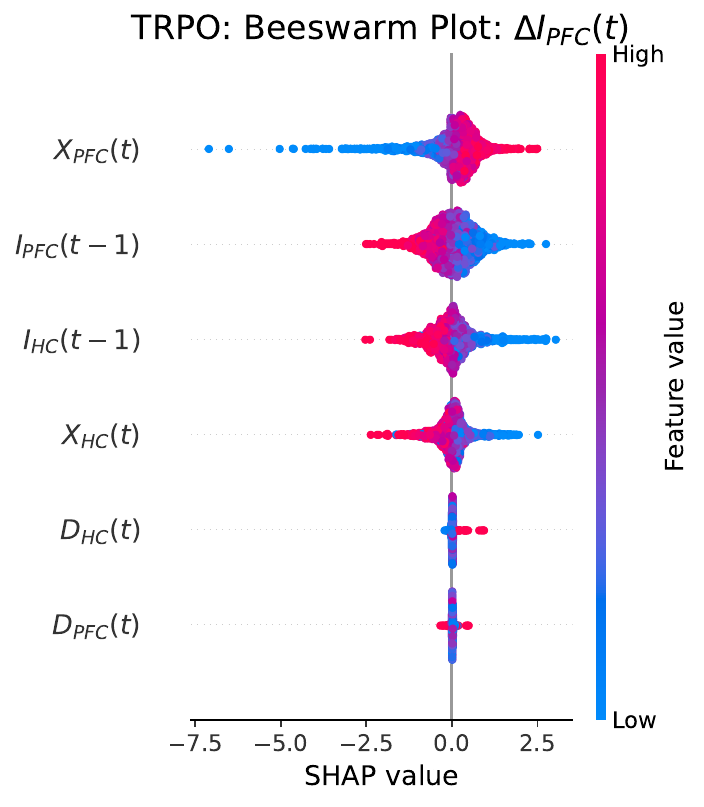}
    \end{subfigure}\\
    
    \begin{subfigure}{0.3\textwidth}
        \includegraphics[width=\linewidth]{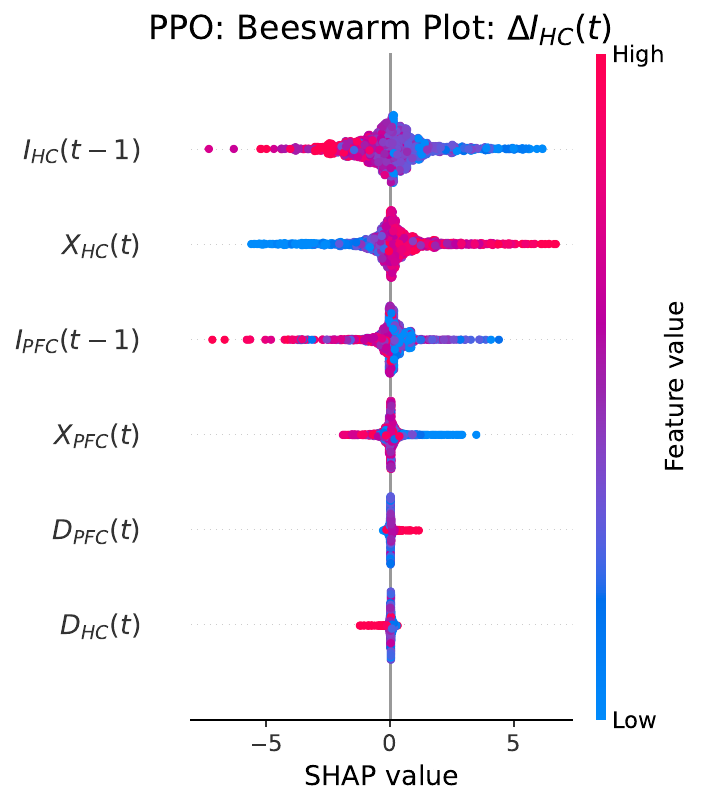}
    \end{subfigure}%
    \hspace{2cm}
    \begin{subfigure}{0.3\textwidth}
        \includegraphics[width=\linewidth]{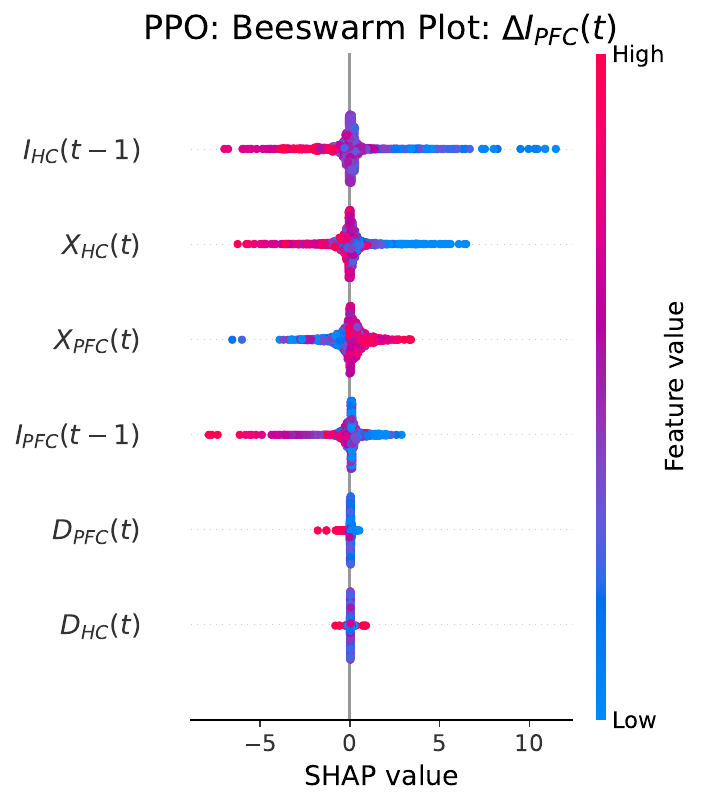}
    \end{subfigure}\\

    \begin{subfigure}{0.3\textwidth}
        \includegraphics[width=\linewidth]{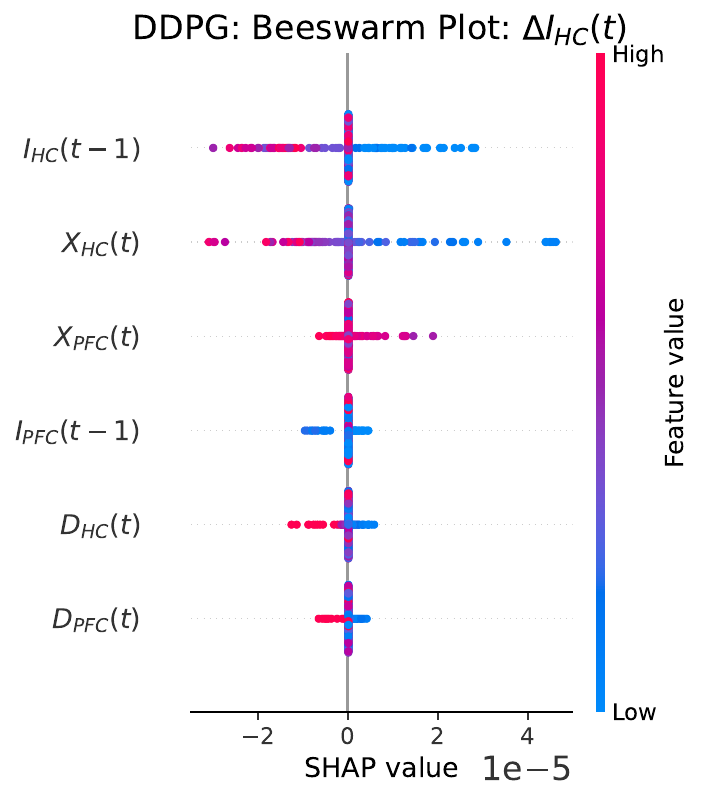}
    \end{subfigure}%
    \hspace{2cm}
    \begin{subfigure}{0.3\textwidth}
        \includegraphics[width=\linewidth]{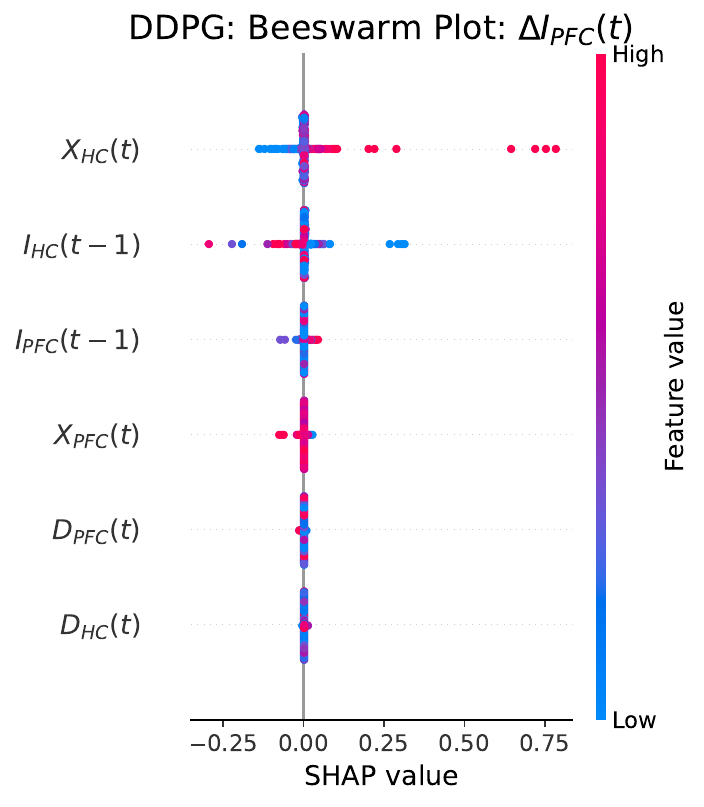}
    \end{subfigure}\\

    \begin{subfigure}{0.3\textwidth}
        \includegraphics[width=\linewidth]{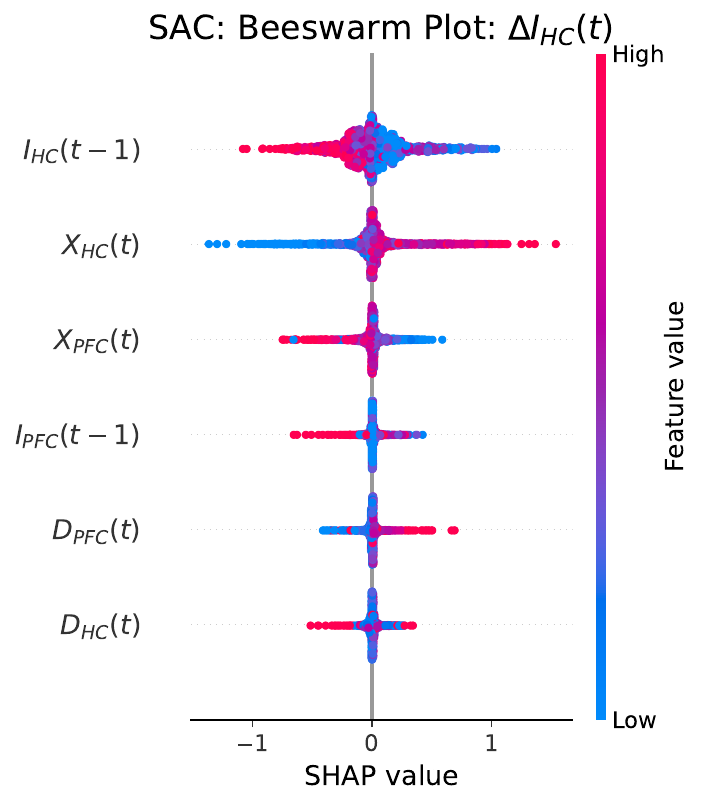}
    \end{subfigure}%
    \hspace{2cm}
    \begin{subfigure}{0.3\textwidth}
        \includegraphics[width=\linewidth]{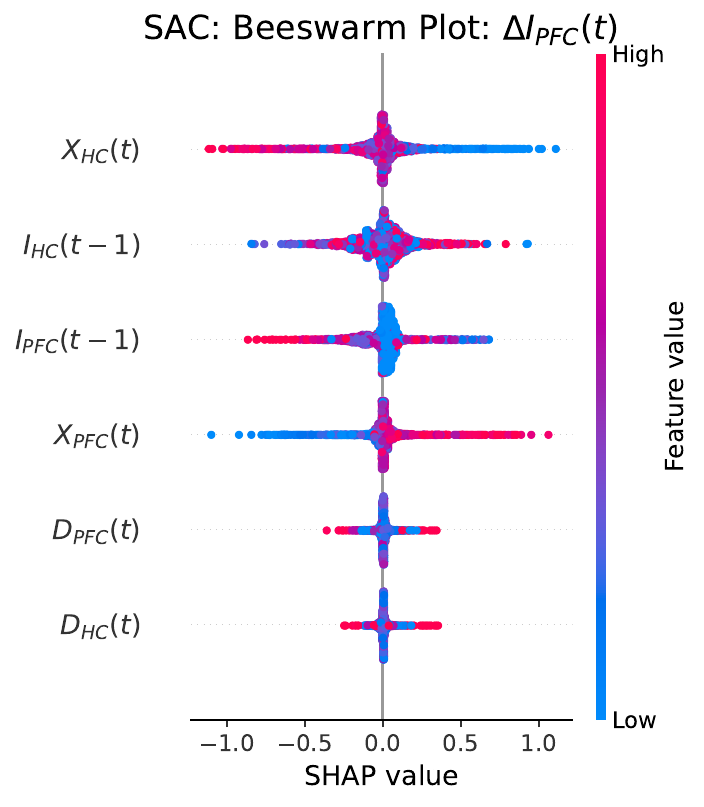}
    \end{subfigure}\\

    \caption{SHAP plots for global predictions of TRPO, PPO, DDPG, and SAC with Beeswarm plots explaining cognition prediction of each region, assigning distinctive colors to sample values (red high, blue low).}
    \label{fig:shap_global_bees}
\end{figure}

\begin{figure}[ht!]
    \centering
    \begin{subfigure}{0.15\textwidth}
        \includegraphics[width=\linewidth]{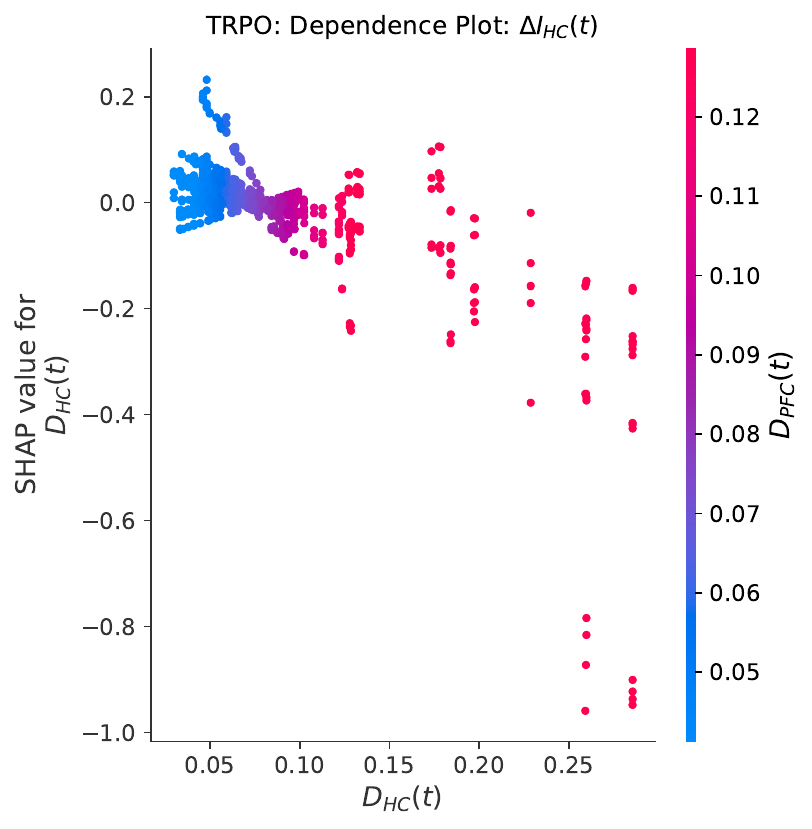}
    \end{subfigure}%
    \begin{subfigure}{0.15\textwidth}
        \includegraphics[width=\linewidth]{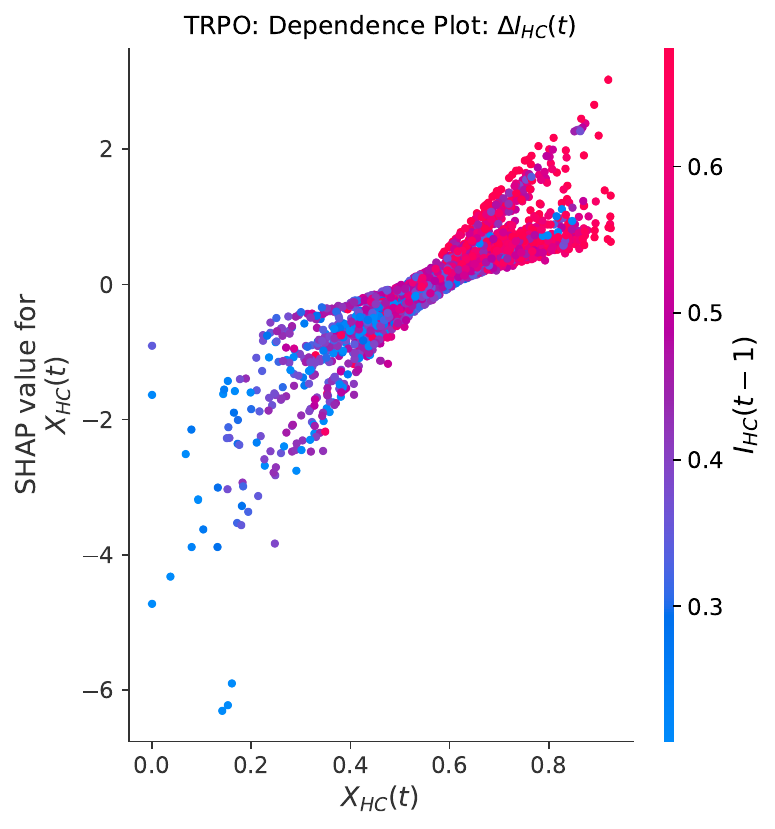}
    \end{subfigure}%
    \begin{subfigure}{0.15\textwidth}
        \includegraphics[width=\linewidth]{plots/shap_global/TRPO_dependence_HC_D_HC.pdf}
    \end{subfigure}%
    \begin{subfigure}{0.15\textwidth}
        \includegraphics[width=\linewidth]{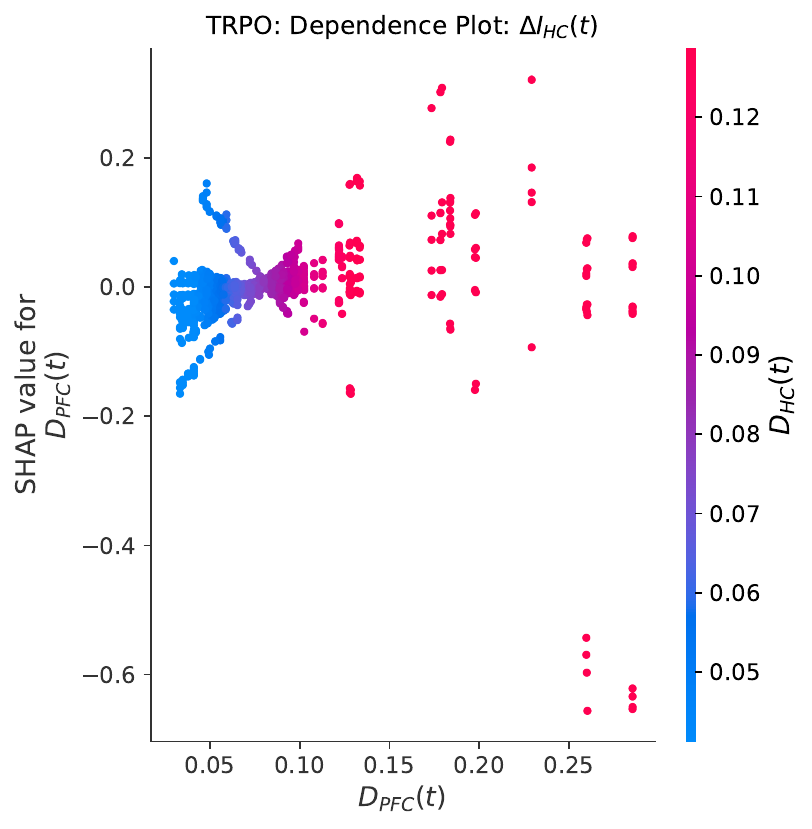}
    \end{subfigure}%
    \begin{subfigure}{0.15\textwidth}
        \includegraphics[width=\linewidth]{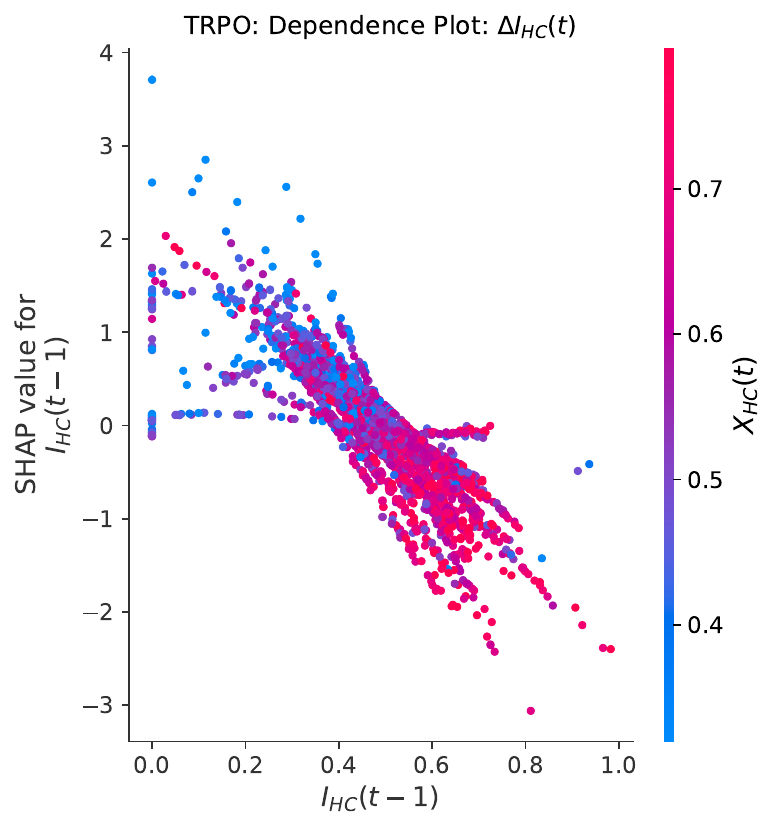}
    \end{subfigure}%
    \begin{subfigure}{0.15\textwidth}
        \includegraphics[width=\linewidth]{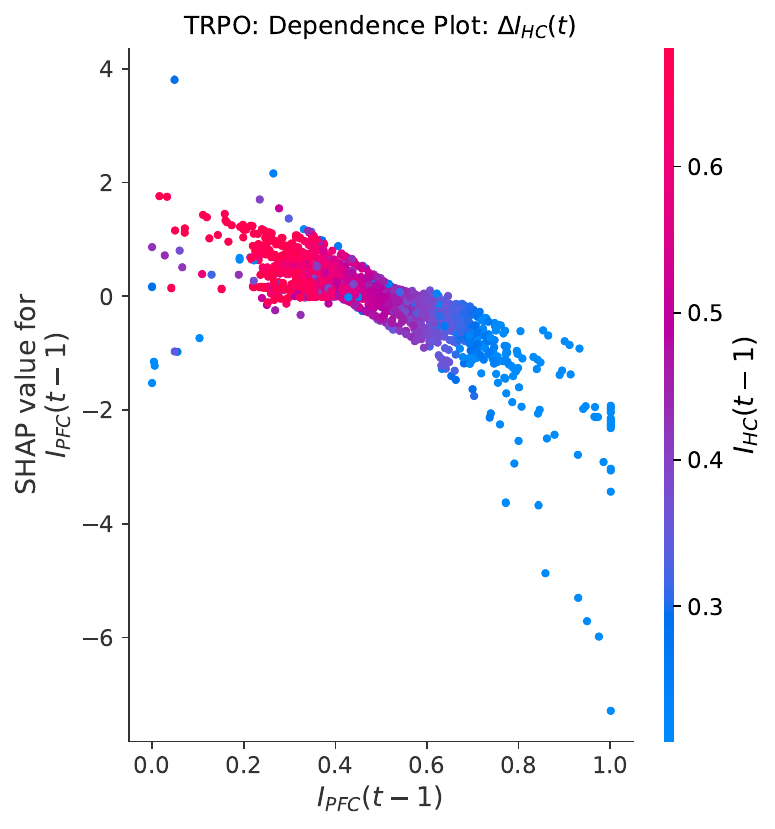}
    \end{subfigure}\\

    \begin{subfigure}{0.15\textwidth}
        \includegraphics[width=\linewidth]{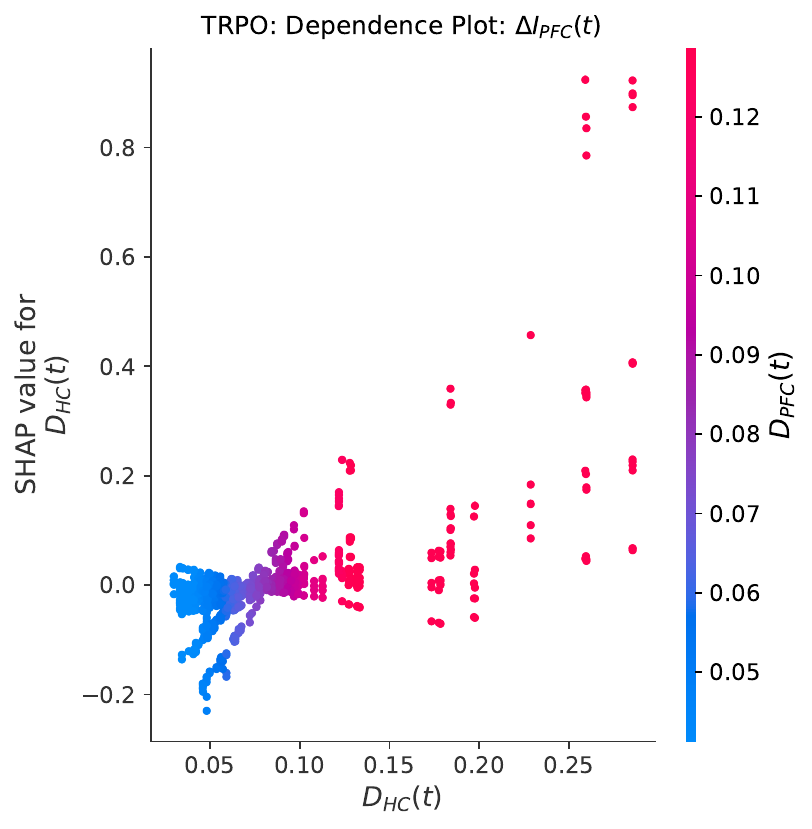}
    \end{subfigure}%
    \begin{subfigure}{0.15\textwidth}
        \includegraphics[width=\linewidth]{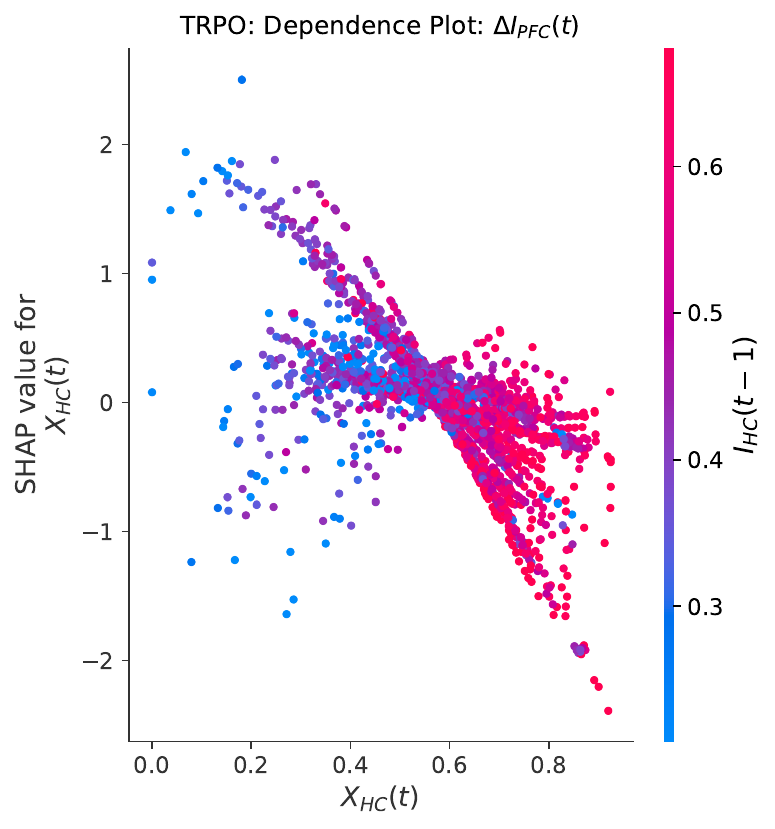}
    \end{subfigure}%
    \begin{subfigure}{0.15\textwidth}
        \includegraphics[width=\linewidth]{plots/shap_global/TRPO_dependence_PFC_D_HC.pdf}
    \end{subfigure}%
    \begin{subfigure}{0.15\textwidth}
        \includegraphics[width=\linewidth]{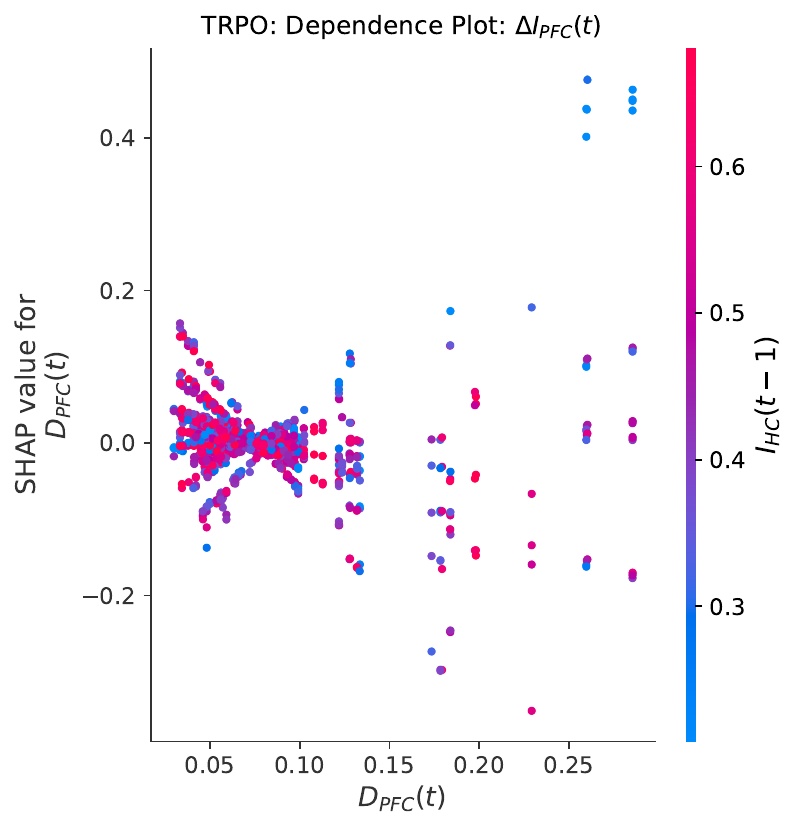}
    \end{subfigure}%
    \begin{subfigure}{0.15\textwidth}
        \includegraphics[width=\linewidth]{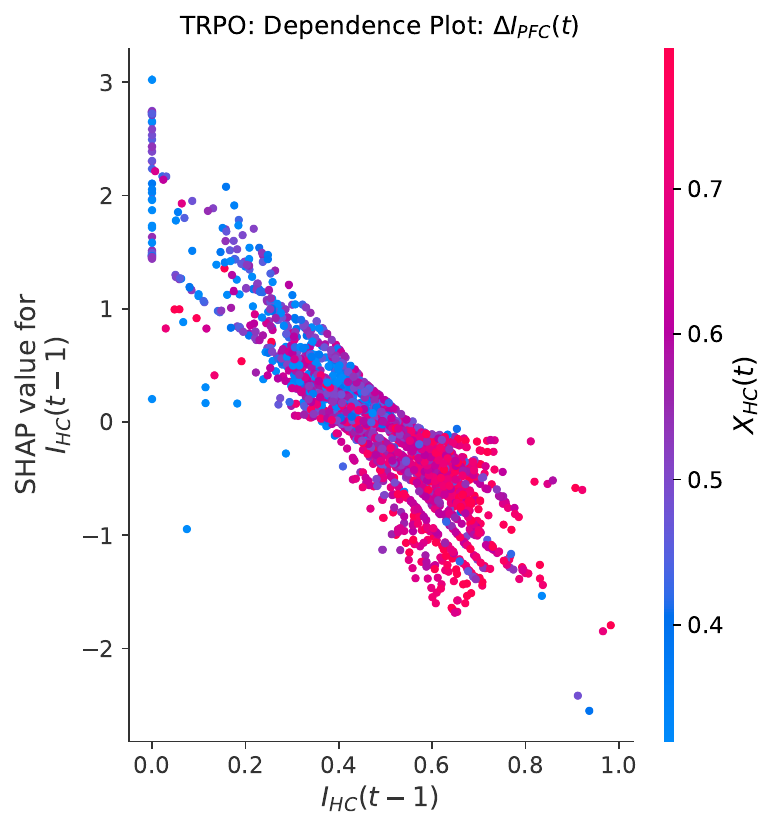}
    \end{subfigure}%
    \begin{subfigure}{0.15\textwidth}
        \includegraphics[width=\linewidth]{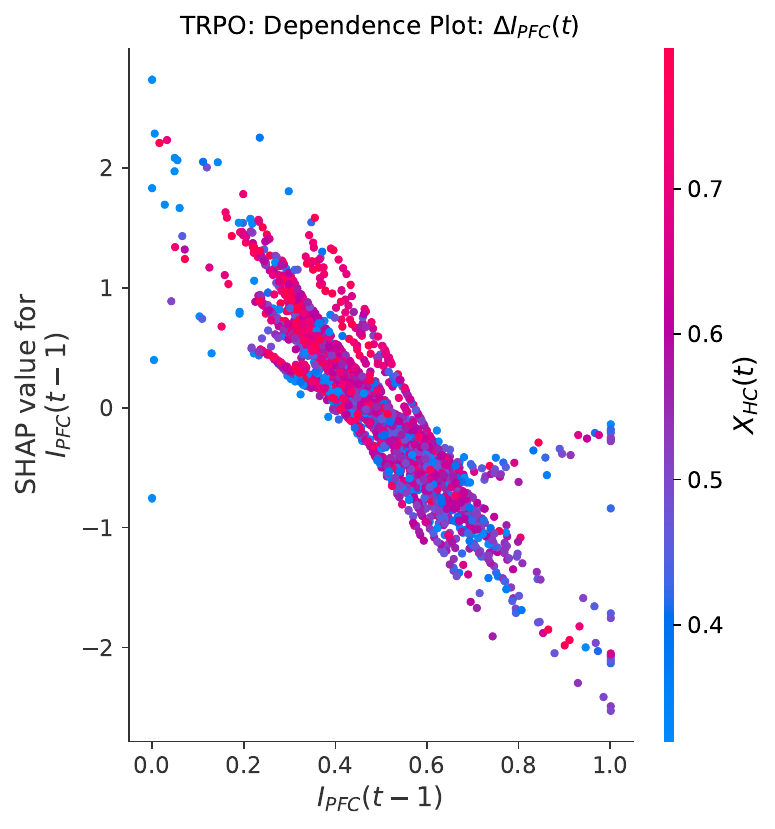}
    \end{subfigure}\\
    
    
    \begin{subfigure}{0.15\textwidth}
         \includegraphics[width=\linewidth]{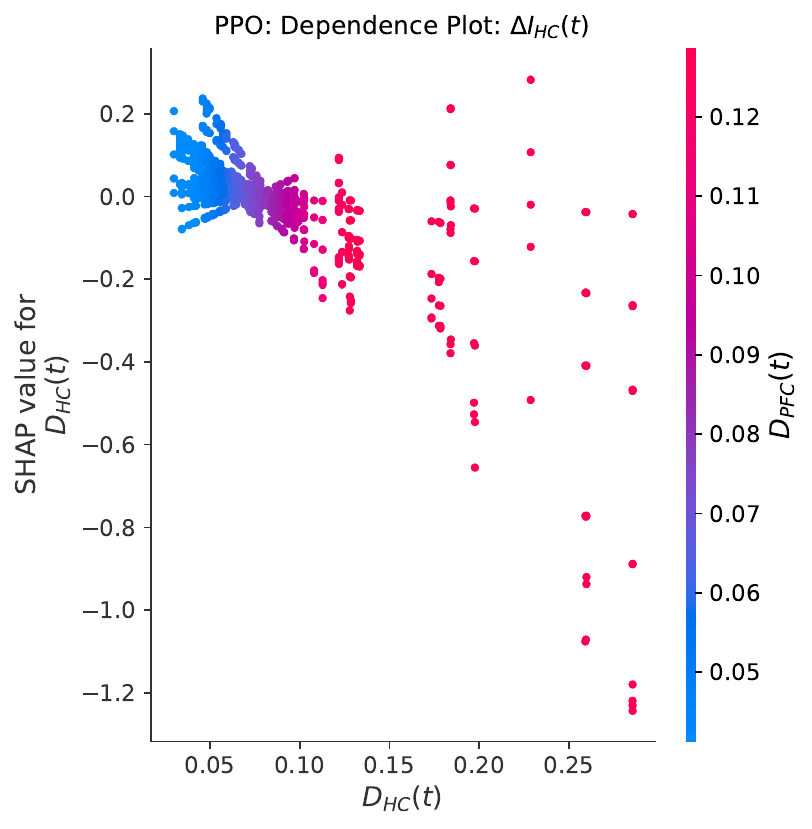}
    \end{subfigure}%
    \begin{subfigure}{0.15\textwidth}
        \includegraphics[width=\linewidth]{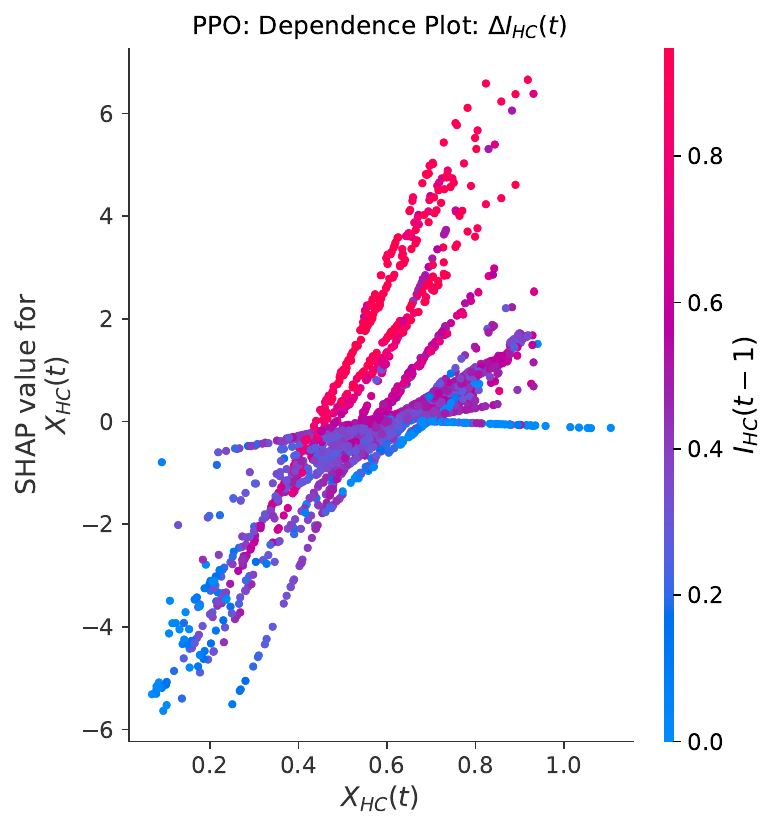}
    \end{subfigure}%
    \begin{subfigure}{0.15\textwidth}
        \includegraphics[width=\linewidth]{plots/shap_global/PPO_dependence_HC_D_HC.pdf}
    \end{subfigure}%
    \begin{subfigure}{0.15\textwidth}
        \includegraphics[width=\linewidth]{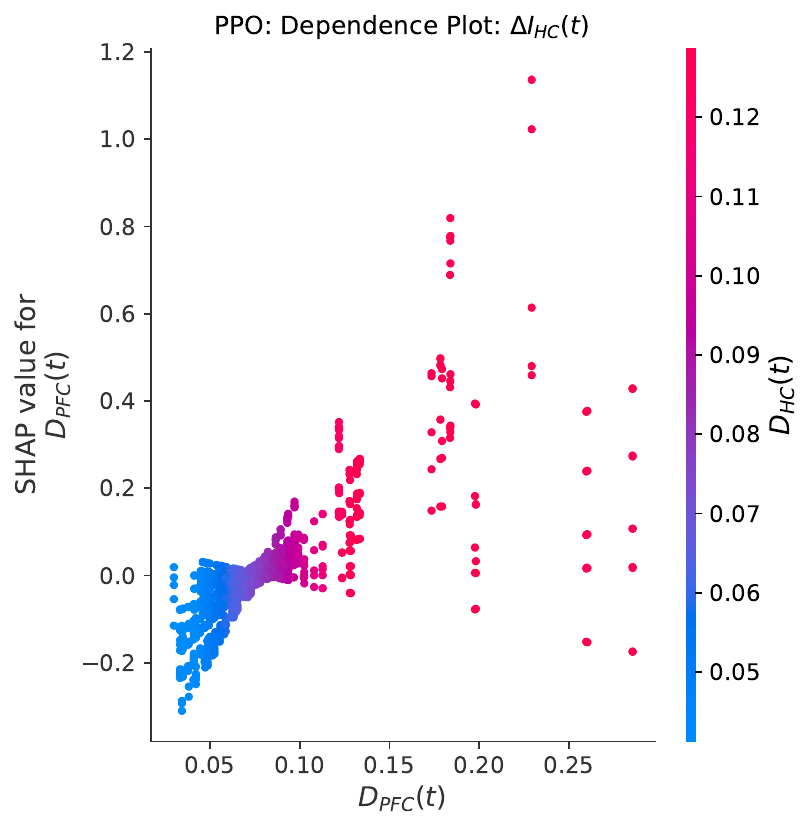}
    \end{subfigure}%
    \begin{subfigure}{0.15\textwidth}
        \includegraphics[width=\linewidth]{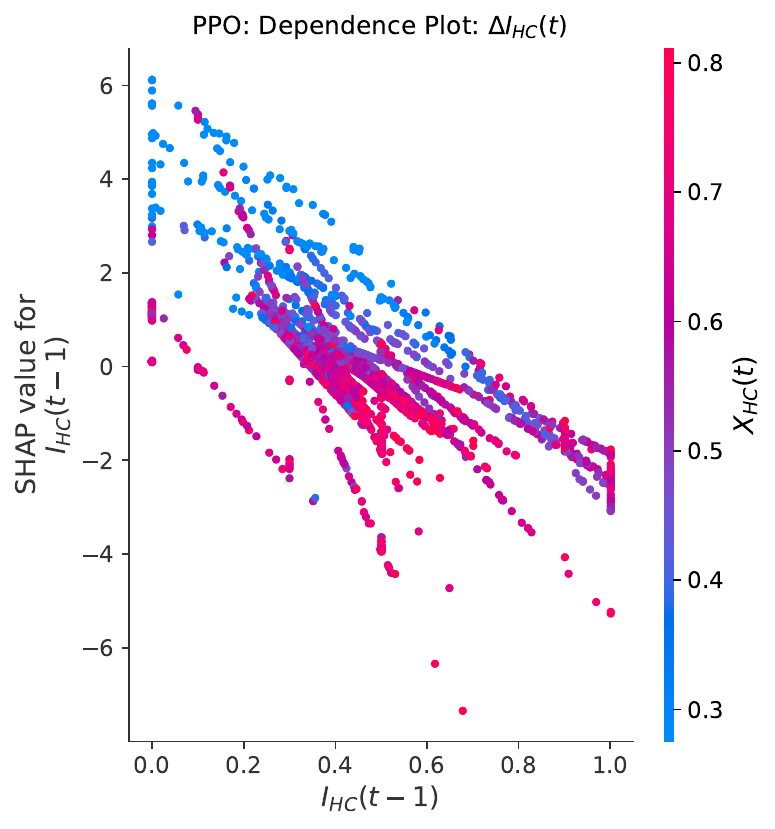}
    \end{subfigure}%
    \begin{subfigure}{0.15\textwidth}
        \includegraphics[width=\linewidth]{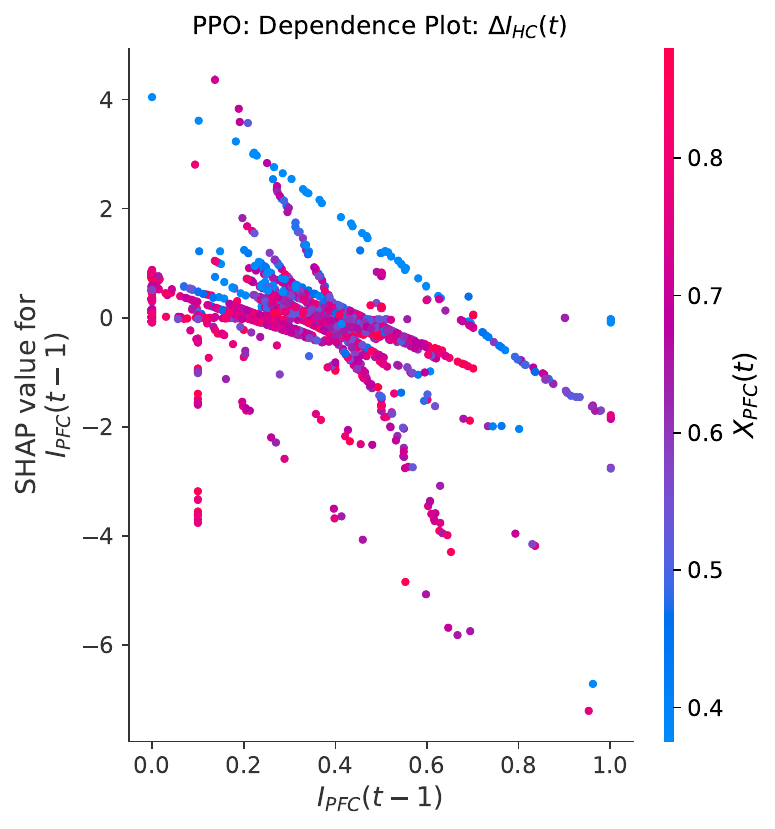}
    \end{subfigure}\\

    \begin{subfigure}{0.15\textwidth}
        \includegraphics[width=\linewidth]{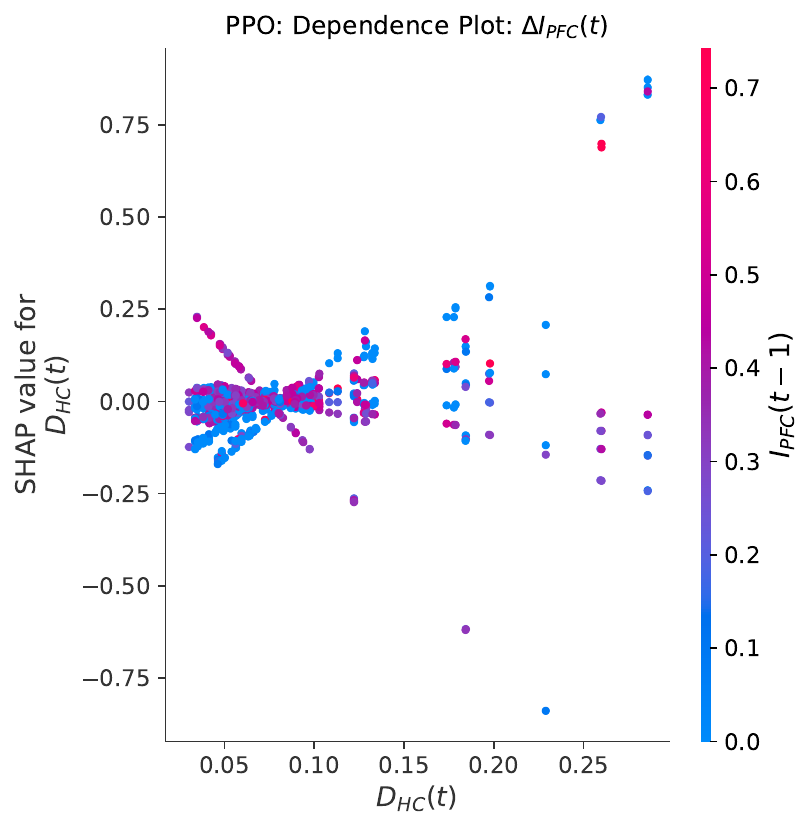}
    \end{subfigure}%
    \begin{subfigure}{0.15\textwidth}
        \includegraphics[width=\linewidth]{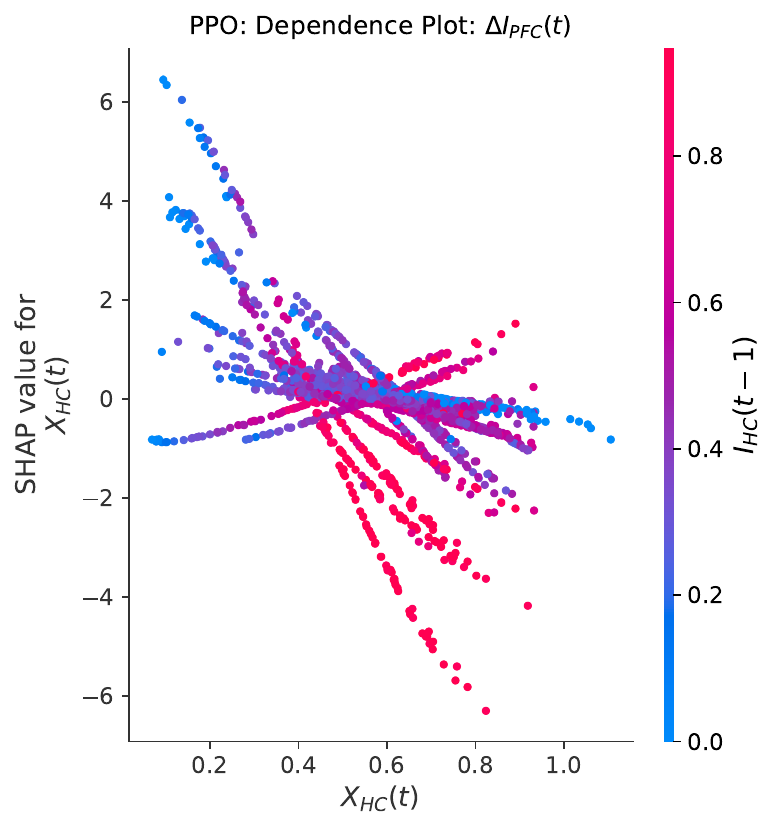}
    \end{subfigure}%
    \begin{subfigure}{0.15\textwidth}
        \includegraphics[width=\linewidth]{plots/shap_global/PPO_dependence_PFC_D_HC.pdf}
    \end{subfigure}%
    \begin{subfigure}{0.15\textwidth}
        \includegraphics[width=\linewidth]{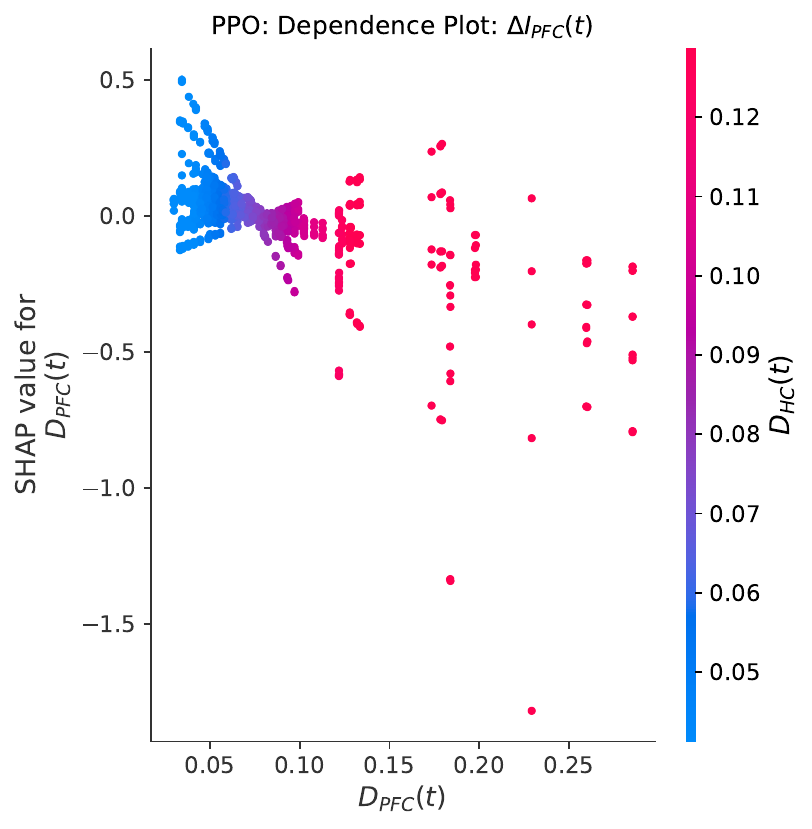}
    \end{subfigure}%
    \begin{subfigure}{0.15\textwidth}
        \includegraphics[width=\linewidth]{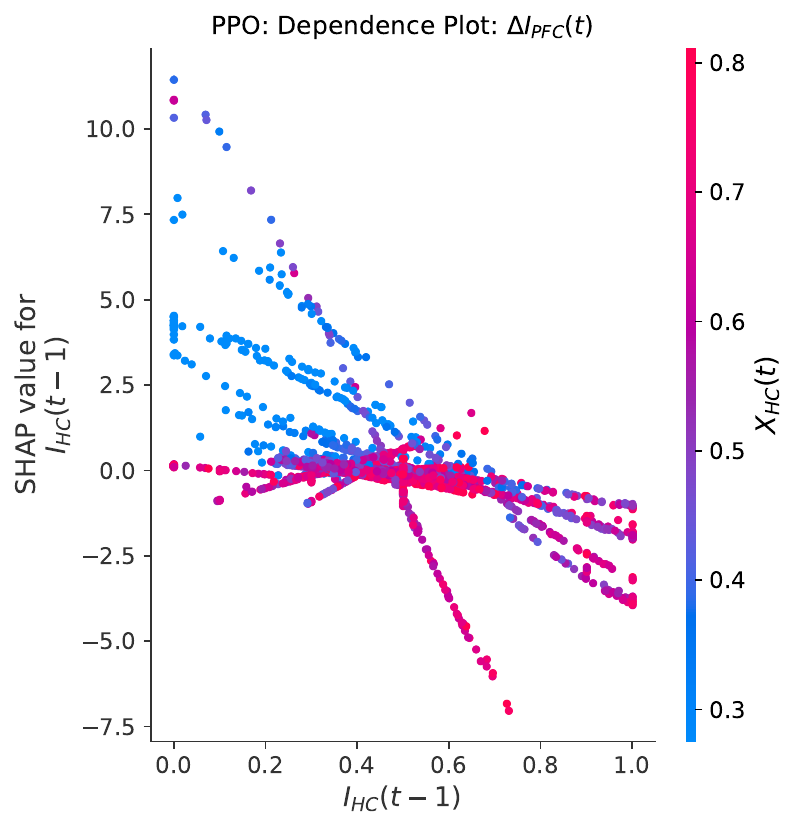}
    \end{subfigure}%
    \begin{subfigure}{0.15\textwidth}
        \includegraphics[width=\linewidth]{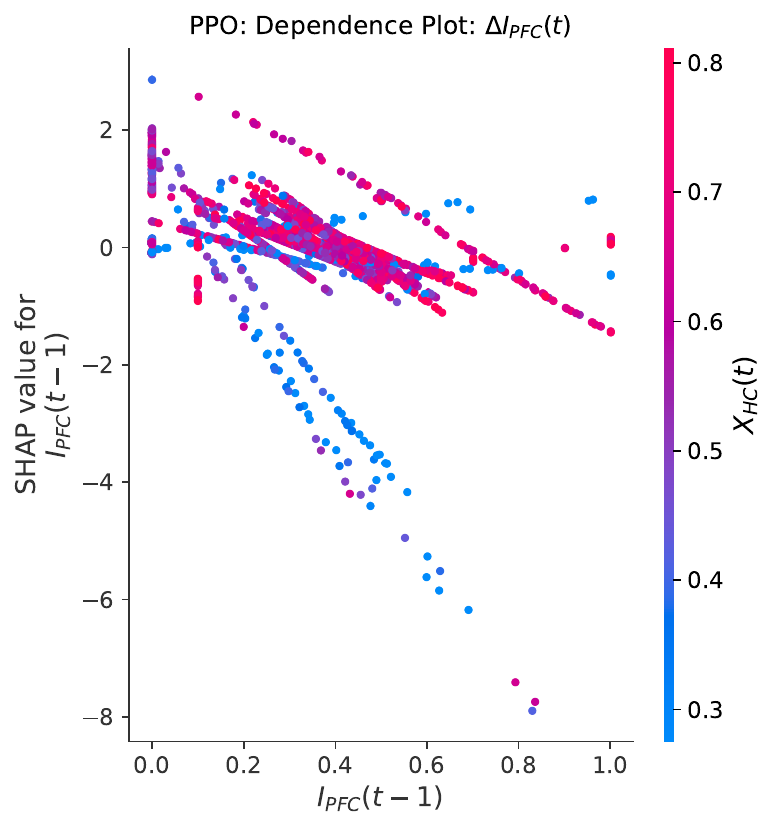}
    \end{subfigure}\\
    
    \begin{subfigure}{0.15\textwidth}
         \includegraphics[width=\linewidth]{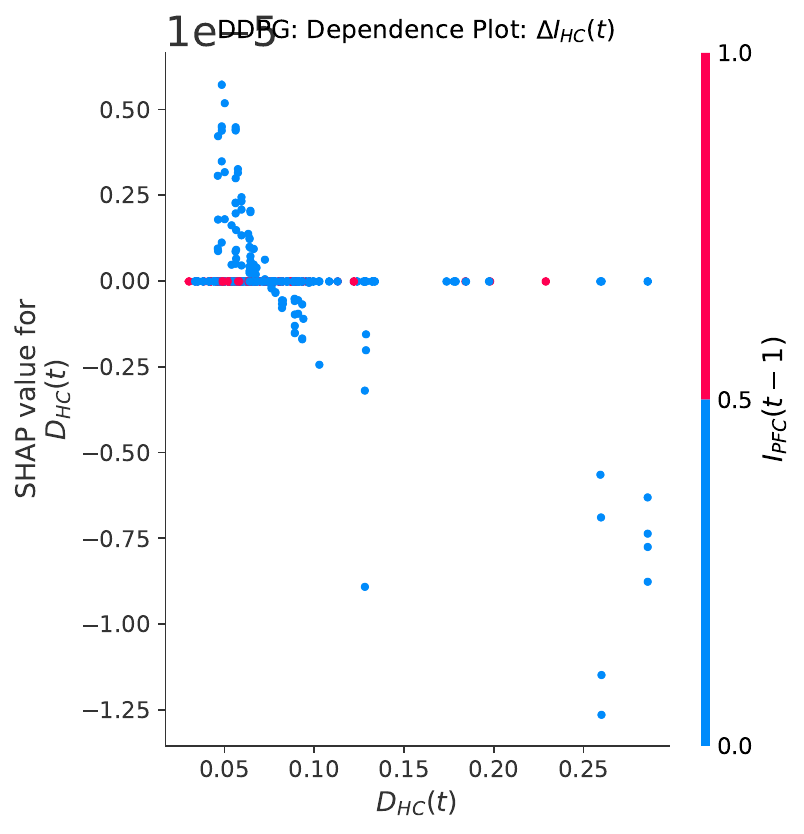}
    \end{subfigure}%
    \begin{subfigure}{0.15\textwidth}
        \includegraphics[width=\linewidth]{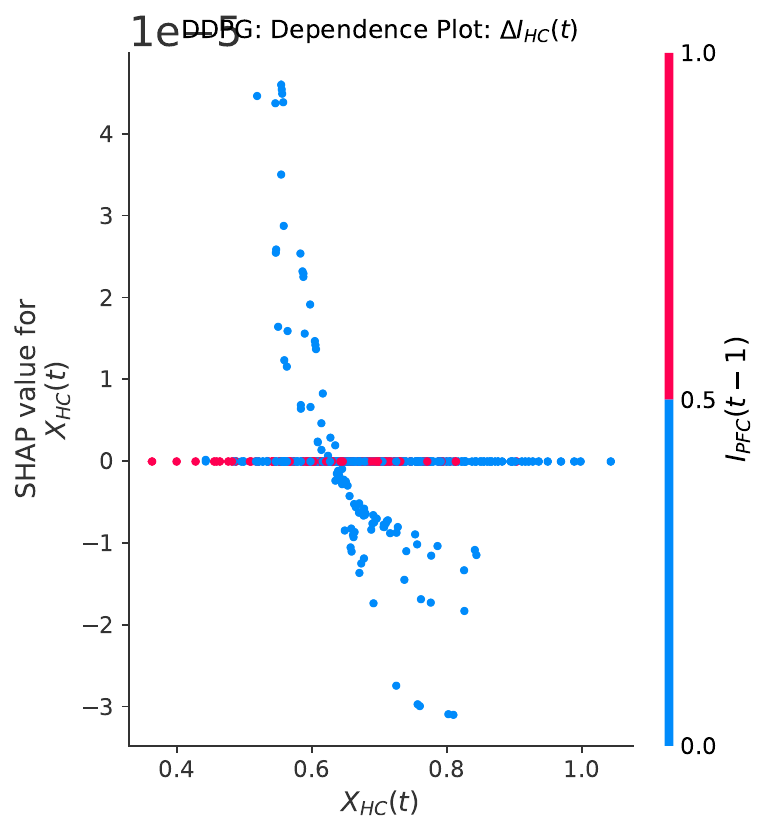}
    \end{subfigure}%
    \begin{subfigure}{0.15\textwidth}
        \includegraphics[width=\linewidth]{plots/shap_global/DDPG_dependence_HC_D_HC.pdf}
    \end{subfigure}%
    \begin{subfigure}{0.15\textwidth}
        \includegraphics[width=\linewidth]{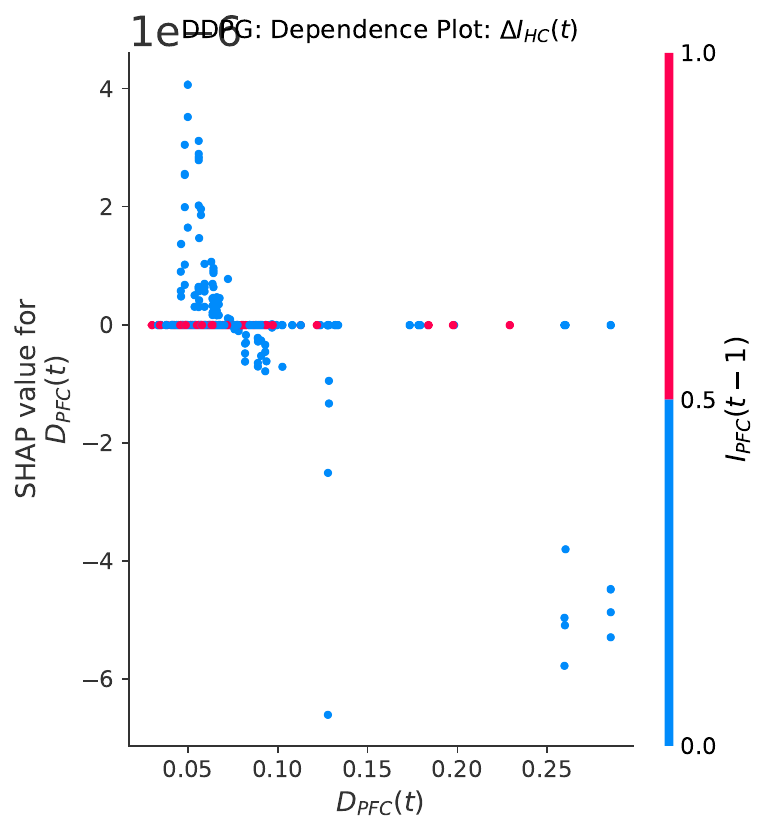}
    \end{subfigure}%
    \begin{subfigure}{0.15\textwidth}
        \includegraphics[width=\linewidth]{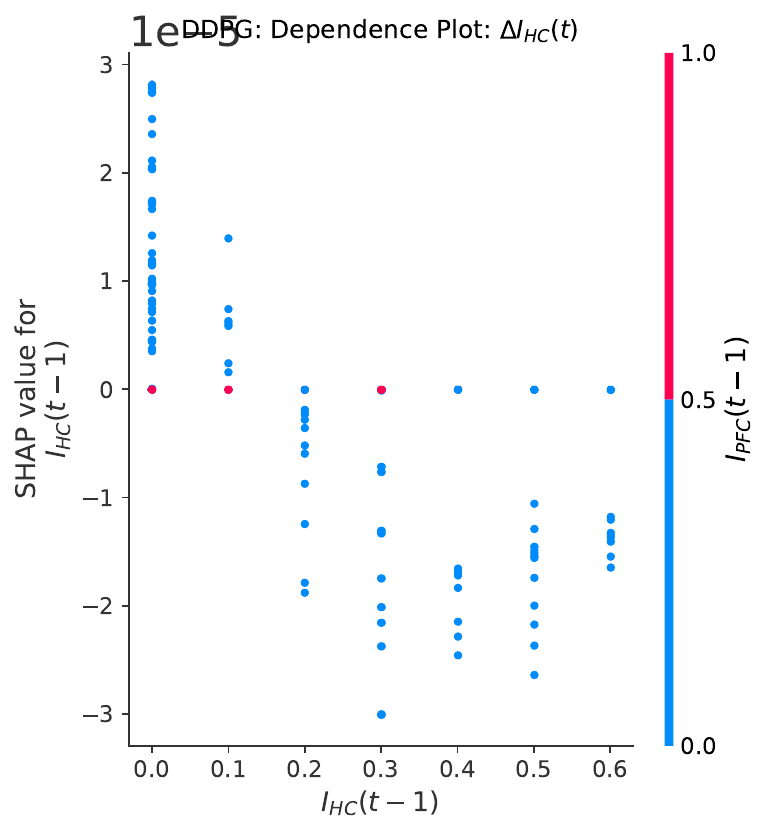}
    \end{subfigure}%
    \begin{subfigure}{0.15\textwidth}
        \includegraphics[width=\linewidth]{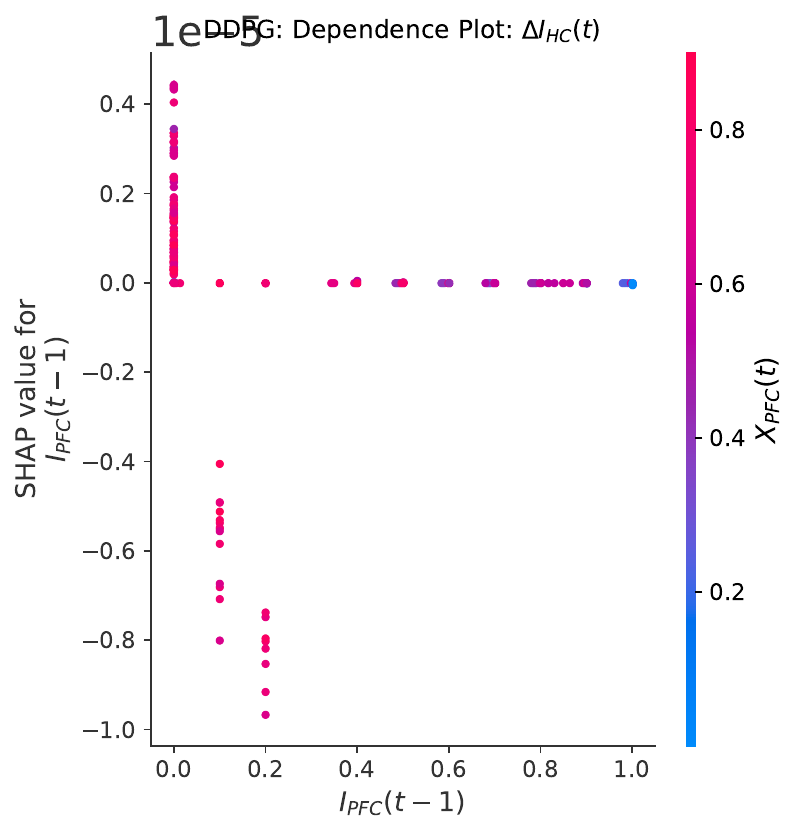}
    \end{subfigure}\\

    \begin{subfigure}{0.15\textwidth}
        \includegraphics[width=\linewidth]{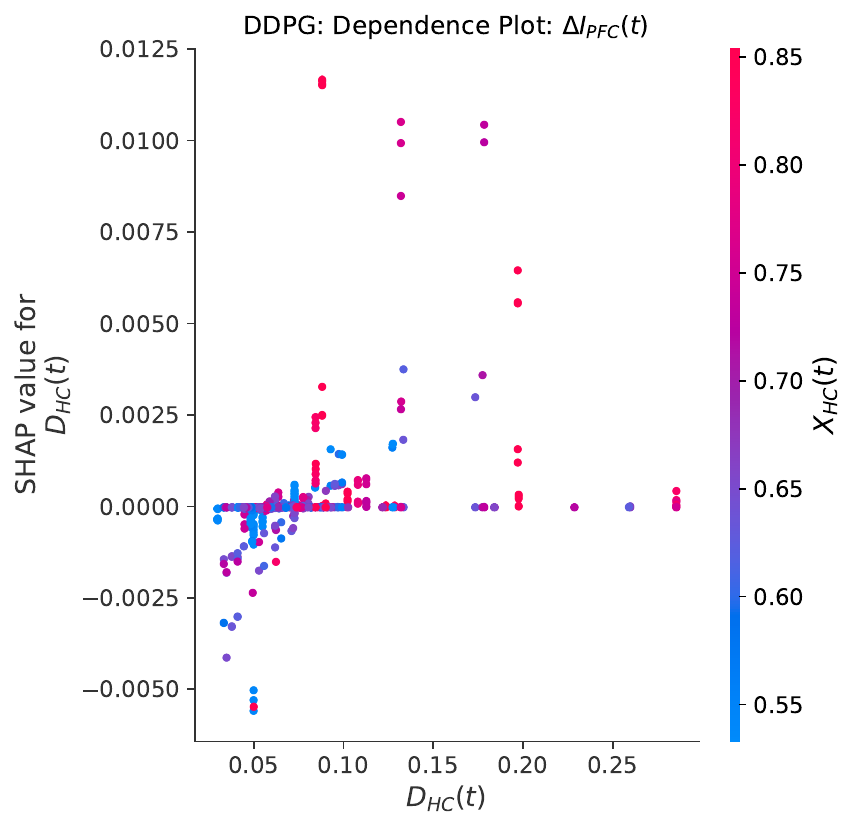}
    \end{subfigure}%
    \begin{subfigure}{0.15\textwidth}
        \includegraphics[width=\linewidth]{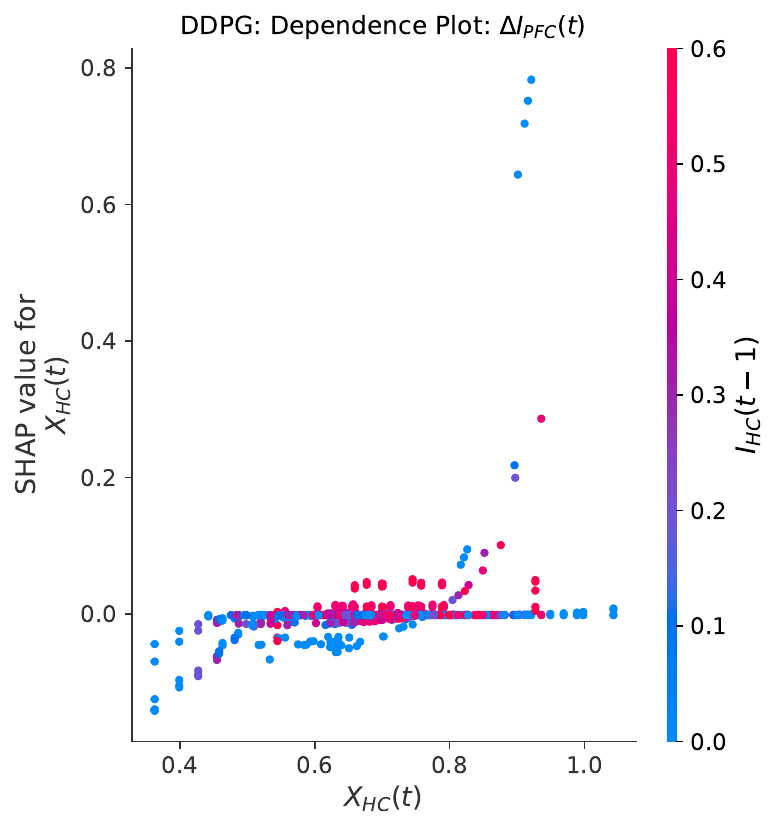}
    \end{subfigure}%
    \begin{subfigure}{0.15\textwidth}
        \includegraphics[width=\linewidth]{plots/shap_global/DDPG_dependence_PFC_D_HC.pdf}
    \end{subfigure}%
    \begin{subfigure}{0.15\textwidth}
        \includegraphics[width=\linewidth]{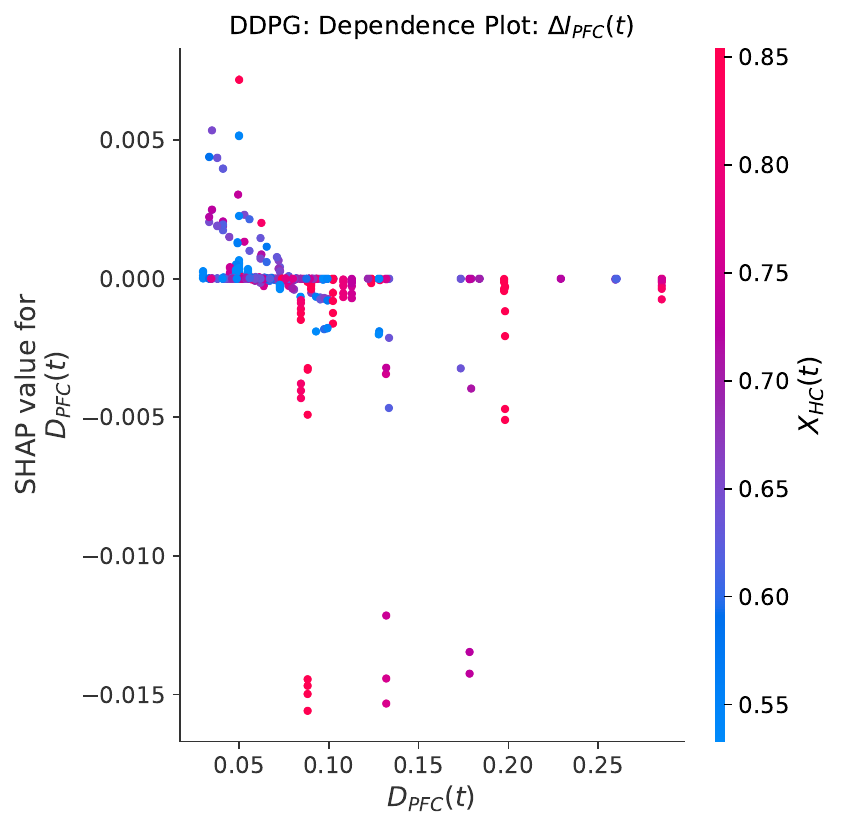}
    \end{subfigure}%
    \begin{subfigure}{0.15\textwidth}
        \includegraphics[width=\linewidth]{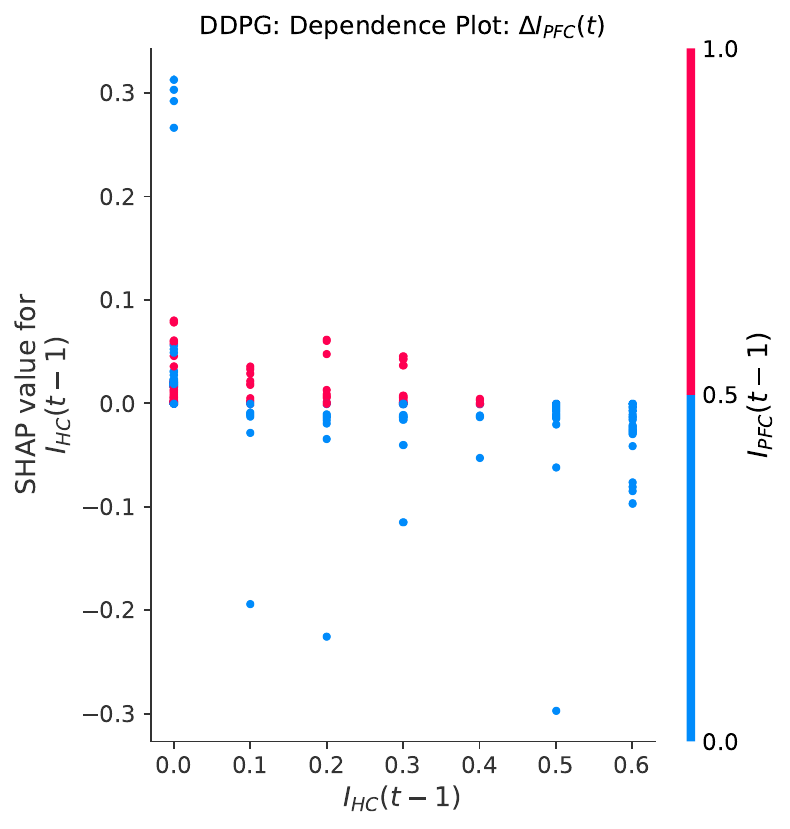}
    \end{subfigure}%
    \begin{subfigure}{0.15\textwidth}
        \includegraphics[width=\linewidth]{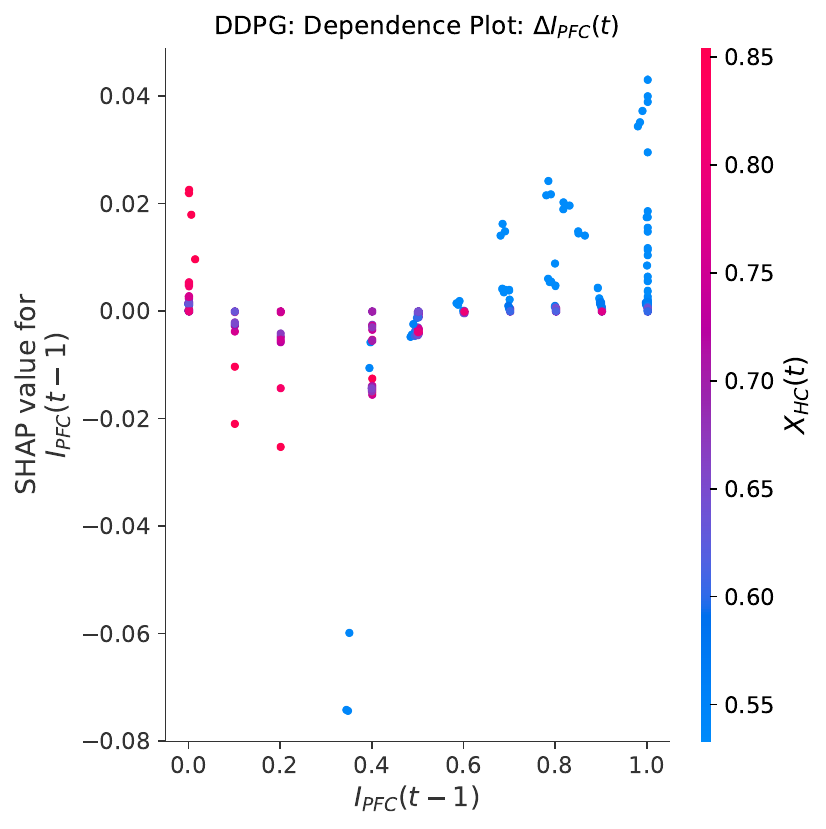}
    \end{subfigure}\\
    
    \begin{subfigure}{0.15\textwidth}
         \includegraphics[width=\linewidth]{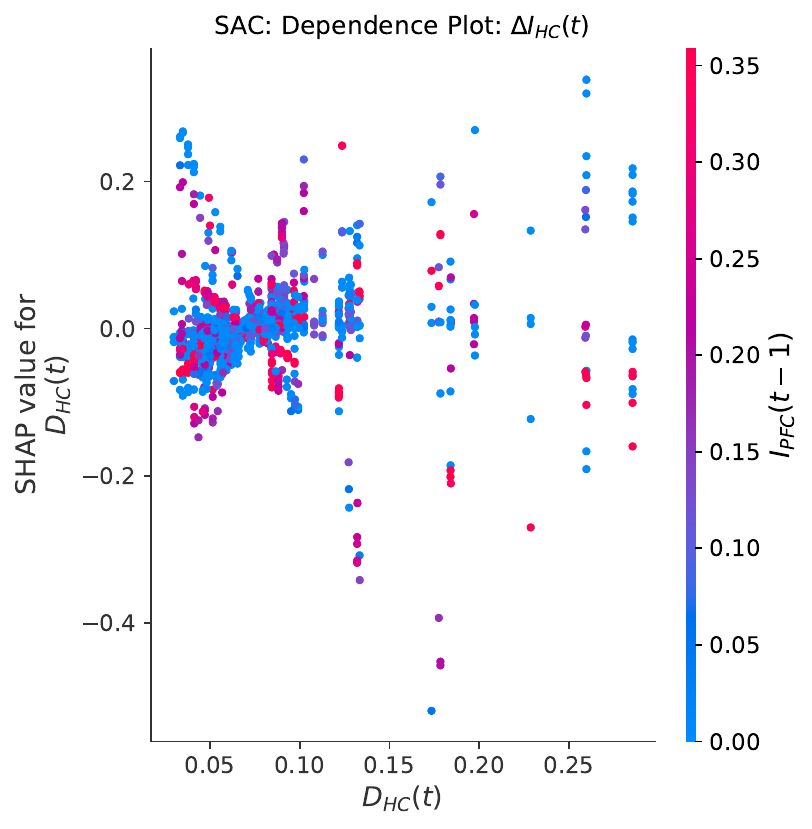}
    \end{subfigure}%
    \begin{subfigure}{0.15\textwidth}
        \includegraphics[width=\linewidth]{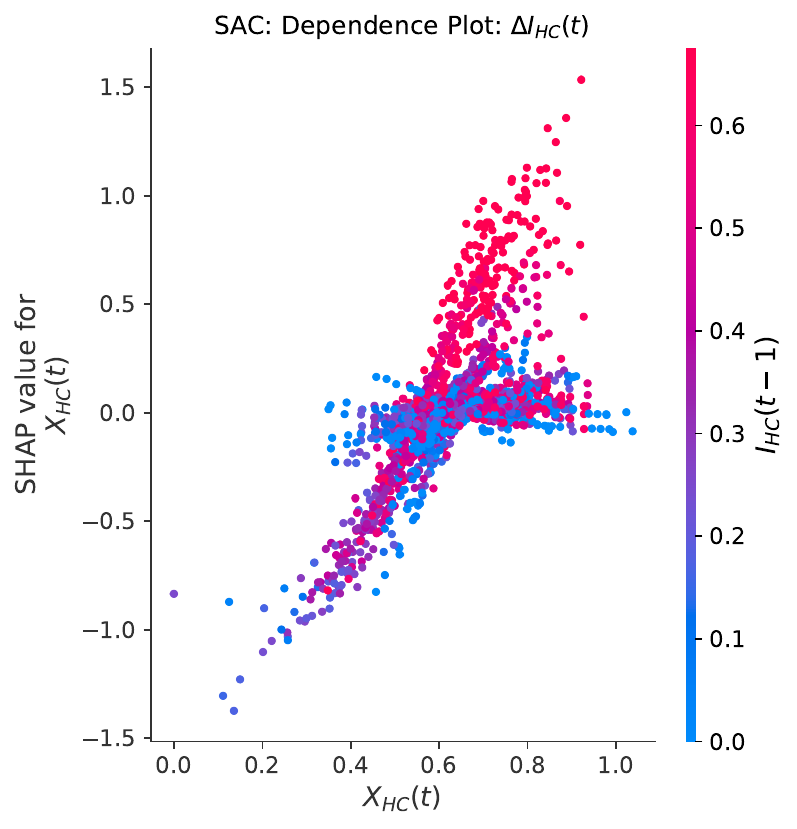}
    \end{subfigure}%
    \begin{subfigure}{0.15\textwidth}
        \includegraphics[width=\linewidth]{plots/shap_global/SAC_dependence_HC_D_HC.pdf}
    \end{subfigure}%
    \begin{subfigure}{0.15\textwidth}
        \includegraphics[width=\linewidth]{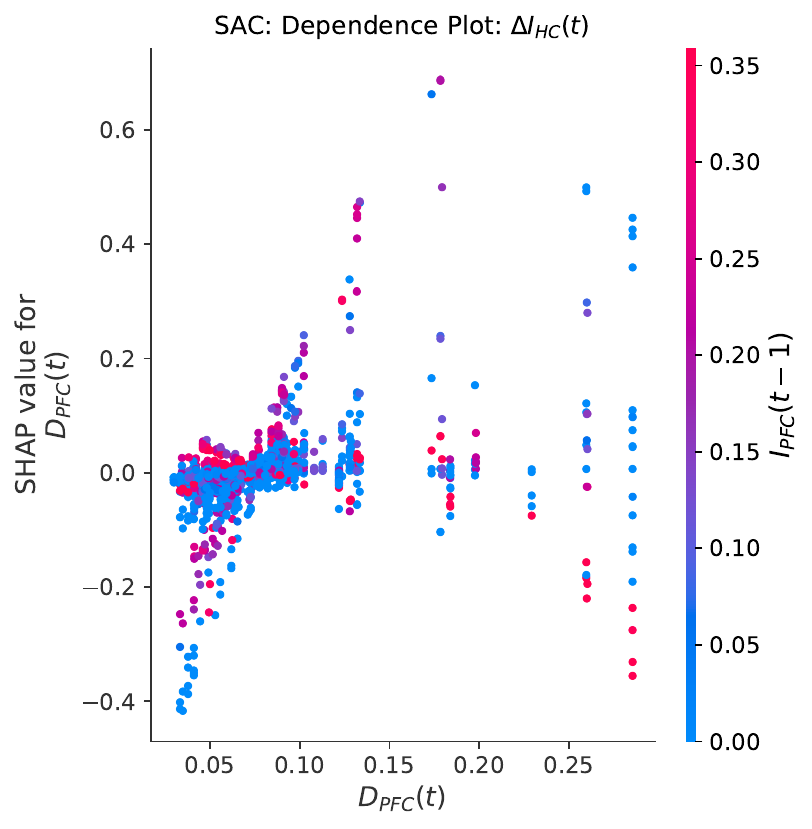}
    \end{subfigure}%
    \begin{subfigure}{0.15\textwidth}
        \includegraphics[width=\linewidth]{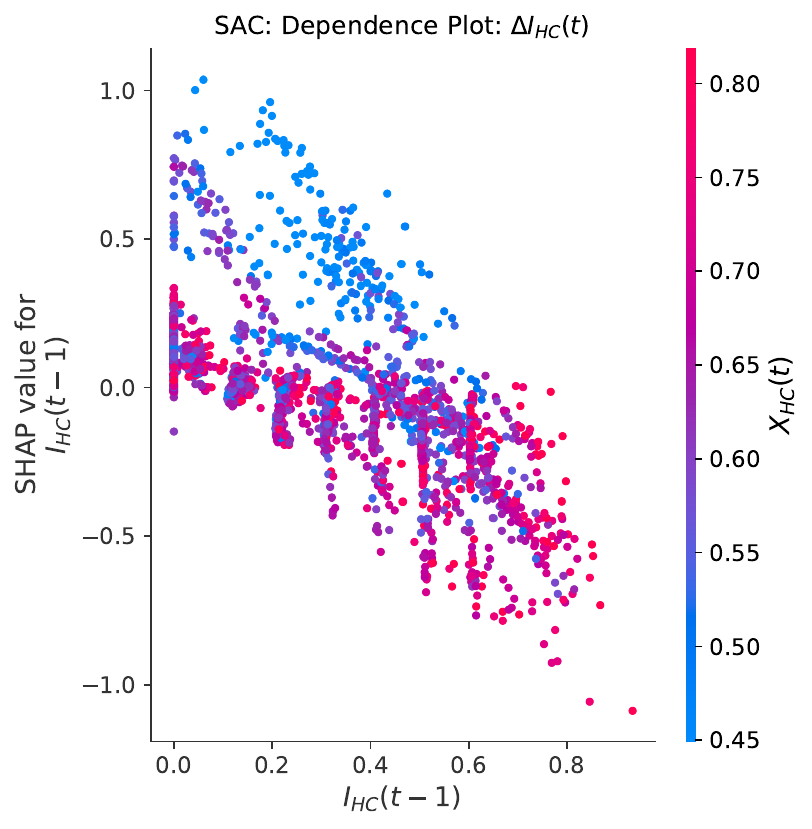}
    \end{subfigure}%
    \begin{subfigure}{0.15\textwidth}
        \includegraphics[width=\linewidth]{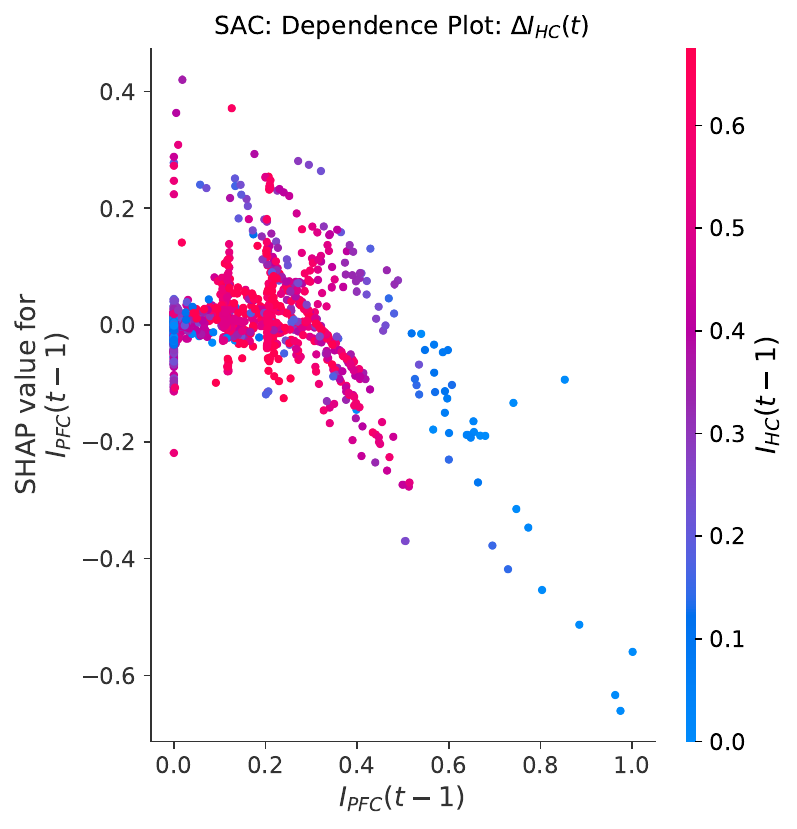}
    \end{subfigure}\\

    \begin{subfigure}{0.15\textwidth}
        \includegraphics[width=\linewidth]{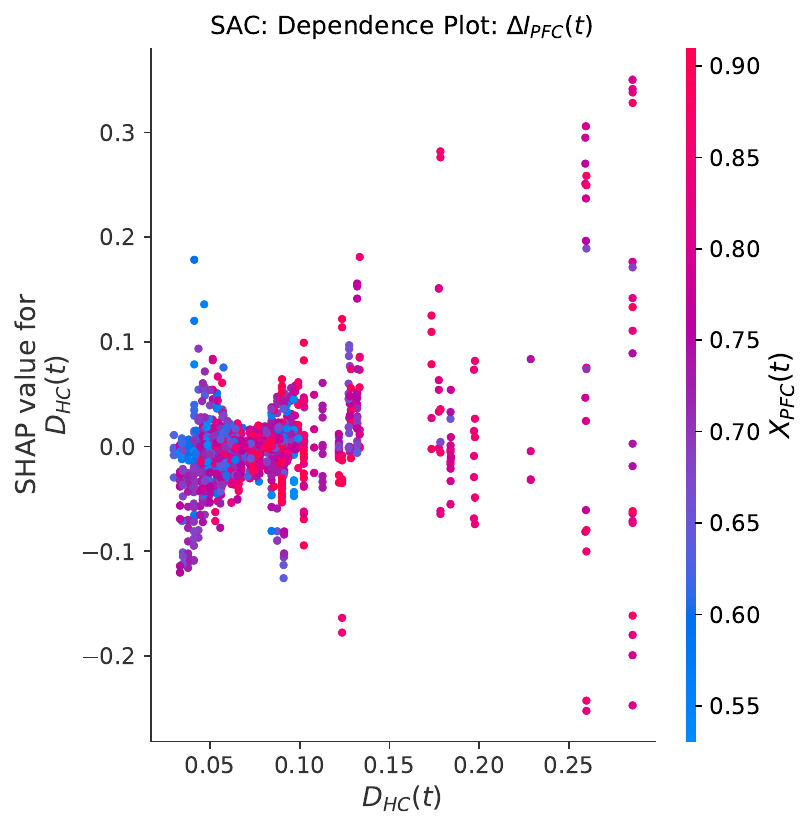}
    \end{subfigure}%
    \begin{subfigure}{0.15\textwidth}
        \includegraphics[width=\linewidth]{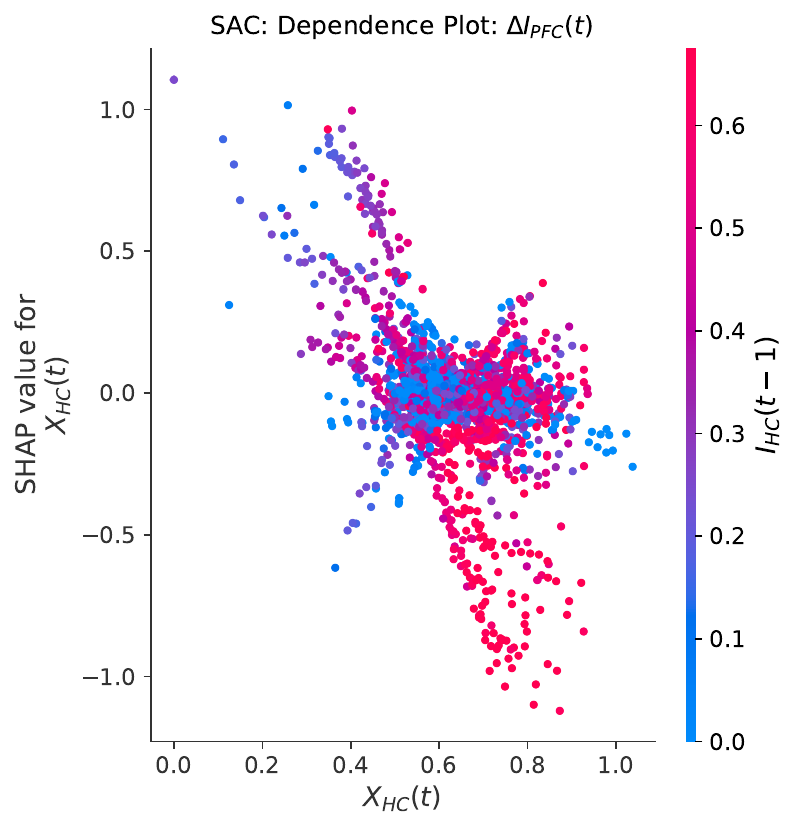}
    \end{subfigure}%
    \begin{subfigure}{0.15\textwidth}
        \includegraphics[width=\linewidth]{plots/shap_global/SAC_dependence_PFC_D_HC.pdf}
    \end{subfigure}%
    \begin{subfigure}{0.15\textwidth}
        \includegraphics[width=\linewidth]{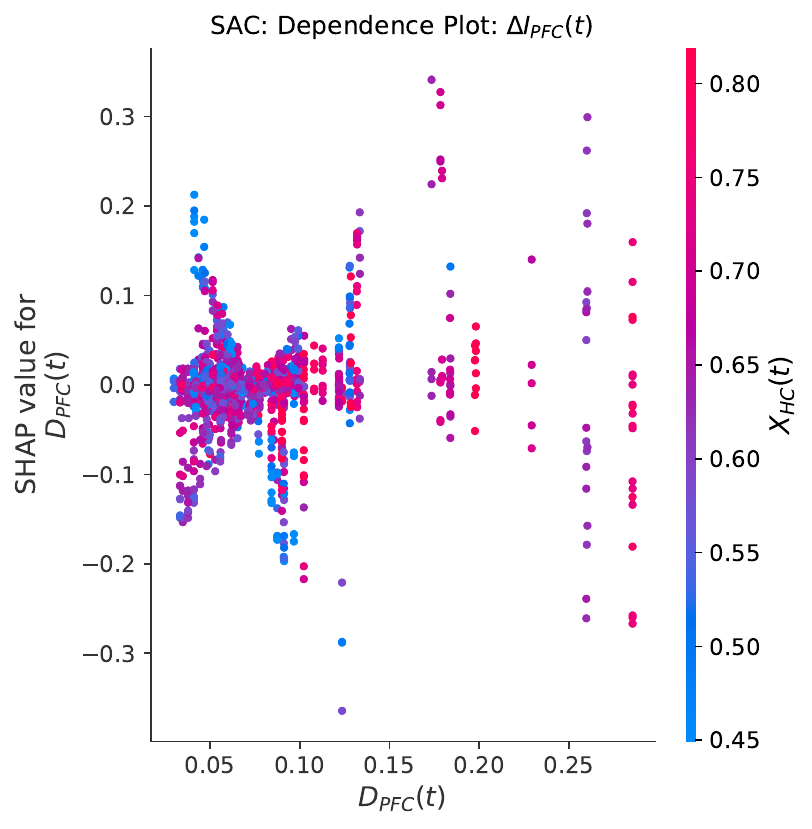}
    \end{subfigure}%
    \begin{subfigure}{0.15\textwidth}
        \includegraphics[width=\linewidth]{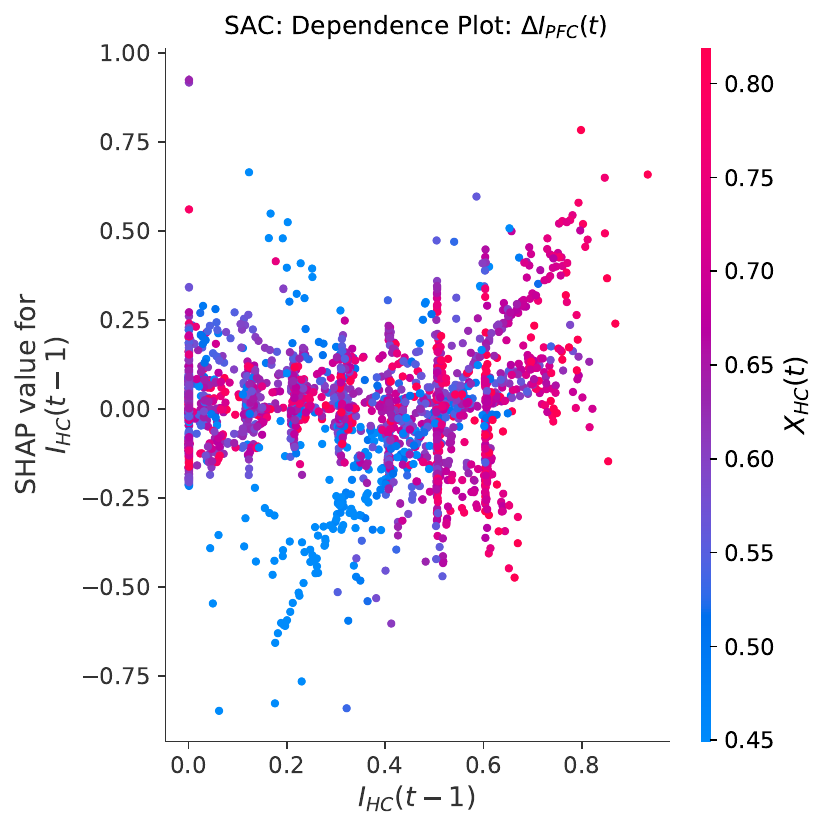}
    \end{subfigure}%
    \begin{subfigure}{0.15\textwidth}
        \includegraphics[width=\linewidth]{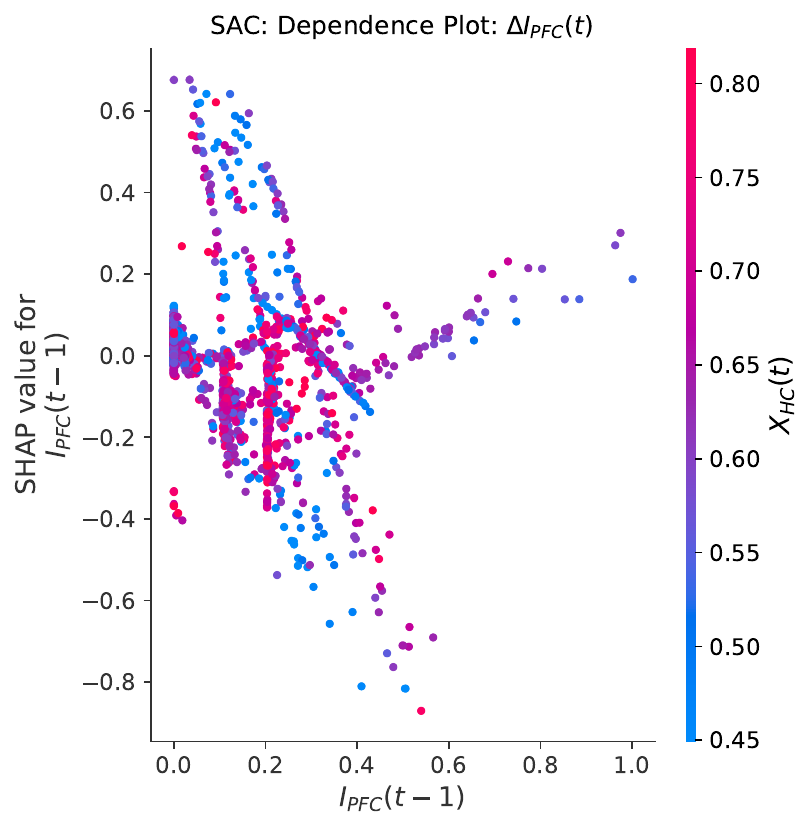}
    \end{subfigure}\\    

    \caption{SHAP dependence plots for global predictions of all four methods. Two rows for each method. 6 input features from left to right. First row corresponds to first action $\Delta I_{HC}$ and second row for action $\Delta I_{PFC}$.}
    \label{fig:shap_global_dependence}
\end{figure}

Figure \ref{fig:shap_global_bees} shows the Beeswarm plots for all four RL methods, for the two model predictions (change in information processing for each region). When comparing with the actual ground truth and RL predictions as shown in Figure \ref{fig:comparison_all_5}, the SHAP plots can shed a better light on why the different methods behave differently. PPO and SAC's SHAP values for $\Delta I_{PFC}$ are too concentrated towards the center when compared to TRPO whereas DDPG's SHAP values are exceedingly small to show that any of these input features affect the model's output. Figure \ref{fig:shap_global_dependence} shows dependence plots for the six input features/states and two outputs/actions, where we generate 12 SHAP dependence plots for each RL algorithm (two rows for each algorithm). These plots provide insights into how variations in individual input states influence the model's decision-making process. For example, we see TRPO being able to show a linear relationship between the increase in size of a region to the change in information processing of that region, capturing a recovery compensatory mechanism that was not in the design of the model. PPO shows some of that characteristic, but DDPG and SAC do not.

\clearpage
\section{Explanations for Patient's Predictions Across 10 Years}
\label{app:shap_patient_global}

The following figures show SHAP plots for a representative patient for predictions made by the RL method for 10 years. Our proposed framework can not only generate SHAP plots for the whole patient dataset but can also filter states and predicted actions to generate a local picture of the model's predictions on a patient. Figure 8 shows the Beeswarm plots, which indicate that for model gave the most importance to the size of the region whose cognition is being predicted. Higher values of $X_{HC}$ (shown in red) increased the resulting information processing for $\Delta I_{HC}$ whereas higher values of information processing at the previous timestep (year) led to a decrease in the change in information processing $\Delta I_{HC}$, signaling that the higher the last known value, the more decline would be anticipated by the model. 

\begin{figure}[ht!]
    \centering
    \begin{subfigure}{0.3\textwidth}
        \includegraphics[width=\linewidth]{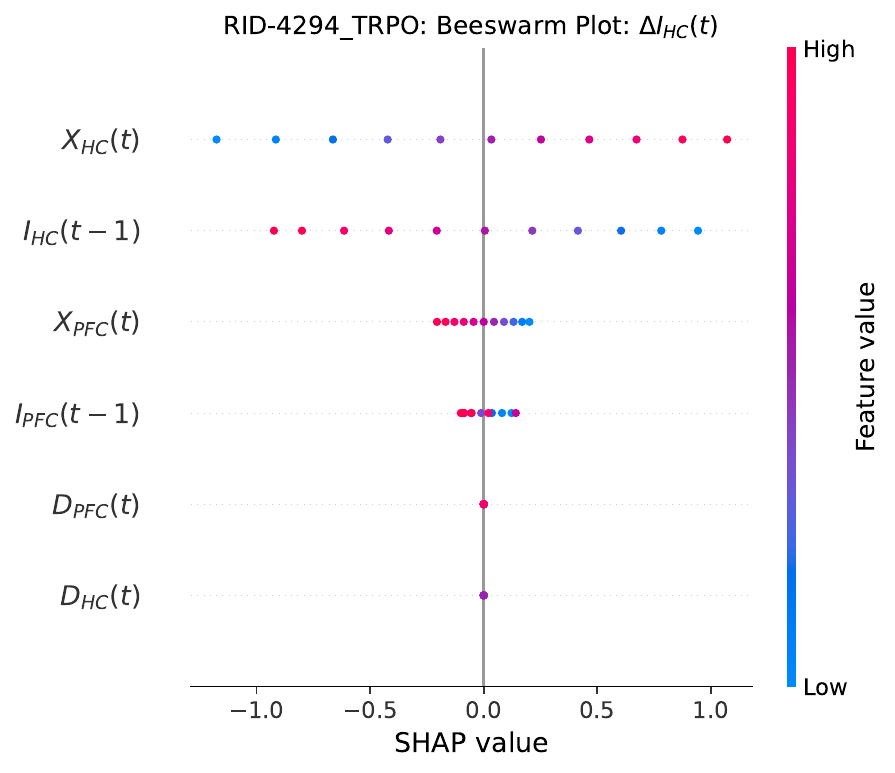}
    \end{subfigure}%
    \hspace{2cm}
    \begin{subfigure}{0.3\textwidth}
        \includegraphics[width=\linewidth]{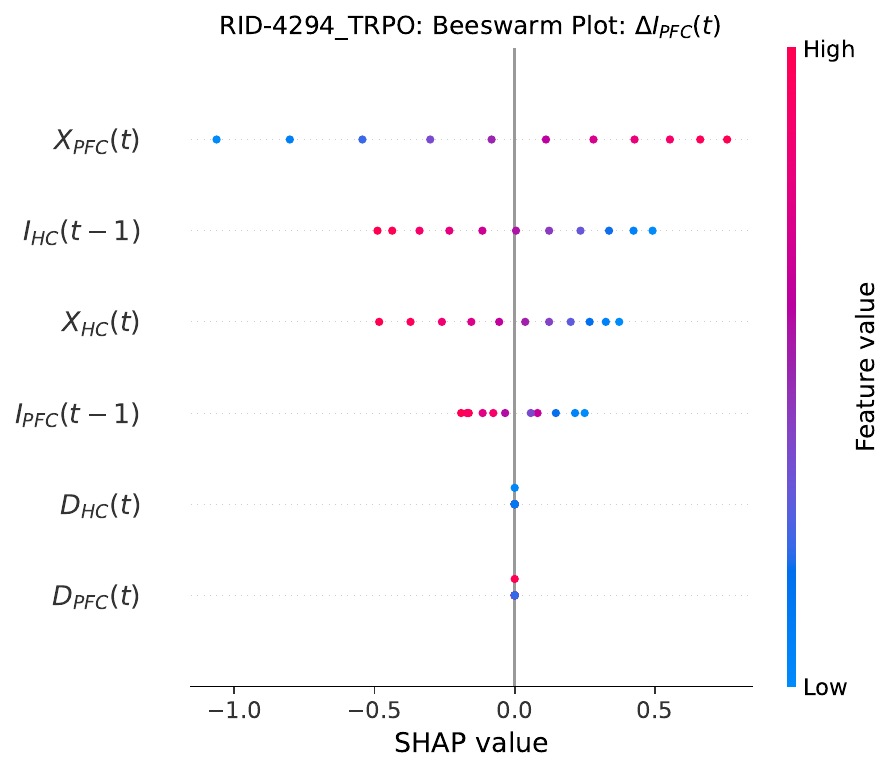}
    \end{subfigure}
    \caption{Beeswarm plots for a representative patient. Predictions made by a trained TRPO agent. Each dot corresponds to predictions for one of the 11 years including baseline year 0 for the two actions (change in cognition in brain region).} 
    
    \begin{subfigure}{0.3\textwidth}
        \includegraphics[width=\linewidth]{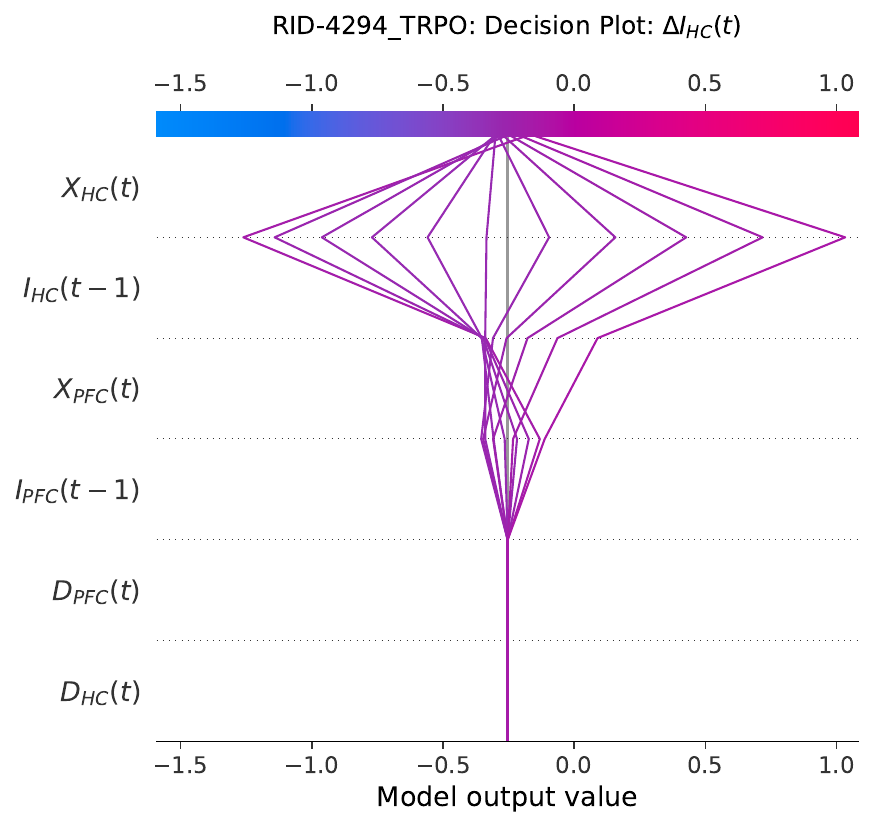}
    \end{subfigure}%
    \hspace{2cm}
    \begin{subfigure}{0.3\textwidth}
        \includegraphics[width=\linewidth]{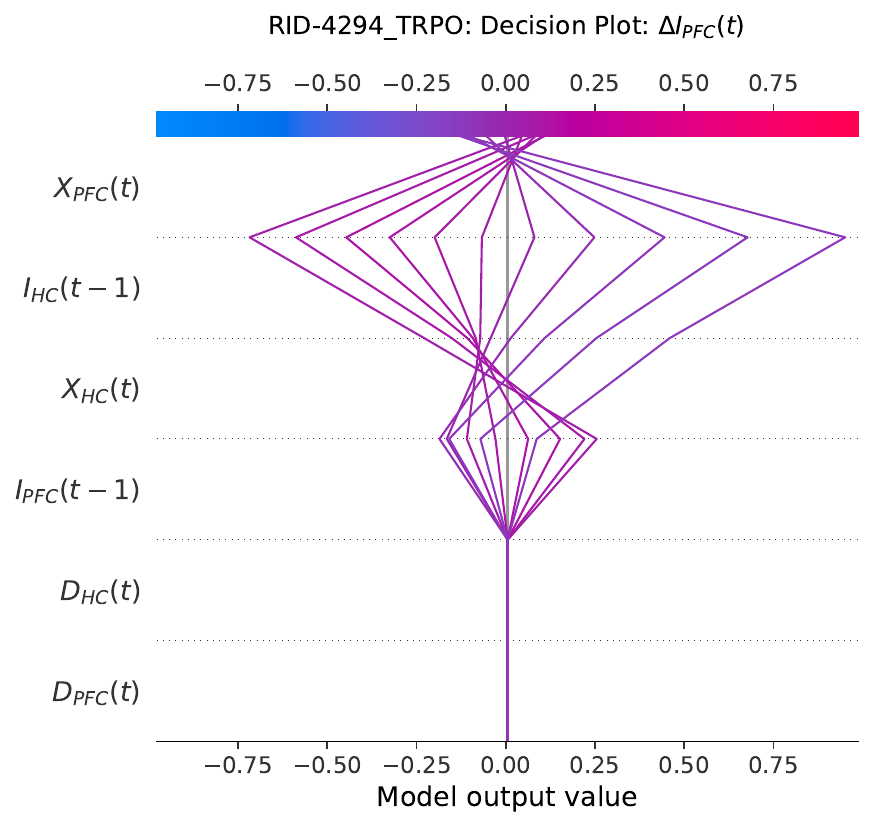}
    \end{subfigure}
    \caption{Decision plots for a representative patient. Predictions made by a trained TRPO agent. Feature influence per prediction. Each line corresponds to predictions for one of the 11 years including baseline year 0 for the two actions (change in cognition in brain region).}

    \begin{subfigure}{0.3\textwidth}
        \includegraphics[width=\linewidth]{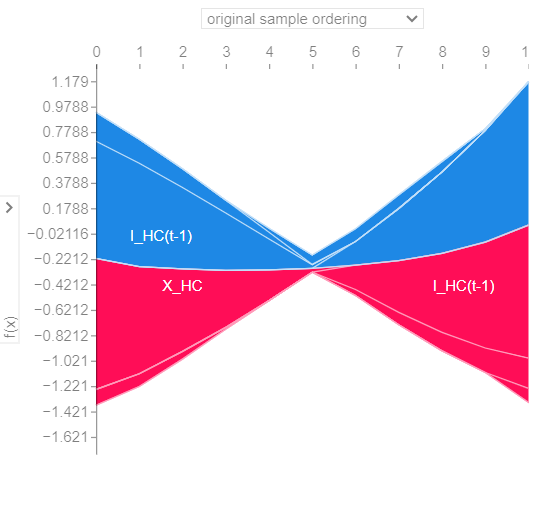}
    \end{subfigure}%
    \hspace{2cm}
    \begin{subfigure}{0.3\textwidth}
        \includegraphics[width=\linewidth]{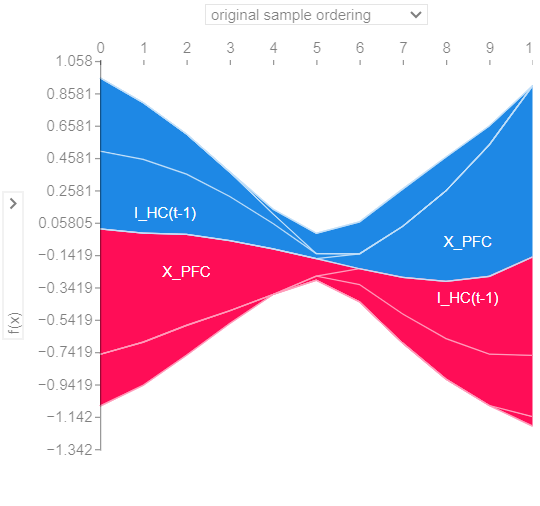}
    \end{subfigure}\\
    \caption{Stacked force plots for a representative patient. Predictions made by a trained TRPO agent. Each dot corresponds to predictions for each of the 11 years including baseline year 0 for the two actions (change in cognition in brain region).}
    \label{fig:shap_patient_global}
\end{figure}

The decision plots show how feature values contributed individually via SHAP values to the model output. Each line depicts one prediction (for a year). The stacked force plots show the contribution of each input feature over 10 years starting from the baseline year, where the function output $f(x)$ is depicted by the line where the blue and red colors meet. As also shown in Figures 7c and 7d of the main paper (where only one feature's effect is shown), the recovery mechanism is visible here as well. For example,  $I_{HC}(t-1)$ from years 1 to 5 pushed the model prediction lower (indicated by the blue color) from the average predicted value of the model, meaning it decreased the predicted cognition value. However, after year 5, it can be seen to be pushing the prediction higher (indicated by the red color).

\section{Explanations for Patient's Predictions for Specific Years}
\label{app:shap_patient_per_year}

\begin{figure}[ht!]
    \centering
    \begin{subfigure}{0.24\textwidth}
        \includegraphics[width=\linewidth]{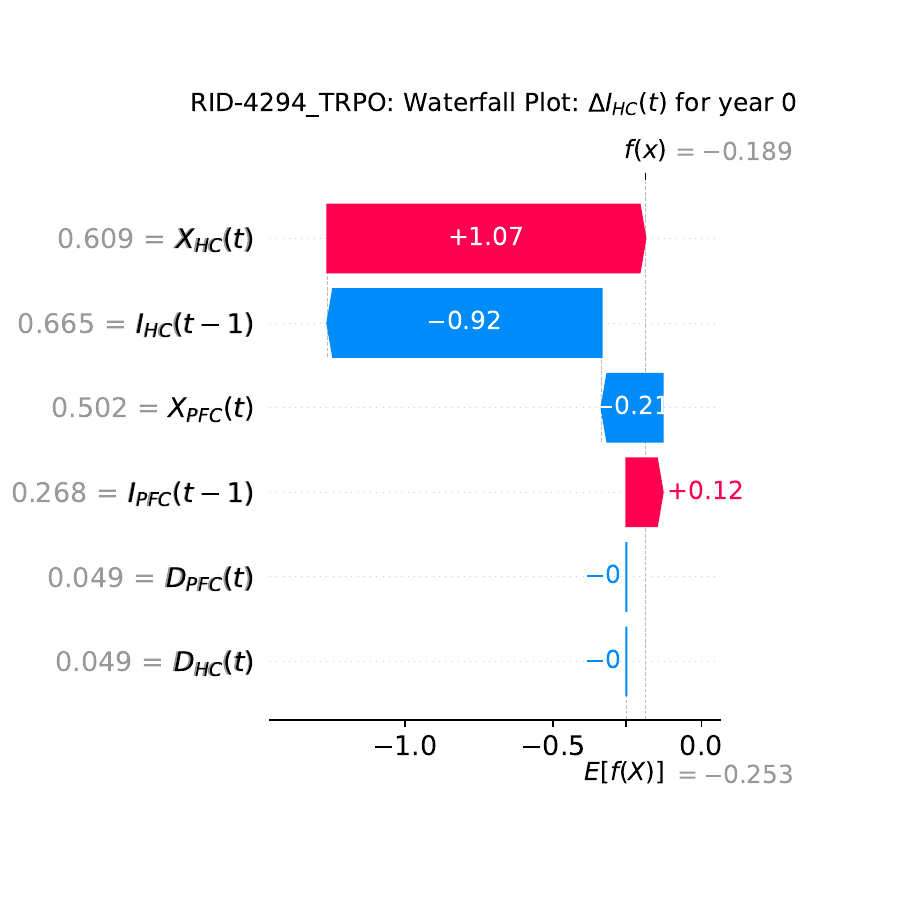}
    \end{subfigure}%
    \begin{subfigure}{0.24\textwidth}
        \includegraphics[width=\linewidth]{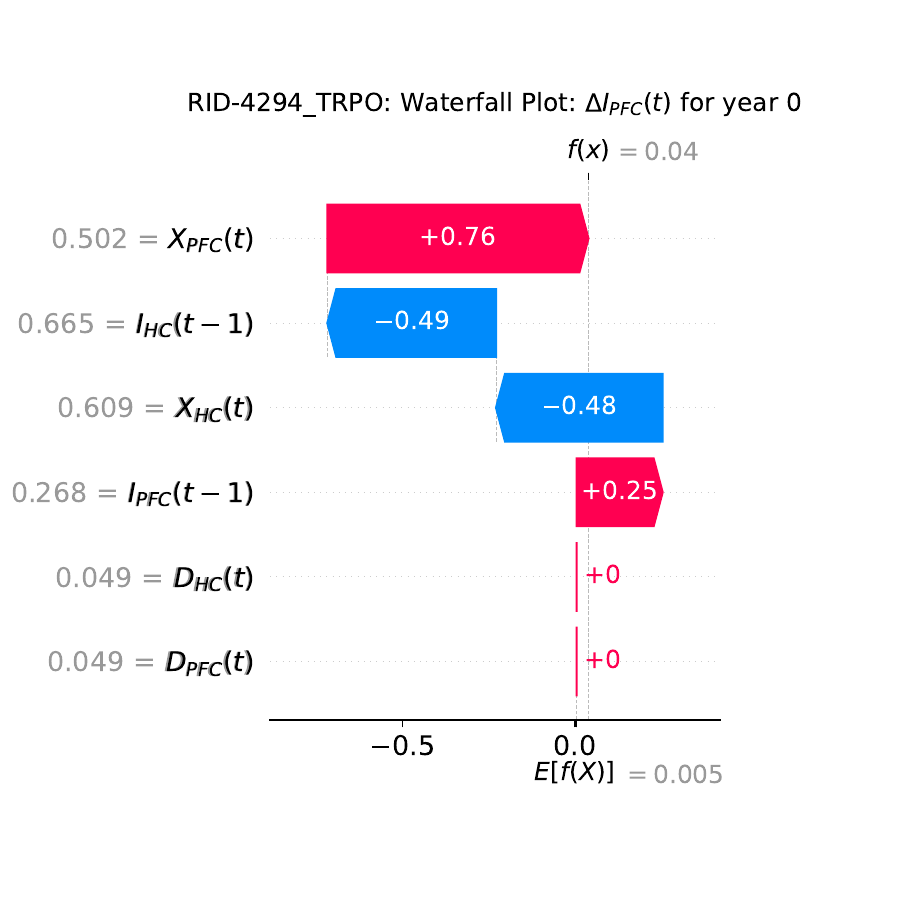}
    \end{subfigure}%
    \begin{subfigure}{0.24\textwidth}
        \includegraphics[width=\linewidth]{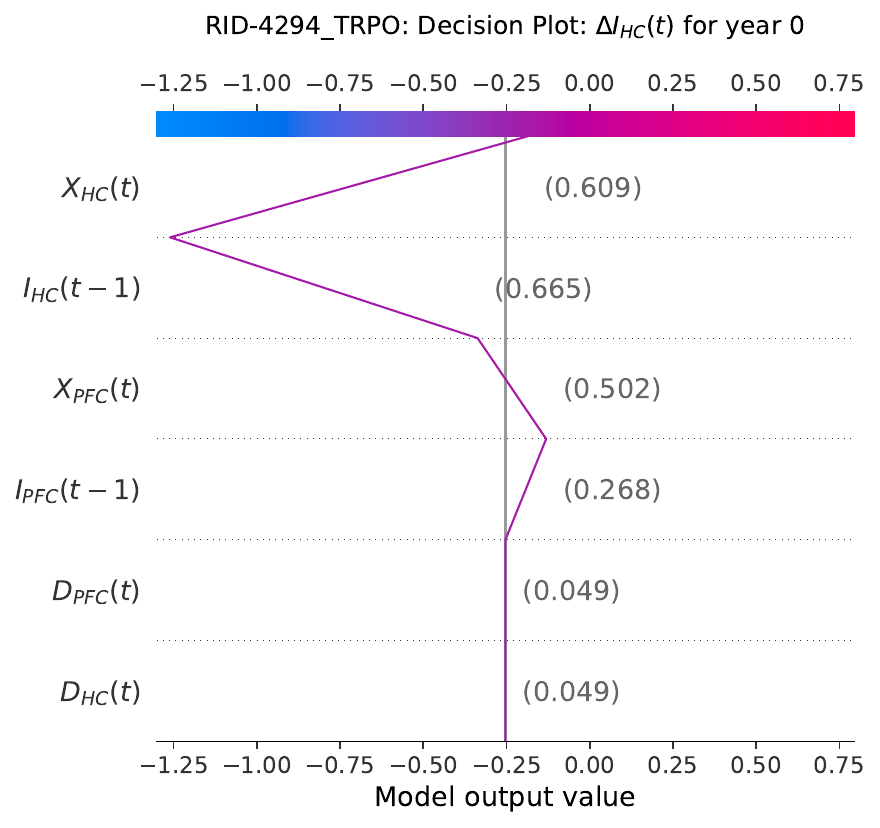}
    \end{subfigure}%
    \begin{subfigure}{0.24\textwidth}
        \includegraphics[width=\linewidth]{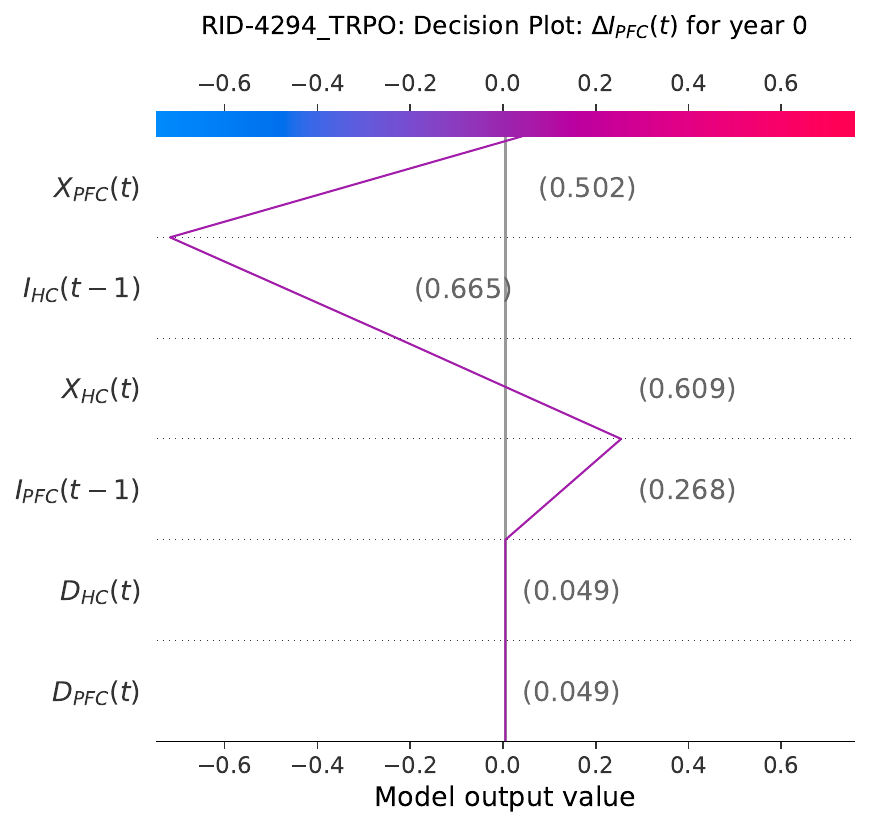}
    \end{subfigure}\\

     \begin{subfigure}{0.24\textwidth}
        \includegraphics[width=\linewidth]{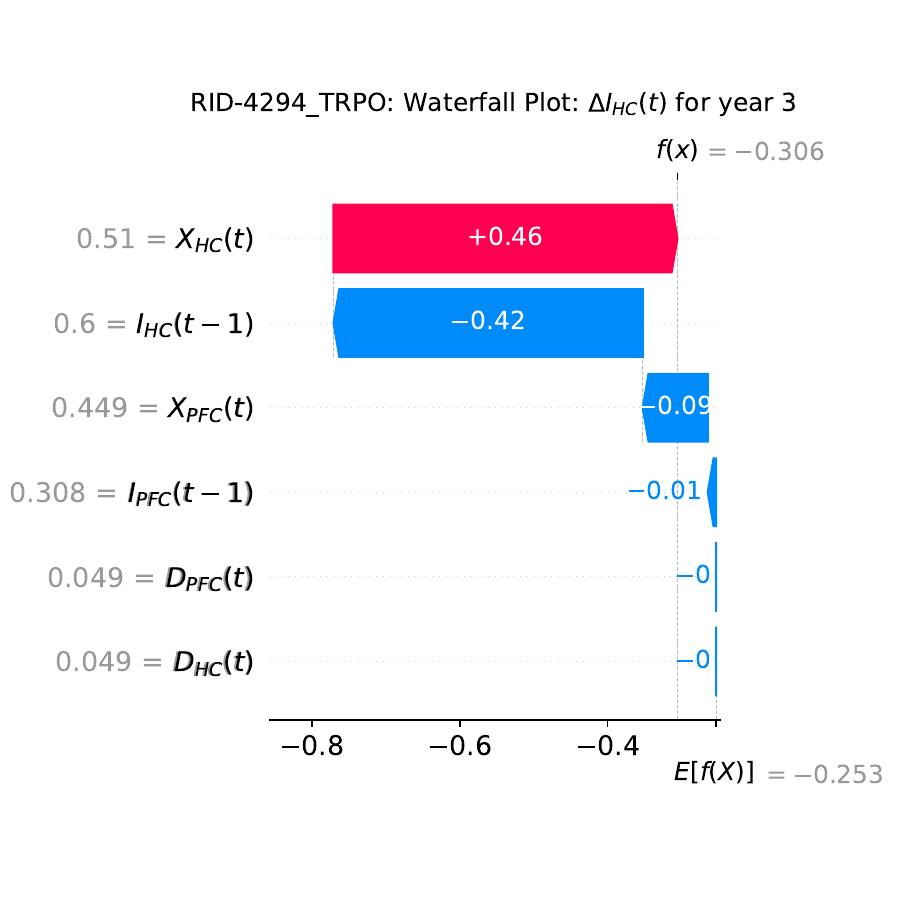}
    \end{subfigure}%
    \begin{subfigure}{0.24\textwidth}
        \includegraphics[width=\linewidth]{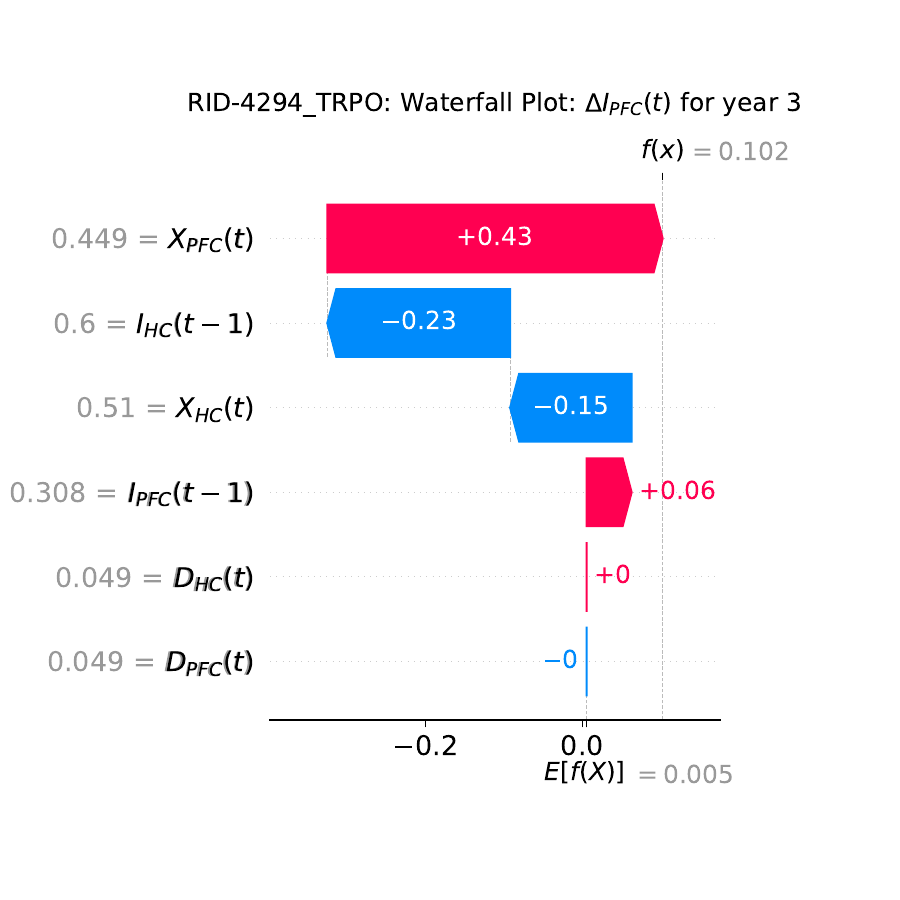}
    \end{subfigure}%
    \begin{subfigure}{0.24\textwidth}
        \includegraphics[width=\linewidth]{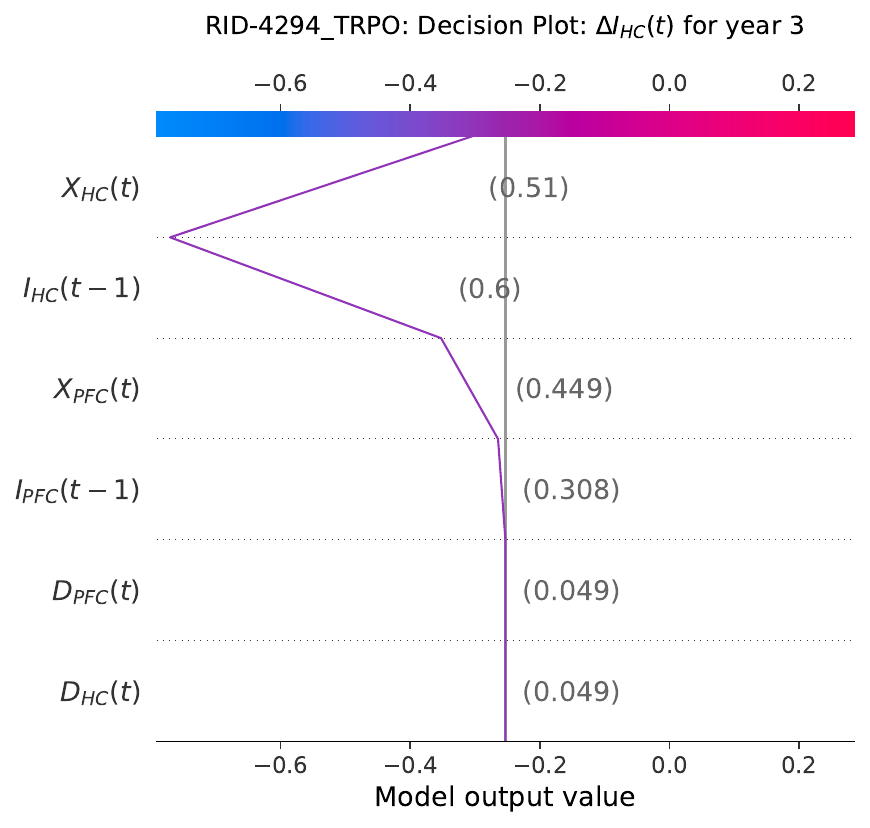}
    \end{subfigure}%
    \begin{subfigure}{0.24\textwidth}
        \includegraphics[width=\linewidth]{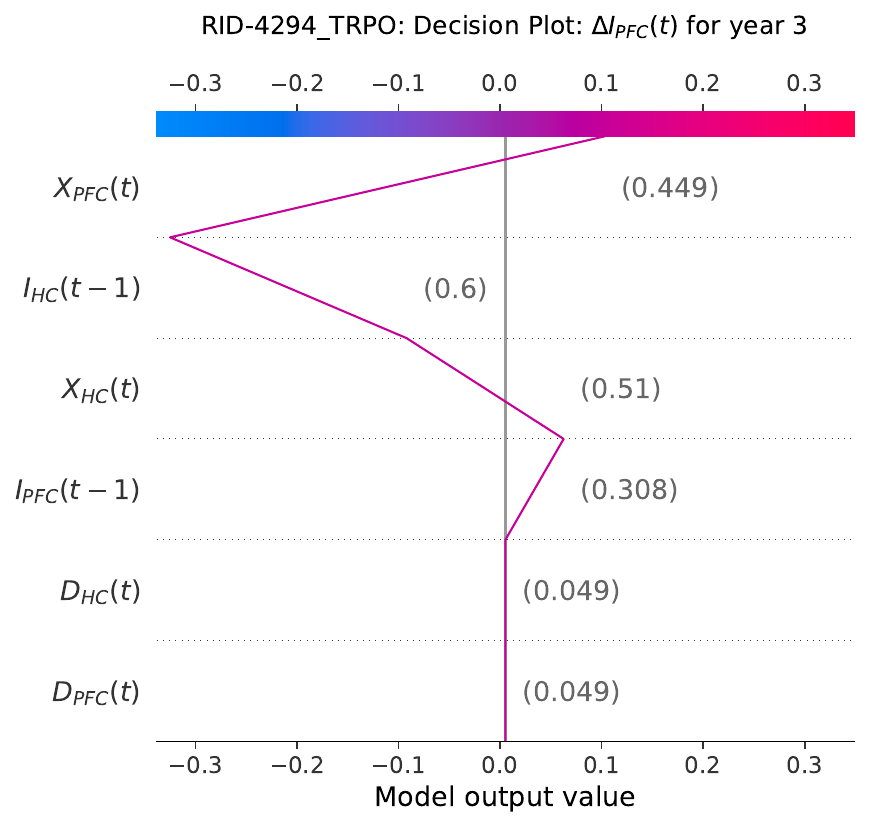}
    \end{subfigure}\\
    
    \begin{subfigure}{0.24\textwidth}
        \includegraphics[width=\linewidth]{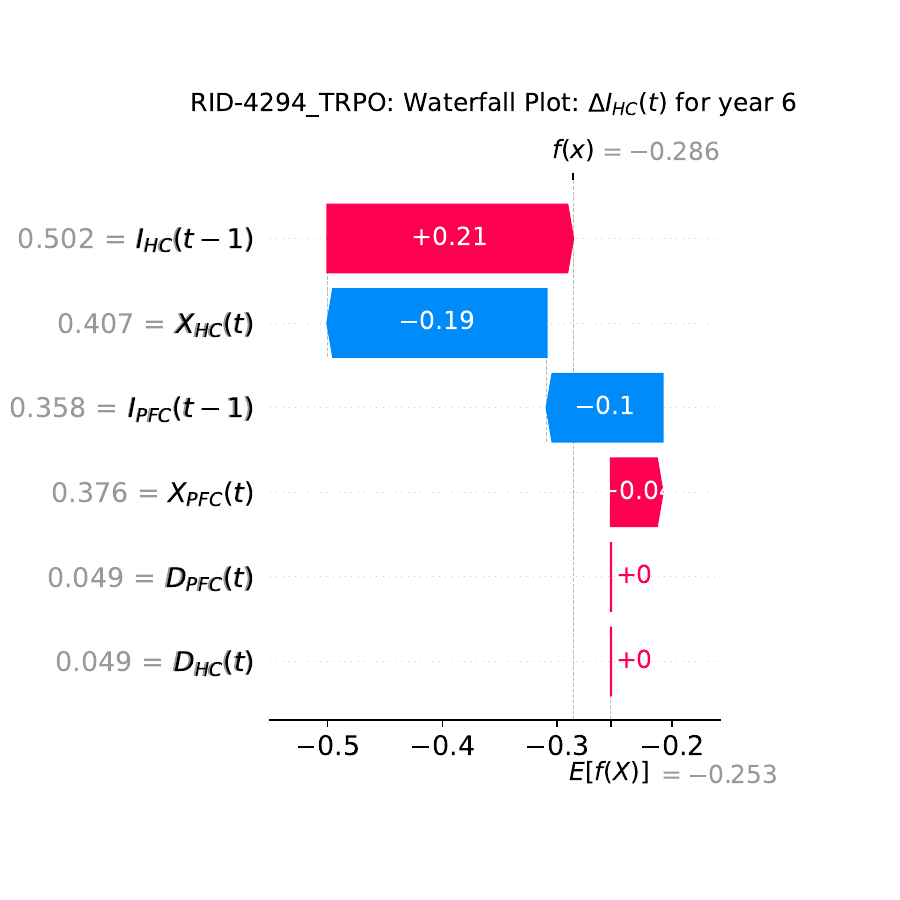}
    \end{subfigure}%
    \begin{subfigure}{0.24\textwidth}
        \includegraphics[width=\linewidth]{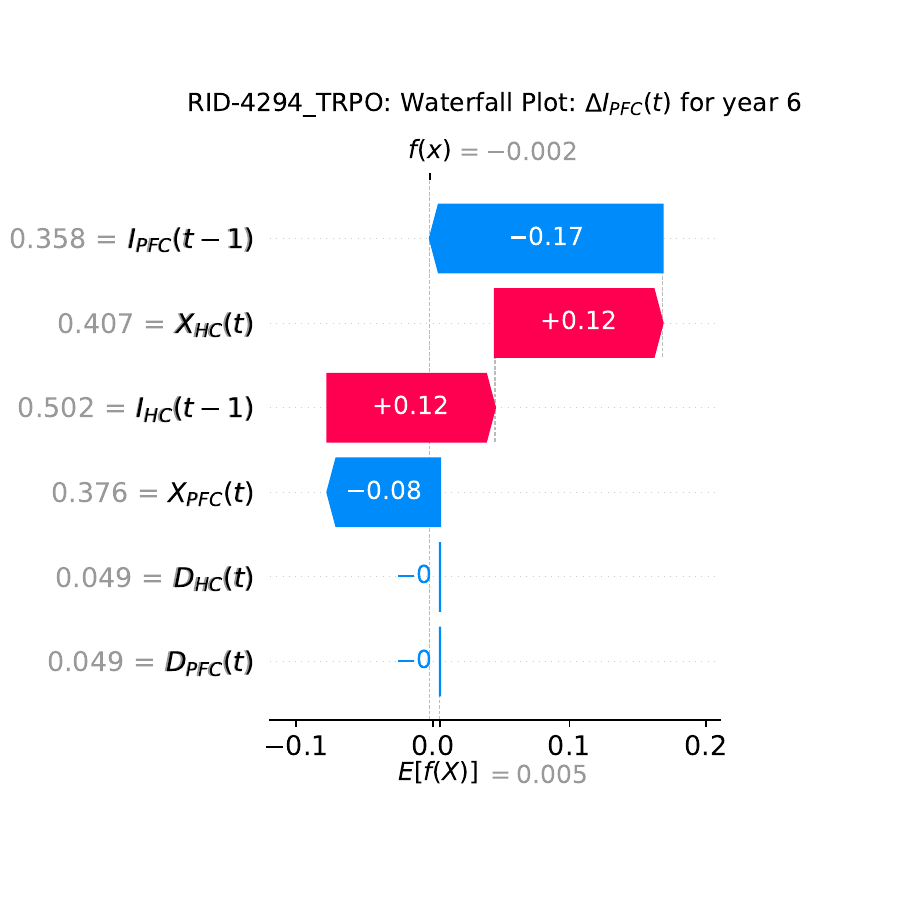}
    \end{subfigure}%
    \begin{subfigure}{0.24\textwidth}
        \includegraphics[width=\linewidth]{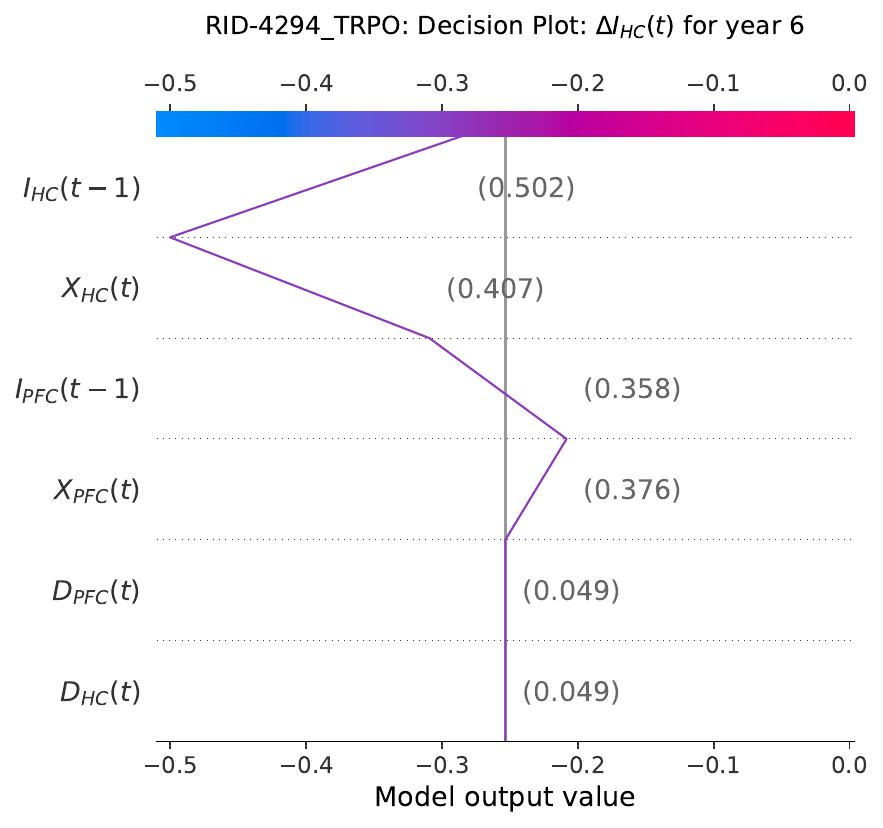}
    \end{subfigure}%
    \begin{subfigure}{0.24\textwidth}
        \includegraphics[width=\linewidth]{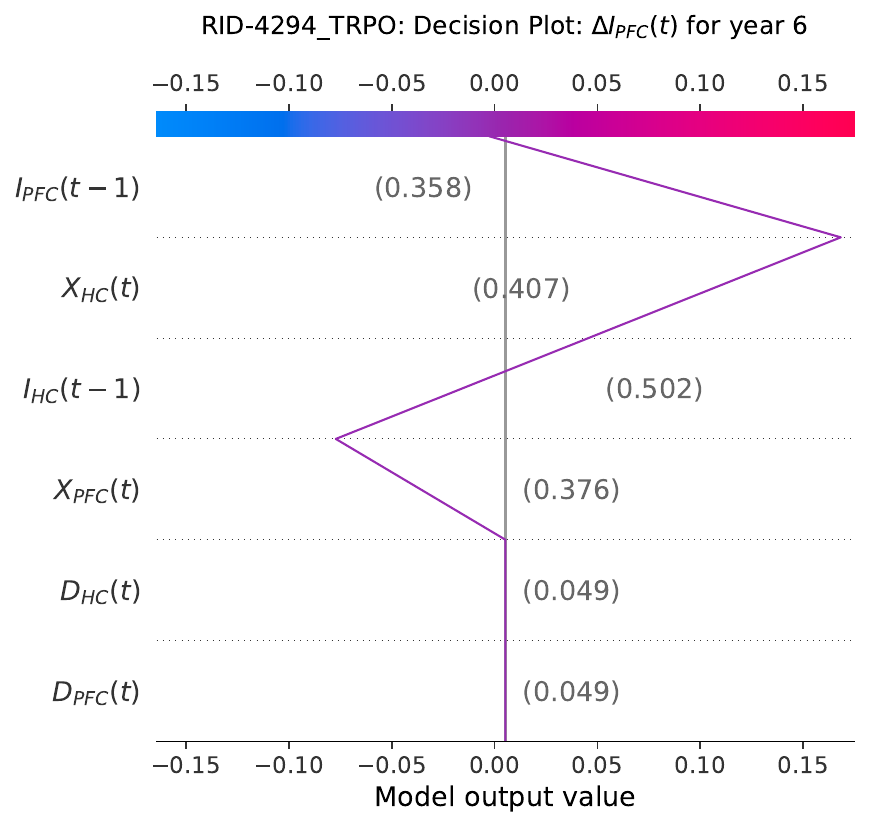}
    \end{subfigure}\\

    \begin{subfigure}{0.24\textwidth}
        \includegraphics[width=\linewidth]{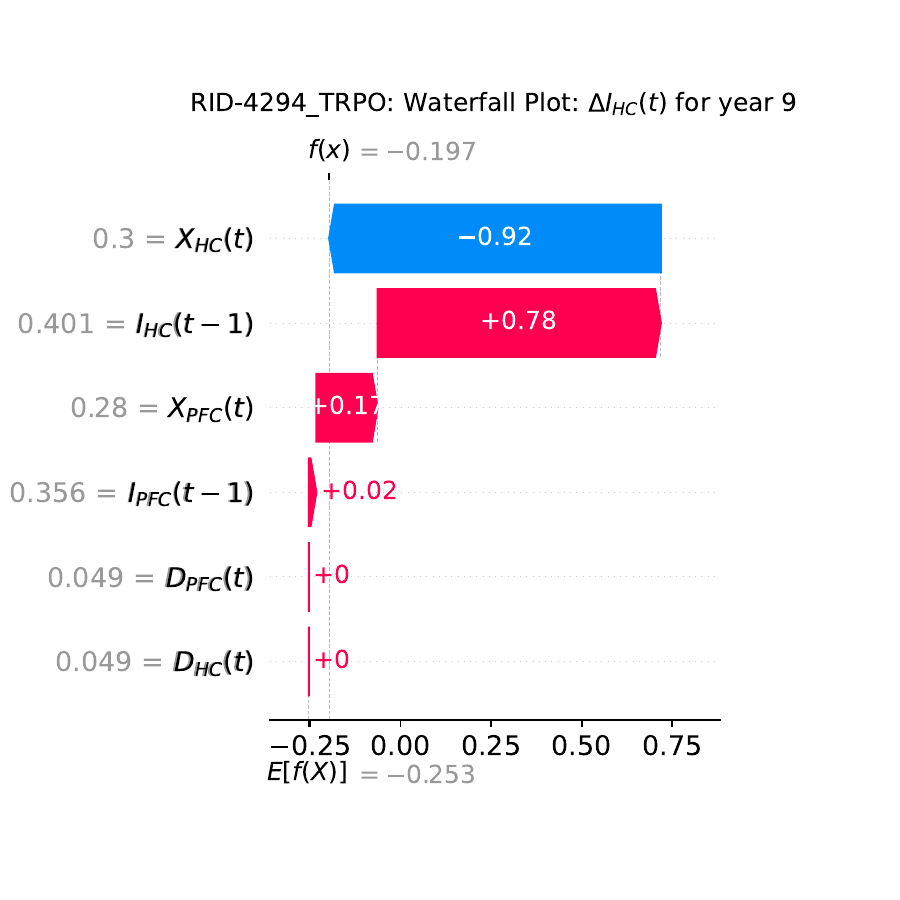}
    \end{subfigure}%
    \begin{subfigure}{0.24\textwidth}
        \includegraphics[width=\linewidth]{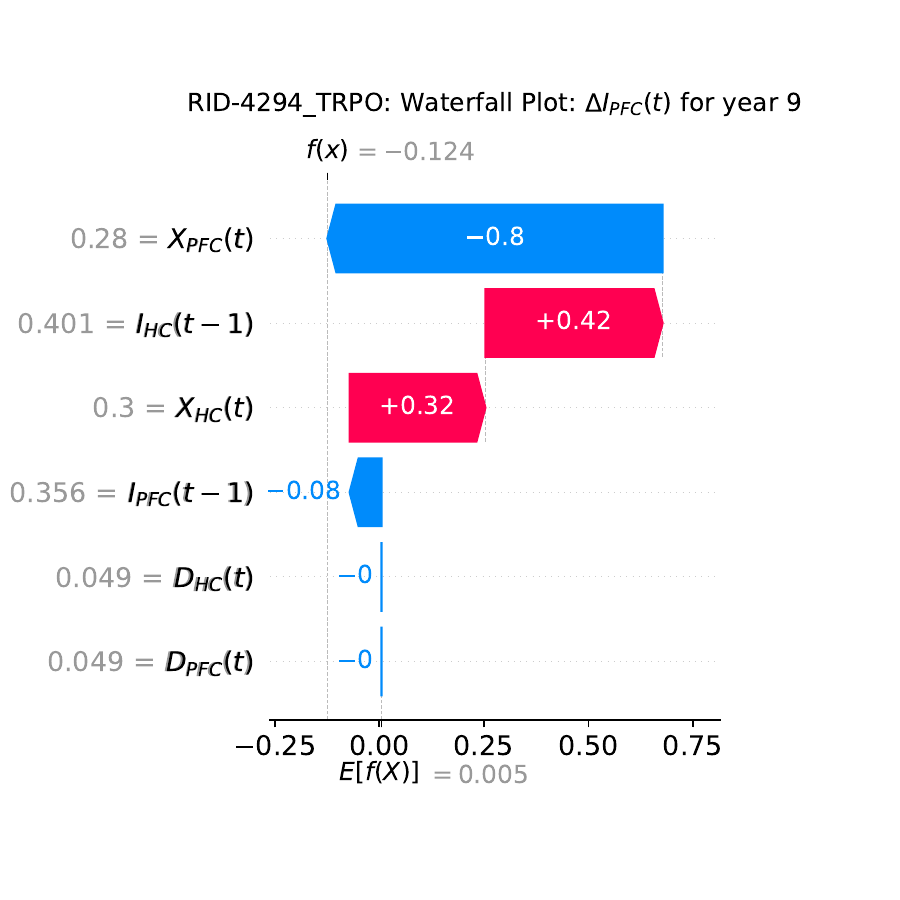}
    \end{subfigure}%
    \begin{subfigure}{0.24\textwidth}
        \includegraphics[width=\linewidth]{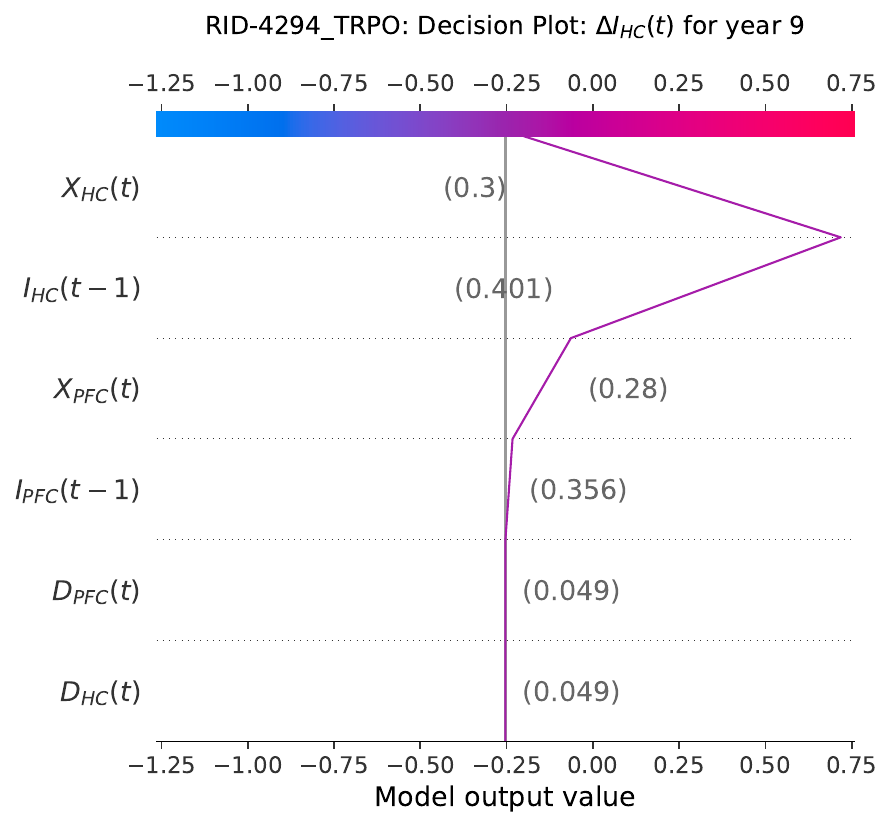}
    \end{subfigure}%
    \begin{subfigure}{0.24\textwidth}
        \includegraphics[width=\linewidth]{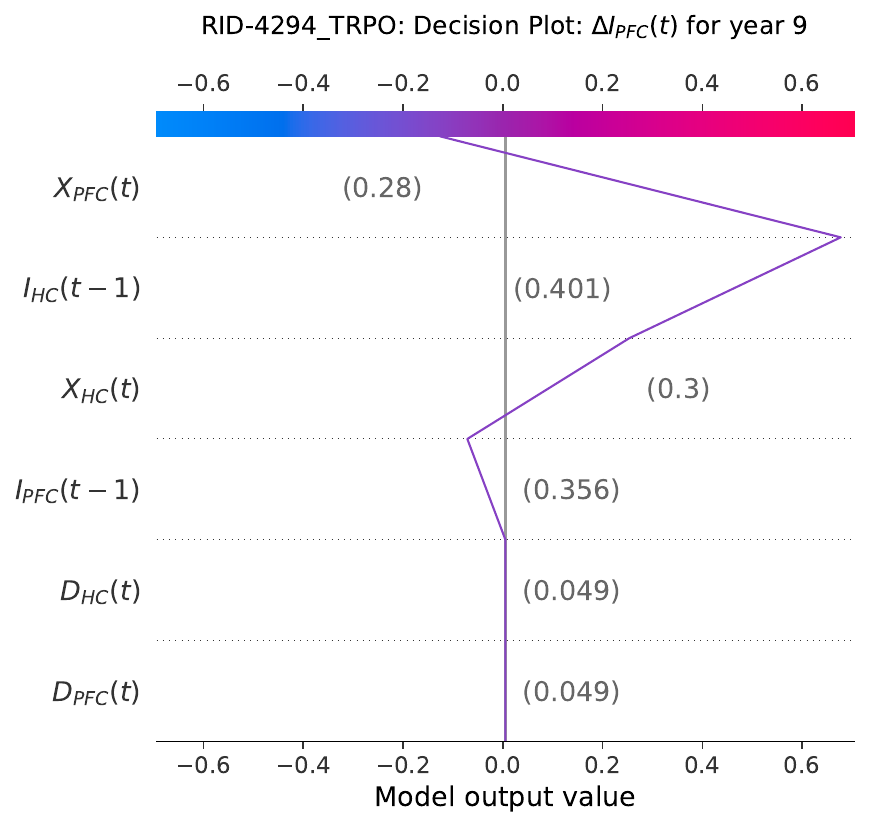}
    \end{subfigure}\\
    
    \caption{SHAP plots for a representative patient sample over 10 years (years 0, 3, 6, and 9 shown here). These local explanations show the decision-making of the framework for predicting the change in cognition in each of the two studied brain regions. Years from top to bottom. \textbf{Left Column}: Waterfall Plot for $\Delta I{HC}$, \textbf{Left Center}: Waterfall Plot for $\Delta I{PFC}$, \textbf{Right Center Column}: Decision Plot for $\Delta I{HC}$, \textbf{Right Column}: Decision Plot for $\Delta I{PFC}$,}
    \label{fig:shap_patient_local_waterfall_decision}
\end{figure}

\begin{figure}[ht!]
    \centering
    \begin{subfigure}{0.7\textwidth}
        \includegraphics[width=\linewidth]{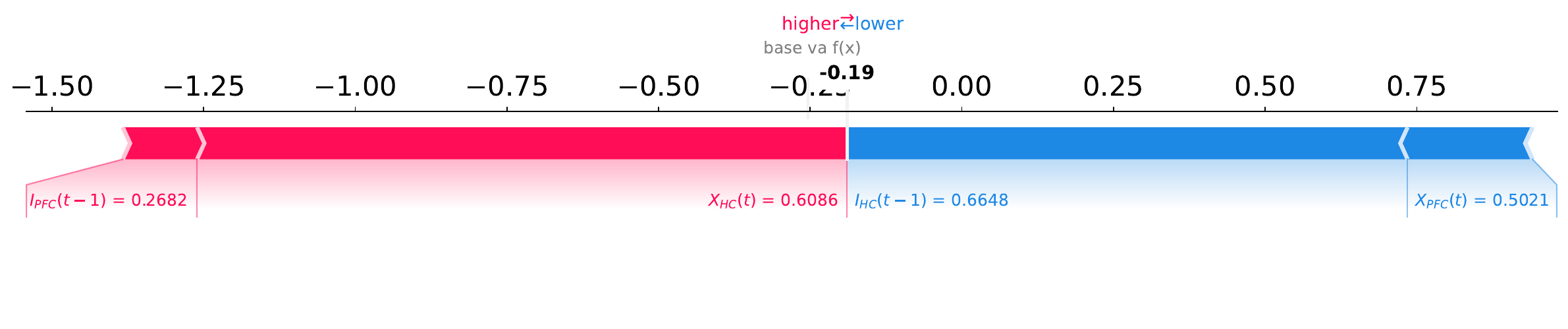}
    \end{subfigure}
	\begin{subfigure}{0.7\textwidth}
        \includegraphics[width=\linewidth]{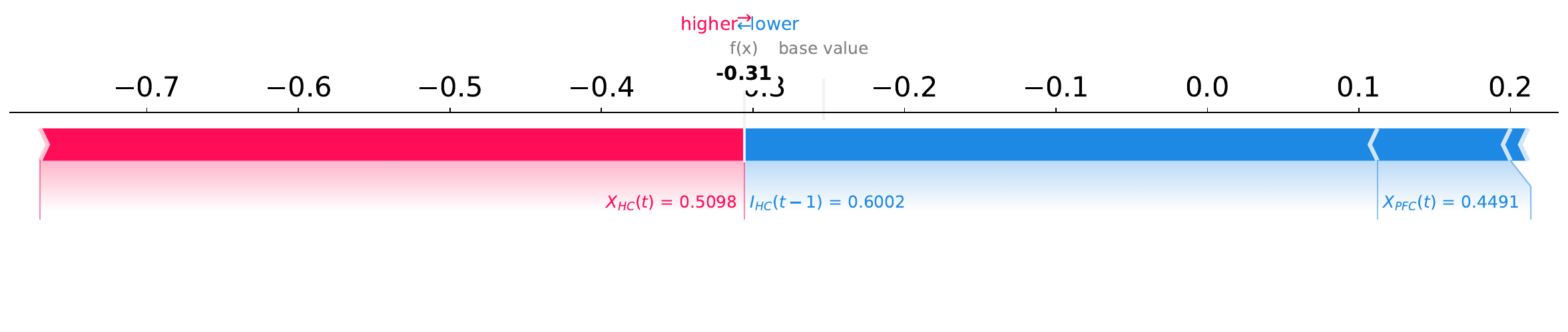}
    \end{subfigure}
	\begin{subfigure}{0.7\textwidth}
        \includegraphics[width=\linewidth]{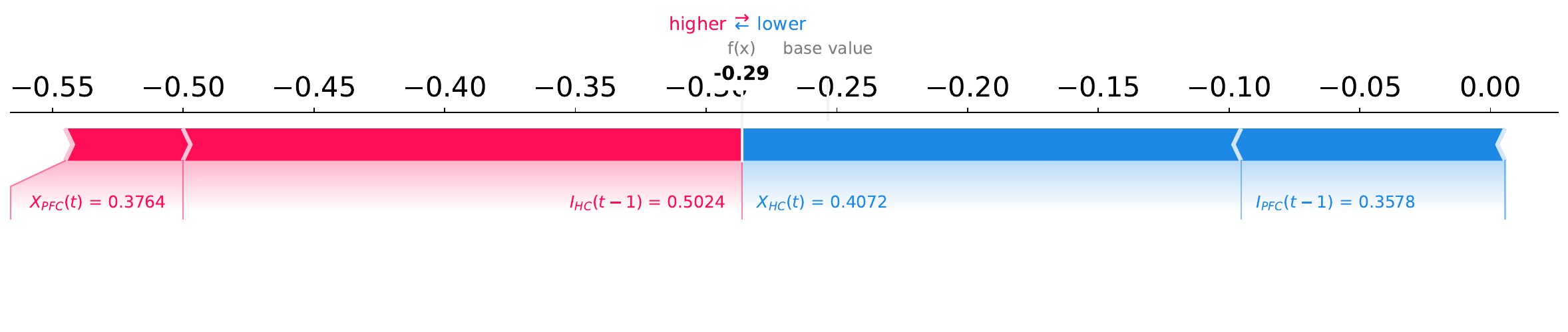}
    \end{subfigure}
	\begin{subfigure}{0.7\textwidth}
        \includegraphics[width=\linewidth]{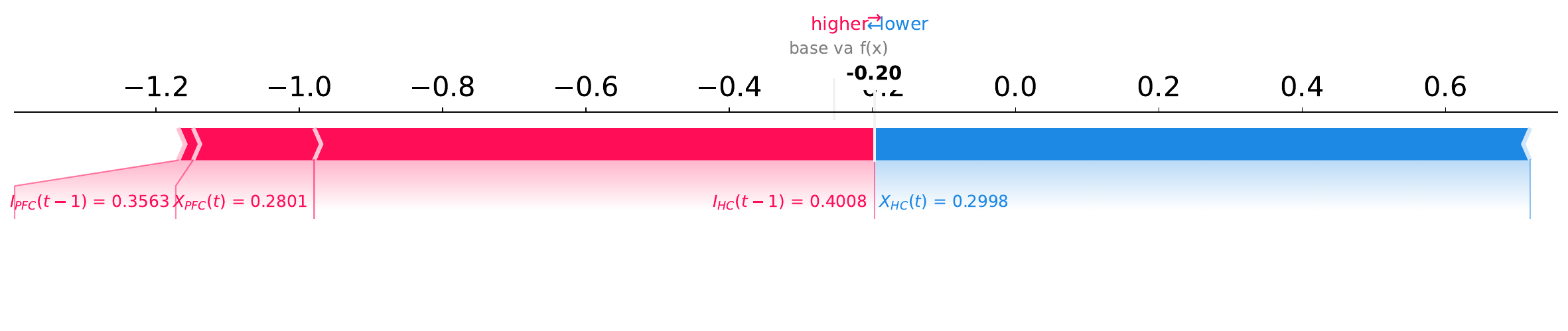}
        \caption{Force Plot for $\Delta I{HC}$ for years 0, 3, 6, and 9 (top to bottom)}
    \end{subfigure}\\
	\vspace{1cm}
    \begin{subfigure}{0.7\textwidth}
        \includegraphics[width=\linewidth]{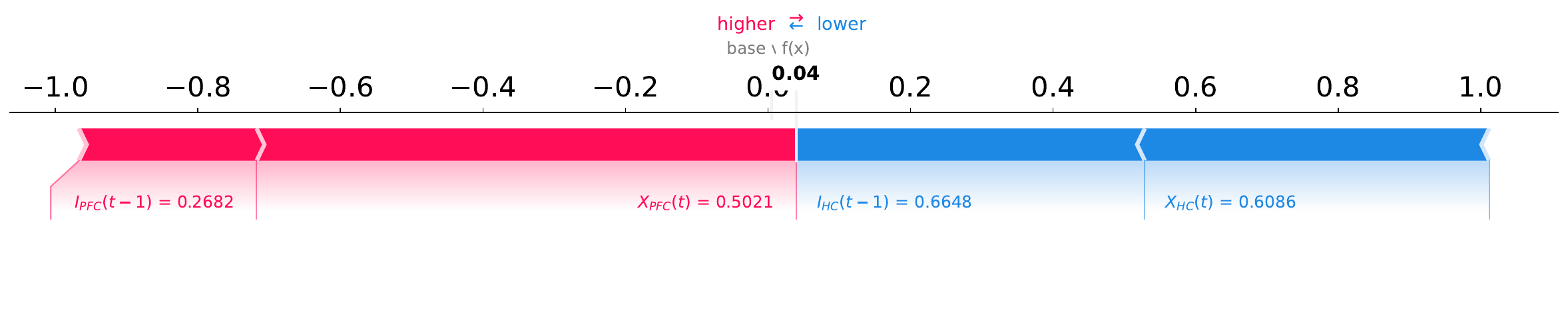}
    \end{subfigure}
    \begin{subfigure}{0.7\textwidth}
        \includegraphics[width=\linewidth]{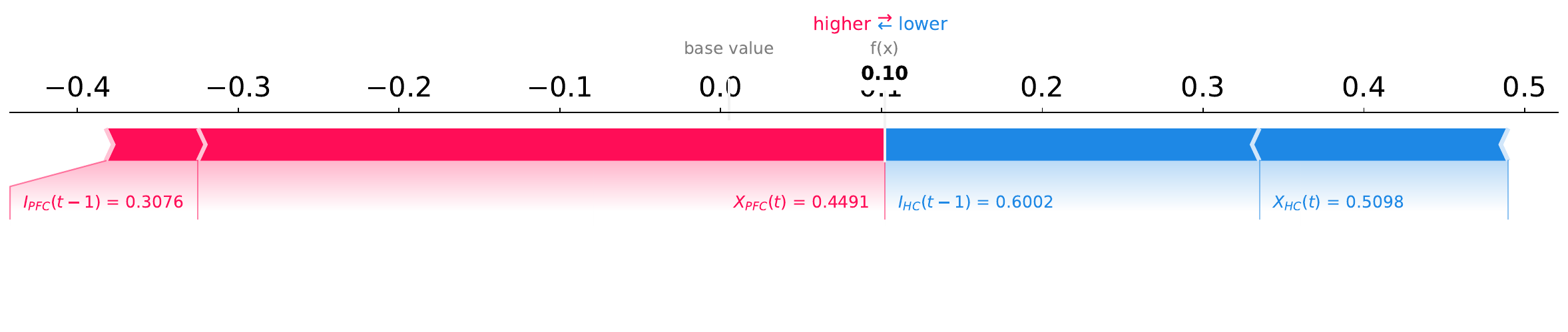}
    \end{subfigure}
    \begin{subfigure}{0.7\textwidth}
        \includegraphics[width=\linewidth]{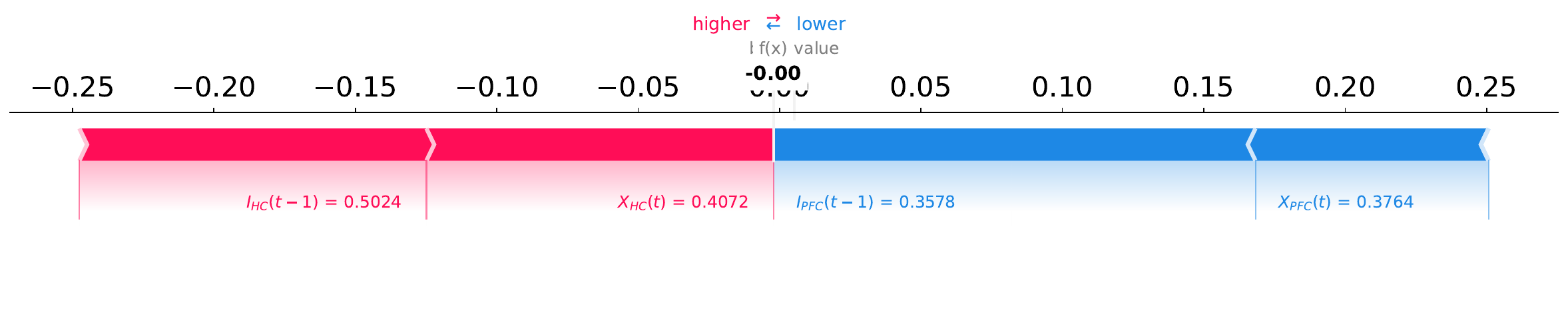}
    \end{subfigure}
    \begin{subfigure}{0.7\textwidth}
        \includegraphics[width=\linewidth]{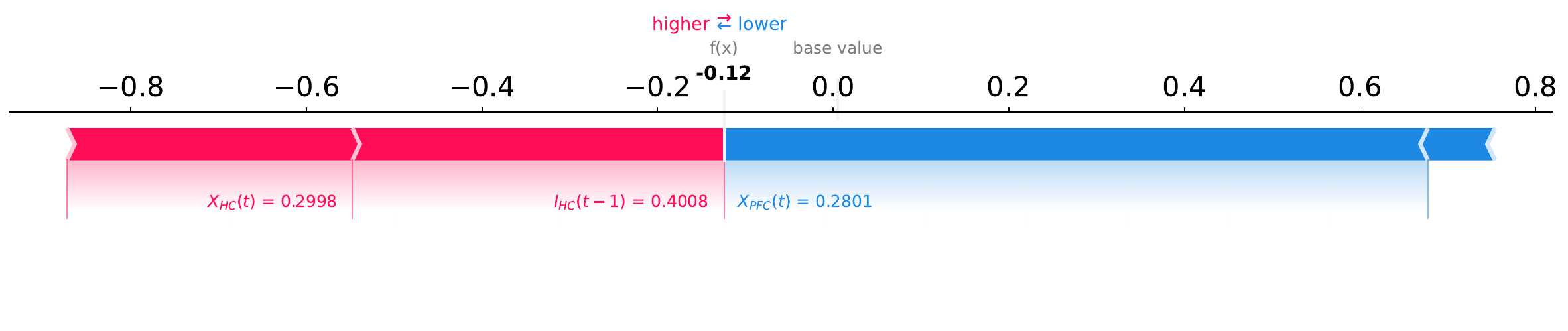}
        \caption{Force Plot for $\Delta I{PFC}$ for years 0, 3, 6, and 9 (top to bottom)}
    \end{subfigure}
	
    \caption{SHAP force plots for a representative patient sample over 10 years (years 0, 3, 6, and 9 shown here). These local explanations show the decision-making of the framework for predicting the change in cognition in each of the two studied brain regions and show how each input feature/state contributed towards the final prediction.}
    \label{fig:shap_patient_local_force}
\end{figure}

Figures \ref{fig:shap_patient_local_waterfall_decision} and \ref{fig:shap_patient_local_force} are explanations of predictions for a representative patient for the years 0, 3, 6, and 9. 

\clearpage

\section{Additional Related Work}
\label{app:additional_related_work}

\subsection{Explainable AI for AD Classification}
\label{app:related_work_XAI_AD}

Along with the increased application of AI to AD diagnosis and progression prediction, there has been a recent surge in research dedicated to the explainability of these models \cite{viswan2024explainable}. A bulk of these efforts have focused on adding explainability to \textit{opaque}, data-driven machine learning models. 
These commonly include the application of tree-based methods like XGBoost and Random Forest on patient datasets for diagnosis classification (e.g., MCI to AD) with the addition of SHAP or LIME \cite{ribeiro2016should} for explainability \cite{danso2021developing,elsappagh2021multilayer,hernandez2022explainable,bogdanovic2022depth}, or the use of deep learning methods like CNNs, Attention Networks, and Vision Transformers on multimodal data including MRI imaging \cite{essemlali2020understanding,qiu2020development,zhang2021explainable,khatri2023explainable} and adding explainability via counterfactual maps \cite{oh2022learn} or post-hoc attention methods (e.g., Grad-CAM, Score-CAM) \cite{jahan2023comparison}. 

Some other works explore combining MRI imaging with gene expression data while using LIME for explainability of the latter \cite{kamal2021alzheimer}, using polygenic risk scores and conventional risk factors to predict AD while employing SHAP for explainability \cite{gao2023explainable}, assessing the impact of cognitive and clinical measures (e.g., ADAS, MMSE, MOCA, FAQ, RAVLT, Ecog) on diagnosis classification \cite{lombardi2022robust} and adding rule-extraction approaches for model explainability while validating them with SHAP/LIME \cite{alatrany2024explainable}. However, as discussed in Section \ref{sec:introduction}, explanations generated on top of data-driven models alone without incorporating domain knowledge can be misleading and untrustworthy since they are only revealing correlations the model identified as factors behind AD progression, not all the factors or their causal relationships.

\subsection{Surveys on eXplainable RL (XRL)}
\label{app:xrl_surveys}
Research in XRL can be categorized in various ways. The most prominent method of categorization splits XRL works into (a) transparent methods and (b) post-hoc explainability \cite{arrieta2020explainable}. Transparent methods include RL models that can be explained by themselves, and post-hoc explainability provides explanations of RL algorithms after the training phase. This work uses the SHAP method, which can be categorized as a post-hoc explainability method with Interaction Data to explain the RL model's predictions \cite{heuillet2021explainability}. SHAP uses the magnitude of influence from each variable in the environment after training to quantify the interactions between and contributions of each variable towards the final prediction of the model. 

XRL methods can also fall into categories of (a) Feature Importance - FI (b) Learning Process and MDP - LPM and (c) Policy-Level - PL \cite{milani2023explainable}. FI explanations describe the reasoning behind taking an action, LPM explanations describe the particularly influential experiences of the model, and PL explanations summarize the long-term behavior of the model. Each category is then broken up into subcategories. In particular, the FI category is broken up into (a) Learn Intrinsically Interpretable Policy (b) Convert to Interpretable Format, and (c) Directly Generate Explanation. SHAP falls within the ``Directly Generate Explanation" subcategory within FI. SHAP generates an explanation after training from a non-interpretable policy. This enables the understanding of the factors that influence a model towards its final predictions. Similarly, as per \cite{qing2022survey}, SHAP falls into the \emph{Model-Explaining and Explanation-Generating} category. This category describes methods that generate explanations from the model without being explicitly self-explainable. 

\section{ADNI Acknowledgment - Standard Statement}
\label{app:adni_ack}
Data used in the preparation of this article were obtained from the Alzheimer’s Disease Neuroimaging Initiative (ADNI) \cite{adni} database (\href{https://adni.loni.usc.edu}{https://adni.loni.usc.edu}). As such, the investigators within the ADNI contributed to the design and implementation of ADNI and/or provided data but did not participate in the analysis or writing of this report. A complete listing of ADNI investigators can be found at: \url{https://adni.loni.usc.edu/wp-content/uploads/how_to_apply/ADNI_Acknowledgement_List.pdf}. The ADNI was launched in 2003 as a public-private partnership, led by Principal Investigator Michael W. Weiner, MD. The primary goal of ADNI has been to test whether serial magnetic resonance imaging (MRI), positron emission tomography (PET), other biological markers, and clinical and neuropsychological assessment can be combined to measure the progression of mild cognitive impairment (MCI) and early Alzheimer’s disease (AD).

\end{document}